\def\year{2021}\relax
\documentclass[letterpaper]{article} 
\usepackage{aaai21}  
\usepackage{times}  
\usepackage{helvet} 
\usepackage{courier}  
\usepackage[hyphens]{url}  
\usepackage{graphicx} 
\usepackage{natbib}  
\usepackage{caption} 
\usepackage{amssymb}
\usepackage{amsmath}
\usepackage{amsmath}
\usepackage{amsthm}
\usepackage[table]{xcolor}
\usepackage{algorithm}
\usepackage{algorithmic}
\usepackage[switch]{lineno}  %
\usepackage{microtype}
\newtheorem{theorem}{Theorem}
\newtheorem*{theorem*}{Theorem}

\newtheorem{lemma}{Lemma}
\newtheorem*{lemma*}{Lemma}
\usepackage{xcolor}

\newcommand{\model}{\mbox{\textsc{Glister}}}
\newcommand{\modelonline}{\mbox{\textsc{Glister-Online}}}
\newcommand{\modelactive}{\mbox{\textsc{Glister-Active}}}
\usepackage{subcaption}
\usepackage{comment}
\usepackage{adjustbox}
\title{\model: Generalization based Data Subset Selection for Efficient and Robust Learning}

\author {
        Krishnateja Killamsetty\textsuperscript{\rm 1}, 
        Durga Sivasubramanian \textsuperscript{\rm 2},
        Ganesh Ramakrishnan \textsuperscript{\rm 2}, 
        Rishabh Iyer\textsuperscript{\rm 1,2}
        \\
}
\affiliations {
   \textsuperscript{\rm 1} University of Texas at Dallas \\
    \textsuperscript{\rm 2} Indian Institute of Technology, Bombay \\
     krishnateja.killamsetty@utdallas.edu, durgas@cse.iitb.ac.in,  ganesh@cse.iitb.ac.in, rishabh.iyer@utdallas.edu}

\begin{document}
\maketitle
\begin{abstract}
Large scale machine learning and deep models are extremely data-hungry. Unfortunately, obtaining large amounts of labeled data is expensive, and training state-of-the-art models (with hyperparameter tuning) requires significant computing resources and time. Secondly, real-world data is noisy and imbalanced. As a result, several recent papers try to make the training process more efficient and robust. However, most existing work either focuses on robustness or efficiency, but not both. In this work, we introduce \model, a GeneraLIzation based data Subset selecTion for Efficient and Robust learning framework. We formulate \model\ as a mixed discrete-continuous bi-level optimization problem to select a subset of the training data, which maximizes the log-likelihood on a held-out validation set. We then analyze \model\ for simple classifiers such as gaussian and multinomial naive-bayes, k-nearest neighbor classifier, and linear regression and show connections to submodularity. 
Next, we propose an iterative online algorithm \modelonline, which performs data selection iteratively along with the parameter updates and can be applied to any loss-based learning algorithm. We then show that for a rich class of loss functions including \emph{cross-entropy, hinge-loss, squared-loss}, and \emph{logistic-loss}, the inner discrete data selection is an instance of (weakly) submodular optimization, and we analyze conditions for which \modelonline\ reduces the validation loss and converges. Finally, we propose \modelactive, an extension to batch active learning, and we empirically demonstrate the performance of \model\ on a wide range of tasks including, (a) data selection to reduce training time, (b) robust learning under label noise and imbalance settings, and (c) batch-active learning with several deep and shallow models. We show that our framework improves upon state of the art both in efficiency and accuracy (in cases (a) and (c)) and is more efficient compared to other state-of-the-art robust learning algorithms in case (b). The code for \model is at:\url{https://github.com/dssresearch/GLISTER}.

\end{abstract}
\section{Introduction}
With the quest to achieve human-like performance for machine learning and deep learning systems, the cost of training and deploying machine learning models has been significantly increasing. The wasted computational and engineering energy becomes evident in deep learning algorithms, wherein extensive hyper-parameter tuning and network architecture search need to be done. This results in staggering compute costs and running times\footnote{\url{https://medium.com/syncedreview/the-staggering-cost-of-training-sota-ai-models-e329e80fa82}}. 

As a result, efficient and robust machine learning is a very relevant and substantial research problem. In this paper, we shall focus on three goals: 
\noindent \textbf{Goal 1: } Train machine learning and deep learning models on effective subsets of data, thereby significantly reducing training time and compute while not sacrificing accuracy. 
\noindent \textbf{Goal 2: } To (iteratively) select effective subsets of labeled data to reduce the labeling cost. 
\noindent \textbf{Goal 3: } Select data subsets to remove noisy labels and class imbalance, which is increasingly common in operational machine learning settings.

\subsection{Background and Related Work}
A number of papers have studied data efficient training and robust training of machine learning and deep learning models. However, the area of data efficient training of models that are  also robust is relatively under-explored. Below, we summarize papers based on  efficiency and robustness.

\noindent \textbf{Reducing Training Time and Compute (Data Selection): } A number of recent papers have used submodular functions as \emph{proxy} functions~\cite{wei2014fast,wei2014unsupervised,kirchhoff2014submodularity,kaushal2019learning}. These approaches have been used in several domains including speech recognition~\cite{wei2014submodular,wei2014unsupervised,liu2015svitchboard}, machine translation~\cite{kirchhoff2014submodularity}, computer vision~\cite{kaushal2019learning}, and NLP~\cite{bairi2015summarization}. Another common approach uses coresets. Coresets are weighted subsets of the data, which approximate certain desirable characteristics of the full data (e.g., the loss function)~\cite{feldman2020core}. Coreset algorithms have been used for several problems including $k$-means clustering \cite{har2004coresets}, SVMs~\cite{clarkson2010coresets} and Bayesian inference \cite{campbell2018bayesian}. Coreset algorithms require specialized (and often very different algorithms) depending on the model and problem at hand and have had limited success in deep learning. A very recent coreset algorithm called \textsc{Craig} \cite{mirzasoleiman2019coresets}, which tries to select representative subsets of the training data that closely approximate the full gradient, has shown promise for several machine learning models. 
The resulting subset selection problem becomes an instance of the facility location problem (which is submodular). Another data selection framework, which is very relevant to this work, poses the data selection problem as that of selecting a subset of the training data such that the resulting \emph{model} (trained on the subset) perform well on the full dataset~\cite{wei2015submodularity}. 
\cite{wei2015submodularity} showed that the resulting problem is a submodular optimization problem for the Nearest Neighbor (NN) and Naive Bayes (NB) classifiers. 
The authors empirically showed that these functions worked well for other classifiers, such as logistic regression and deep models~\cite{kaushal2019learning,wei2015submodularity}. 

\noindent \textbf{Reducing Labeling Cost (Active Learning): } Traditionally, active learning techniques like uncertainty sampling (US) and query by committee (QBC) have shown great promise in several domains of machine learning~\cite{settles2009active}. However, with the emergence of batch active learning~\cite{wei2015submodularity,sener2018active},
simple US and QBC approaches do not capture diversity in the batch. Among the approaches to diversified active learning, one of the first approaches was Filtered Active Submodular Selection (FASS)~\cite{wei2015submodularity} that combines the uncertainty sampling method with a submodular data subset selection framework to label a subset of data points to train a classifier. Another related approach~\cite{sener2018active}  defines active learning as a core-set selection problem and has demonstrated that the model learned over the $k$-centers of the dataset is competitive to the one trained over the entire data. 
Very recently, an algorithm called \textsc{Badge}~\cite{ash2020deep} sampled groups of points that have a diverse and higher magnitude of hypothesized gradients to incorporate both predictive uncertainty and sample diversity into every selected batch. 

\noindent \textbf{Robust Learning: } 
A number of approaches have been proposed to address robust learning in the context of noise, distribution shift and class imbalance. Several methods rely on reweighting training examples either by knowledge distillation from auxilliary models~\cite{han2018co, jiang2018mentornet, malach2017decoupling} or by using a clean held out validation set~\cite{ren2018learning,zhang2018generalized}. In particular, our approach bears similarity to the learning to reweight framework~\cite{ren2018learning} wherein the authors try to reweight the training examples using a validation set, and solve the problem using a online meta-learning based approach. 

\noindent \textbf{Submodular Functions: } Since several of the data selection techniques use the notion of submodularity, we briefly introduce submodular functions and optimization.  Let $V = \{1, 2, \cdots, n\}$ denote a ground set of items (for example, in our case, the set of training data points). Set functions are functions $f: 2^V \rightarrow \mathbf{R}$ that operate on subsets of $V$. A set function $f$ is called a submodular function~\cite{fujishige2005submodular} if it satisfies the diminishing returns property: for subsets $S \subseteq T \subseteq V, f(j | S) \triangleq  f(S \cup j) - f(S) \geq f(j | T)$. Several natural combinatorial functions such as facility location, set cover, concave over modular, {\em etc.}, are submodular functions~\cite{iyer2015submodular,iyer2020submodular}. Submodularity is also very appealing because a simple greedy algorithm achieves a $1 - 1/e$ constant factor approximation guarantee~\cite{nemhauser1978analysis} for the problem of maximizing a submodular function subject to a cardinality constraint (which most data selection approaches involve). Moreover, several variants of the greedy algorithm have been proposed which further scale up submodular maximization to almost linear time complexity~\cite{minoux1978accelerated,mirzasoleiman2014lazier,mirzasoleiman2013distributed}. 

\subsection{Our Contribution}
Most prior work discussed above, either study robustness or efficiency, but not both. For example, the data selection approaches such as~\cite{wei2015submodularity,mirzasoleiman2019coresets,shinohara2014submodular} and others focus on approximating either gradients or performance on the \emph{training sets}, and hence would not be suitable for scenarios such as label noise and imbalance. 
On the other hand, the approaches like~\cite{ren2018learning,jiang2018mentornet} and others, focus on robustness but are not necessarily efficient. For example, the approach of \cite{ren2018learning} requires 3x the standard (deep) training cost, to obtain a robust model. 
\model\ is the first framework, to the best of our knowledge, which focuses on both efficiency and robustness. Our work is closely related to the approaches of \cite{wei2015submodularity} and \cite{ren2018learning}. We build upon the work of \cite{wei2015submodularity}, by first generalizing their framework beyond simple classifiers (like nearest neighbor and naive bayes), but with general loss functions. We do this by proposing an iterative algorithm \modelonline\ which does data selection via a meta-learning based approach along with parameter updates. Furthermore, we pose the problem as optimizing the \emph{validation} set performance as opposed to training set performance, thereby encouraging generalization. Next, our approach also bears similarity to \cite{ren2018learning}, except that we need to solve a \emph{discrete} optimization problem instead of a meta-gradient update. Moreover, we do not run our data selection every iteration, thereby ensuring that we are significantly faster than a single training run. 
Finally, we extend our algorithm to the active learning scenario. We demonstrate that our framework is more efficient and accurate compared to existing data selection and active learning algorithms, and secondly, also generalizes well under noisy data, and class imbalance scenarios. In particular, we show that \model{} achieves a \textbf{3x - 6x speedup on a wide range of models and datasets, with very small loss in accuracy}.

\section{Problem Formulation} \label{prob-form}
\textbf{Notation: } Denote ${\mathcal U}$ to be the full training set with instances $\{(x^i, y^i)\}_{i \in {\mathcal U}}$ and ${\mathcal V}$ to be a held-out validation set $\{(x^i, y^i)\}_{i \in {\mathcal V}}$. Define $L(\theta, S) = \sum_{i \in S} L(\theta, x^i, y^i)$ as the loss on a set $S$ of instances. Denote $L_T$ as the training loss, and hence $L_T(\theta, {\mathcal U})$ is the full training loss. Similarly, denote $L_V$ as the validation loss (i.e. $L_V(\theta, {\mathcal V})$ as the loss on the validation set ${\mathcal V}$). In this paper, we study the following problem:
\begin{align}\label{prob-formloss}
    \underset{{S \subseteq {\mathcal U}, |S| \leq k}}{\operatorname{argmin}} L_V(\mbox{argmin}_{\theta} L_T( \theta, S), {\mathcal V})
\end{align}
Equation~\eqref{prob-formloss} tries to select a subset $S$ of the training set ${\mathcal U}$, such that the loss on the set ${\mathcal V}$ is minimized. We can replace the loss functions $L_V$ and $L_T$ with $\log$-likelihood functions $LL_V$ and $LL_T$ in which case, the $\mbox{argmin}$ becomes $\mbox{argmax}$: 
\begin{align}\label{prob-formll}
    \underset{{S \subseteq {\mathcal U}, |S| \leq k}}{\operatorname{argmax}} LL_V(\mbox{argmax}_{\theta} LL_T( \theta, S), {\mathcal V})
\end{align}
Finally, we point out that we can replace $L_V$ (or $LL_V$) with the training loss $L_T$ (or log-likelihood $LL_T$), in which case we get the problem studied in~\cite{wei2015submodularity}. The authors in~\cite{wei2015submodularity} only consider simple models such as nearest neighbor and naive bayes classifiers. 

\noindent \textbf{Special Cases: } We start with the naive bayes and nearest neighbor cases, which have already been studied in~\cite{wei2015submodularity}. We provide a simple extension here to consider a validation set instead of the training set. First consider the naive bayes model. Let $m_{x_j,y}(S) = \sum_{i \in S} 1[x^i_j = x_j \wedge y_j = y]$ and
$m_y(S) = \sum_{i \in S} 1[y_j = y]$. Also, denote ${\mathcal V}^y \subseteq {\mathcal V}$ as a set of the validation instances with label $y$. Furthermore, define $w(i, j) = d - ||x^i - x^j||^2_2$, where $d = \max_{i, j} ||x^i - x^j||^2_2$. We now define two submodular functions. The first is the naive-bayes submodular function $f_{NB}^v (S) = \sum_{j=1:d} \sum_{x_j \in \mathcal{X}} \sum_{y \in \mathcal{Y}} m_{x_j,y}({\mathcal V}) \log m_{x_j,y}(S)$, and the second is the nearest-neighbor submodular function: $f^{NN}_V (S) = \sum_{y \in \mathcal{Y}} \sum_{i \in {\mathcal V}^y}\max_{s \in S\cap {\mathcal V}^{y^i}} w(i,s)$.

The following Lemma analyzes Problem~\eqref{prob-formll} in the case of naive bayes and nearest neighbor classifiers.
\begin{lemma}
Maximizing equation~\eqref{prob-formll} in the context of naive bayes and nearest neighbor classifiers is equivalent to optimizing $f^{NB}_V(S)$ and $f^{NN}_V (S)$ under the constraint that \(|S \cap {\mathcal V}^y| = k \frac{|{\mathcal V}^y|}{|{\mathcal V}|}\) with \(|S| = k\), which is essentially a partition matroid constraint.
\end{lemma}
This result is proved in the Appendix, and follows a very similar proof technique from~\cite{wei2015submodularity}. However, in order to achieve this formulation, we make a natural assumption that the  distribution over class labels in
\(S\) is same as that of \(U\). Furthermore, the Lemma above implied that solving equation~\eqref{prob-formll} is basically a form of submodular maximization, for the NB and NN classifiers. In the interest of space, we defer the analysis of Linear Regression (LR) and Gaussian Naive Bayes (GNB) to the appendix. Both these models enable closed form for the inner problem, and the resulting  problems are closely related to submodularity.

\begin{figure}[!t]
    \centering
    \includegraphics[width=0.48\textwidth, height=5.5cm]{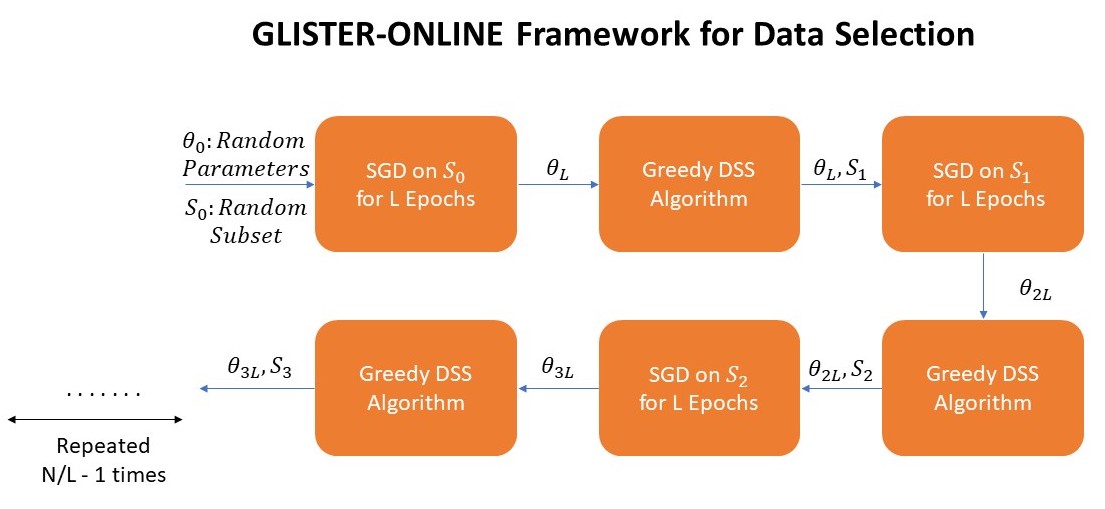}
    \caption{ Main flowchart of the \modelonline\ framework for Data Selection.}
    \label{fig:framework}
\end{figure}

\section{\modelonline\ Framework}
In this section, we present \modelonline, which performs data selection jointly with the parameter learning. This allows us to handle arbitrary loss functions $L_V$ and $L_T$. A careful inspection of equation~\eqref{prob-formll} reveals that it is a nested bi-level optimization problem:


\begin{equation}
\label{nested-equation}
    \overbrace{\underset{{S \subseteq {\mathcal U}, |S| \leq k}}{\operatorname{argmax\hspace{0.7mm}}} LL_V(\underbrace{\underset{\theta}{\operatorname{argmax\hspace{0.7mm}}} LL_T( \theta, S)}_{inner-level}, {\mathcal V})}^{outer-level}
\end{equation}

The outer layer, tries to select a subset from the training set ${\mathcal U}$, such that the model trained on the subset will have the best log-likelihood $LL_{V}$ on the validation set ${\mathcal V}$. Whereas in the inner layer, we optimize the model parameters by maximizing training log-likelihood $LL_{T}$ on the subset selected $S$. Due to the fact that equation~\eqref{nested-equation} is a bi-level optimization, it is expensive and impractical to solve for general loss functions. This is because in most cases, the inner optimization problem cannot be solved in closed form. Hence we need to make approximations to solve the optimization problem efficiently.

\begin{algorithm}[!tb]
\caption{\modelonline\ Algorithm}
\label{alg:algorithm1}
\begin{algorithmic}[1]
\REQUIRE Training data ${\mathcal U}$, Validation data ${\mathcal V}$, Initial subset  $S^{(0)}$ of size=$k$, $\theta^{(0)}$ model parameters initialization
\REQUIRE $\eta$: learning rate. $T=$ total epochs, $L=$ epoch interval for selection, , $r=$ No of Taylor approximations, $\lambda= $ Regularization coefficient
\FORALL {epoch $t$ in $T$}
        \IF{$t \mbox{ mod } L == 0$}
            \STATE $S^{(t)} = \operatorname{GreedyDSS}(\mathcal{U},\mathcal{V},\theta^{(t-1)}, \eta, k, r, \lambda)$ 
        \ELSE
            \STATE $S^{(t)} = S^{(t-1)}$
        \ENDIF
        \STATE Perform one epoch of batch SGD to  update $\theta_t$ on $LL_T$ and using the data subset $S^{(t)}$.
\ENDFOR
\STATE Output final subset $S^{(T)}$ and parameters $\theta^{(T)}$
\end{algorithmic}
\end{algorithm}

\begin{figure*}[!htbp]
\centering
    \begin{subfigure}[b]{0.28\textwidth}
        \centering
        \includegraphics[width=\textwidth, height=2.7cm]{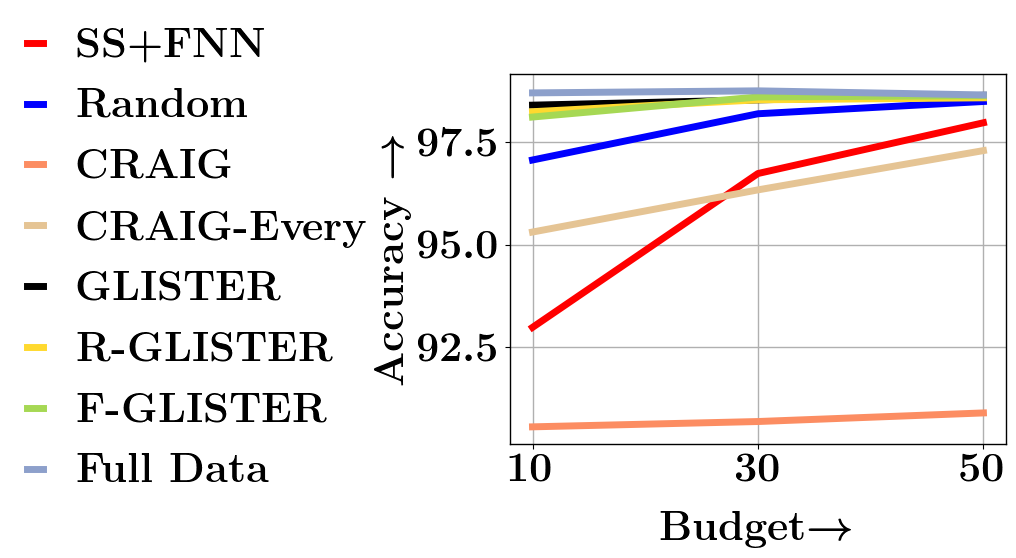}
        \subcaption{}
    \end{subfigure}
    \begin{subfigure}[b]{0.18\textwidth}
        \centering
        \includegraphics[trim=270 0 0 0, clip,width=\textwidth, height=2.7cm]{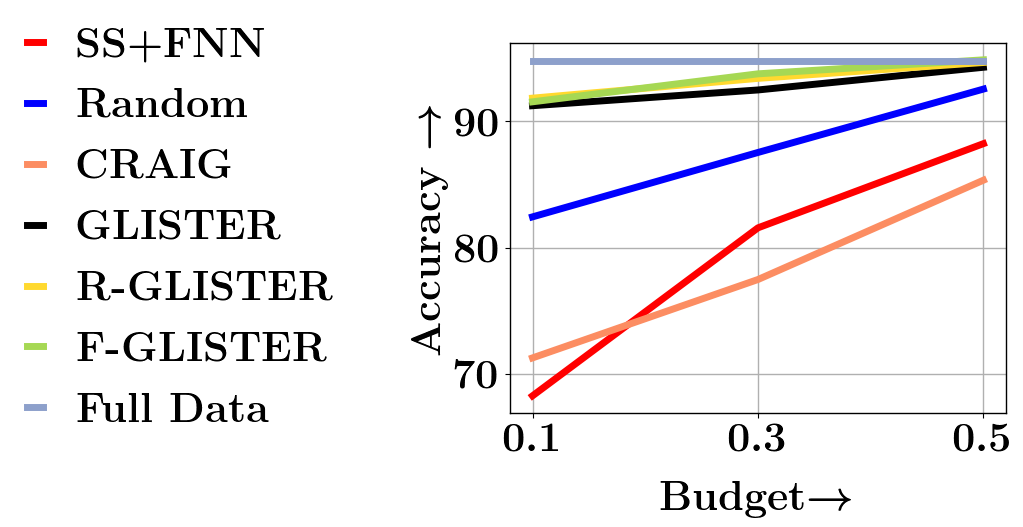}
        \subcaption{}
    \end{subfigure}
    \centering
    \begin{subfigure}[b]{0.28\textwidth}
        \centering
        \includegraphics[width=\textwidth, height=2.7cm]{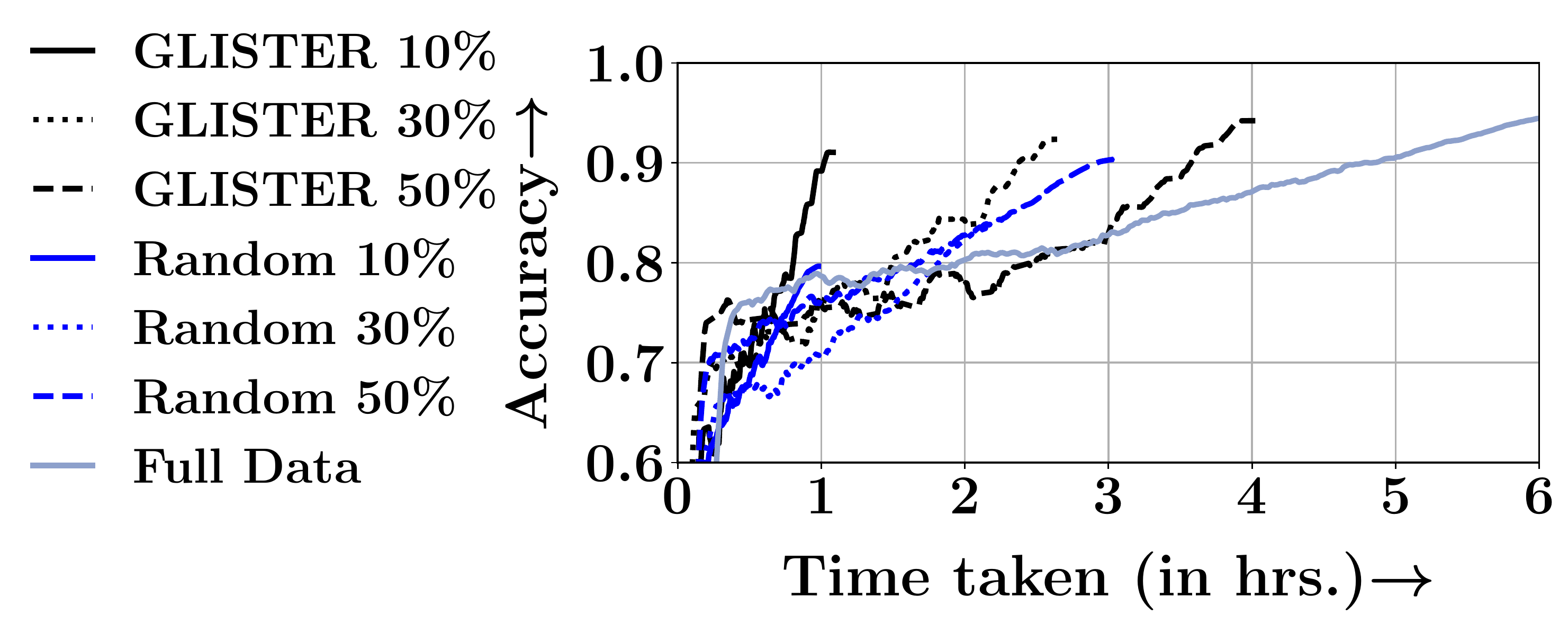}
        \subcaption{}
    \end{subfigure}
    \begin{subfigure}[b]{0.24\textwidth}
        \centering
        \includegraphics[width=\textwidth, height=2.7cm]{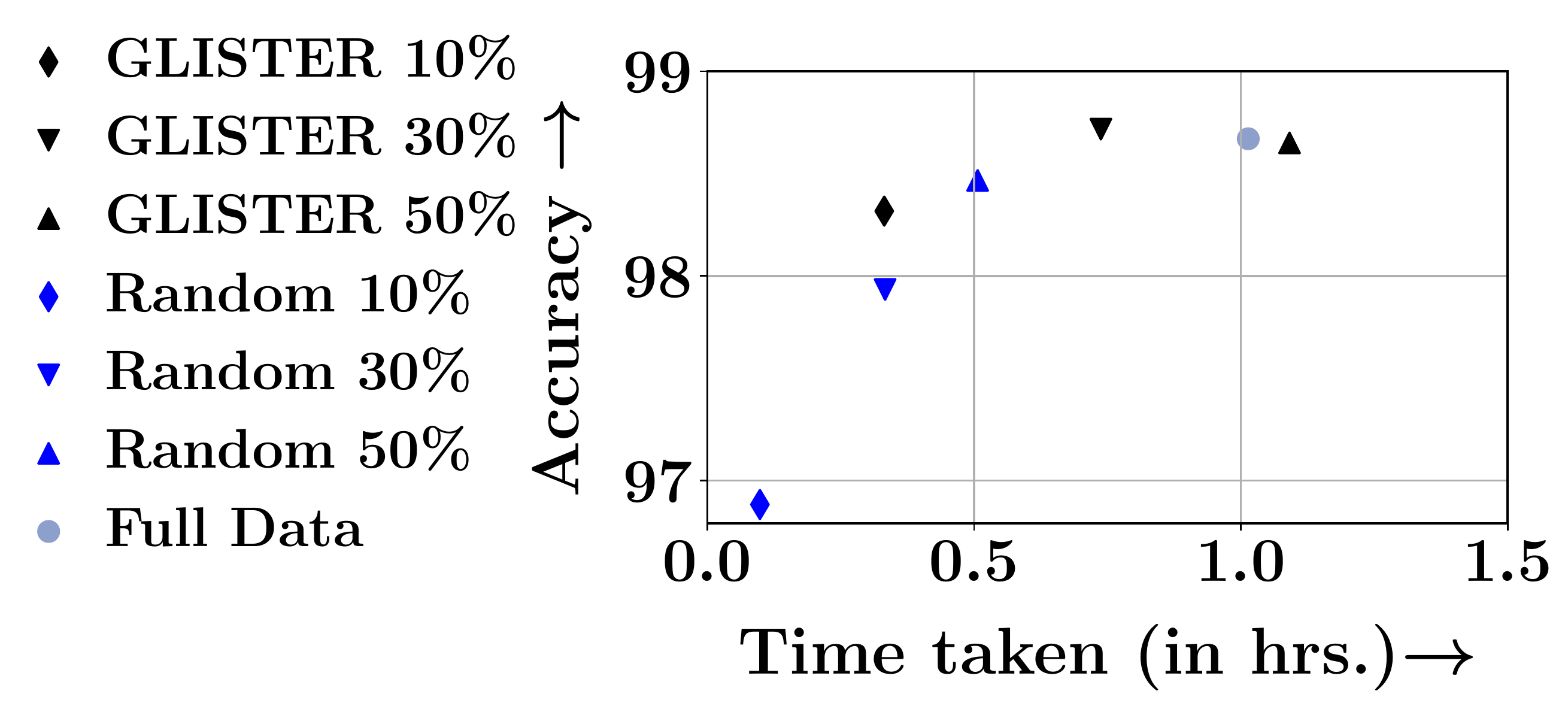}
        \subcaption{}
    \end{subfigure}
    \begin{subfigure}[b]{0.30\textwidth}
        \centering
        \includegraphics[width=\textwidth, height=2.7cm]{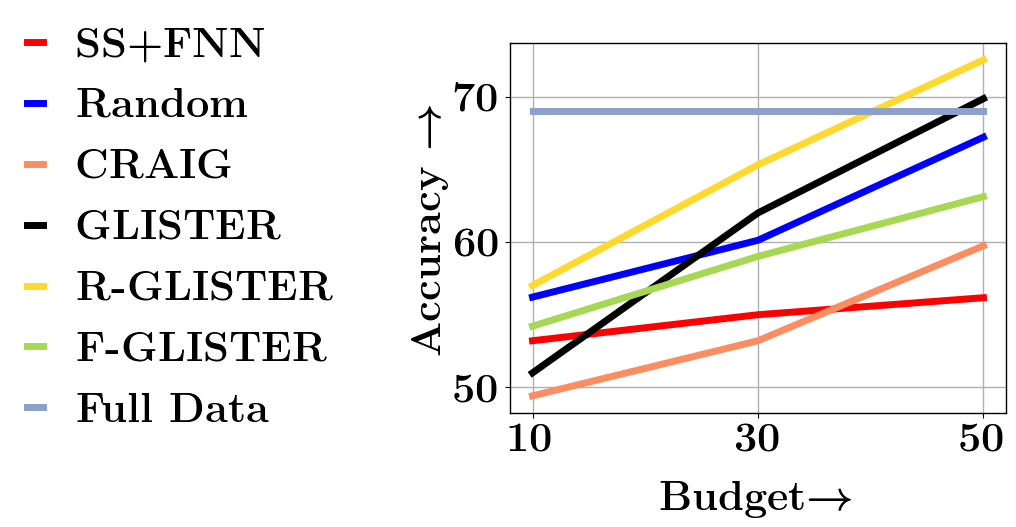}
        \subcaption{}
    \end{subfigure}
    \begin{subfigure}[b]{0.22\textwidth}
        \centering
        \includegraphics[trim=270 0 0 0, clip,width=\textwidth, height=2.7cm]{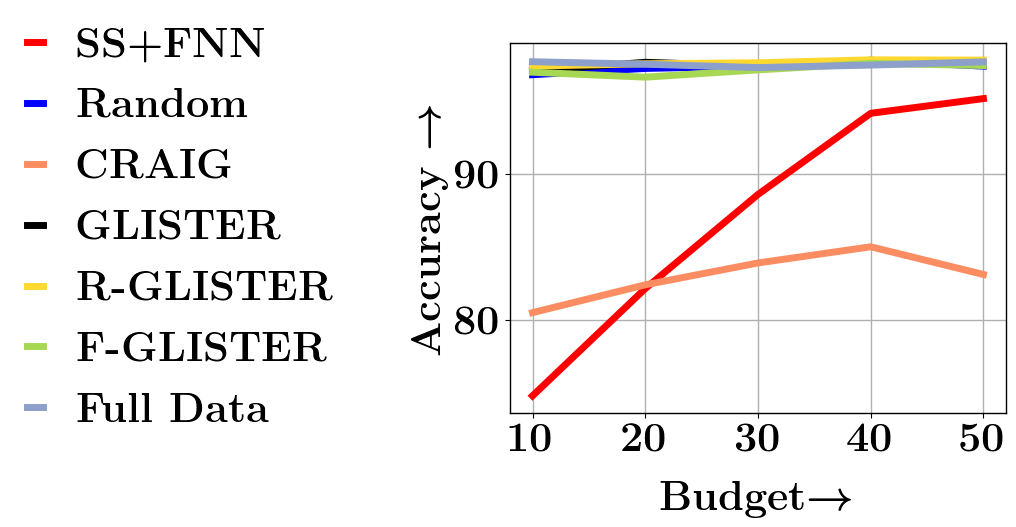}
        \subcaption{}
    \end{subfigure}
    \begin{subfigure}[b]{0.22\textwidth}
        \centering
        \includegraphics[trim=270 0 0 0, clip,width=\textwidth, height=2.7cm]{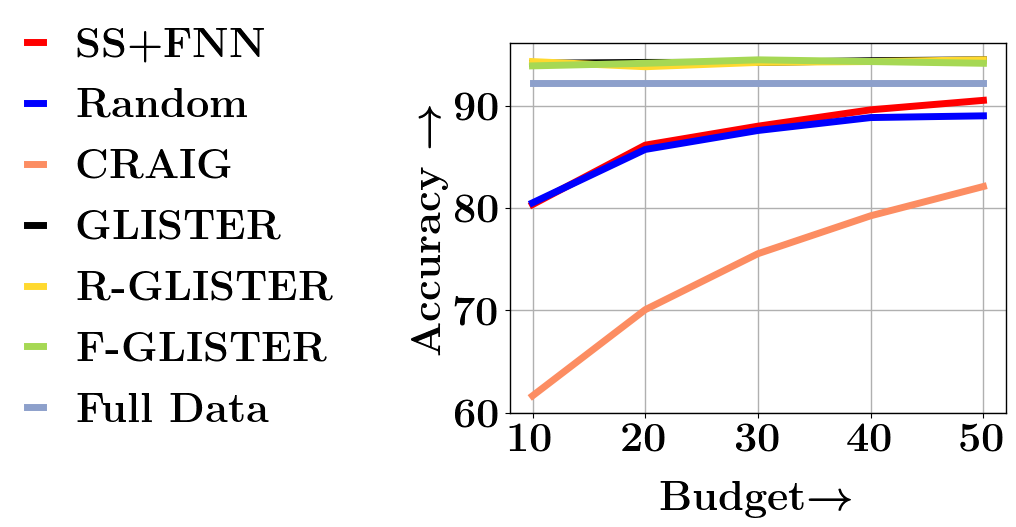}
        \subcaption{}
    \end{subfigure}
    \begin{subfigure}[b]{0.22\textwidth}
        \centering
        \includegraphics[trim=270 0 0 0, clip,width=\textwidth, height=2.7cm]{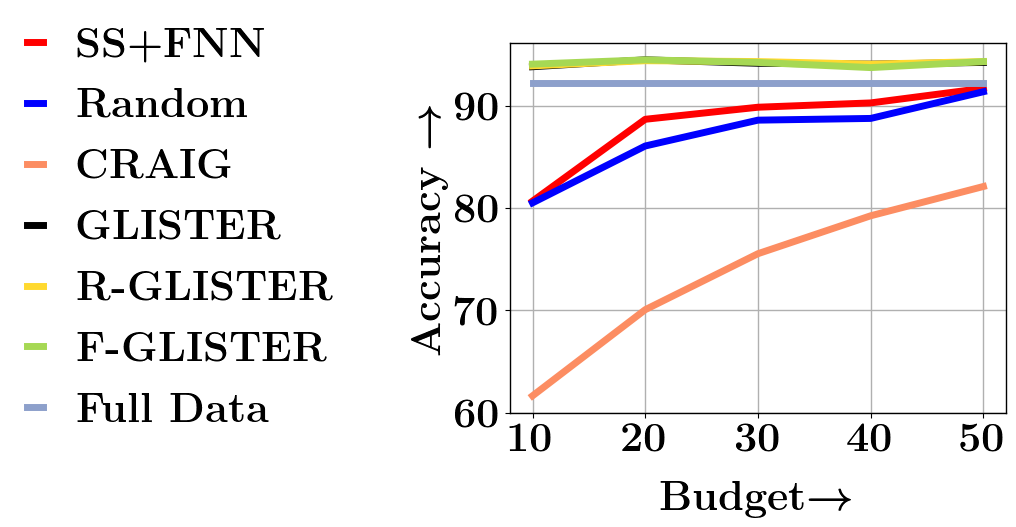}
        \subcaption{}
    \end{subfigure}
    \caption{Top Row: Data Selection for Efficient Learning. 
    (a) MNIST Accuracy vs Budgets, (b) CIFAR-10 Accuracy vs Budget , (c) CIFAR 10 Convergence plot, and (d) MNIST Accuracy vs Total time taken. Bottom Row: Data Selection in Class Imbalance and Noise: Accuracy vs Budget (e) CIFAR-10 (Class Imb), (f) MNIST (Class Imb), (g) DNA (Class Imb), and (d) DNA (Noise).}
    \label{fig:dss_general_imb_experiments}
\end{figure*}

\subsubsection{Online Meta Approximation Algorithm}
Our first approximation is that instead of solving the inner optimization problem entirely, we optimize it by iteratively doing a meta-approximation, which takes a single step towards the training subset log-likelihood $LL_{T}$ ascent direction. This algorithm is iterative, in that it proceeds by simultaneously updating the model parameters and selecting subsets. Figure~\ref{fig:framework} gives a flowchart of \modelonline{}. Note that instead of performing data selection every epoch, we perform data selection every $L$ epochs, for computational reasons. We will study the tradeoffs associated with $L$ in our experiments.

\modelonline{} proceeds as follows. We update the model parameters $\theta_t$ on a subset obtained in the last subset selection round. We perform subset selection only every $L$ epochs, which we do as follows. At training time step $t$, if we take one gradient step on a subset $S$, we achieve: $\theta^{t+1}(S) = \theta^{t} + \eta \nabla_{\theta} LL_T(\theta^{t}, S)$. We can then plug this \emph{approximation} into equation~\eqref{nested-equation} and obtain the following discrete optimization problem. Define $G_{\theta}(S) = LL_V(\theta + \eta \nabla_{\theta} LL_T(\theta, S), {\mathcal V})$ below: 
\begin{align} \label{glister-disc}
    S^{t+1} &= \underset{{S \subseteq {\mathcal U}, |S| \leq k}}{\operatorname{argmax\hspace{0.7mm}}} G_{\theta^t}(S)
\end{align}
Next, we show that the optimization problem (equation~\eqref{glister-disc}) is NP hard.
\begin{lemma}
\label{np-hard}
There exists log-likelihood functions $LL_V$ and $LL_T$ and training and validation datasets, such that equation~\eqref{glister-disc} is NP hard.
\end{lemma}
The proof of this result is in the supplementary material. While, the problem is NP hard in general, we show that for many important log-likelihood functions such as the negative cross entropy, negative logistic loss, and others, $G_{\theta^t}(S)$ is submodular in $S$ for a given $\theta^t$.

\begin{theorem}
\label{theorem-1}
If the validation set log-likelihood function $LL_V$ is either the negative logistic loss, the negative squared loss, negative hinge loss, or the negative perceptron loss, the optimization problem in equation~\eqref{glister-disc} is an instance of cardinality constrained submodular maximization. When $LL_V$ is the negative cross-entropy loss, the optimization problem in equation~\eqref{glister-disc} is an instance of cardinality constrained weakly submodular maximization. 
\end{theorem}
The proof of this result, along with the exact forms of the (weakly) submodular functions are in the supplementary material. Except for negative squared loss, the submodular functions for all other losses (including cross-entropy) are monotone, and hence the lazy greedy~\cite{minoux1978accelerated} or stochastic greedy~\cite{mirzasoleiman2014lazier} give $1-1/e$ approximation guarantees. For the case of the negative squared loss, the randomized greedy algorithm achieves a $1/e$ approximation guarantee~\cite{buchbinder2014submodular}. The lazy greedy algorithm in practice is amortized linear-time complexity, but in the worst case, can have a complexity $O(nk)$. On the other hand, the stochastic greedy algorithm is linear time, i.e. it obtains a $1 - 1/e - \epsilon$ approximation in $O(n\log 1/\epsilon)$ iterations.

Before moving forward, we analyze the computational complexity of \modelonline{}. Denote $m = |\mathcal V$ (validation set size), and $n = |\mathcal U|$ (training set size). Furthermore, let $F$ be the complexity of a forward pass and $B$ the complexity of backward pass. Denote $T$ as the number of epochs. The complexity of full training with SGD is $O(nTB)$. Using the naive (or lazy) greedy algorithm, the worst case complexity is $O(nkmFT/L + kTB)$. With stochastic greedy, we get a slightly improved complexity of $O(nmFT/L \log 1/\epsilon + kTB)$. Since $B \approx 2F$, and $m$ is typically a fraction of $n$ (like 5-10\% of $n$), and $L$ is a constant (e.g. $L = 20$), the complexity of subset selection (i.e. the first part) can be significantly more than the complexity of training (for large values of $n$), thereby defeating the purpose of data selection. Below, we study a number of approximations which will make the subset selection more efficient, while not sacrificing on accuracy.

\subsubsection{Approximations:} We start with a simple approximation of just using the \textbf{Last layer approximation} of a deep model, while computing equation~\eqref{glister-disc}. Note that this can be done in closed form for many of the loss functions described in Theorem 1. Denote $f$ as the complexity of computing the function on the last layer (which is much lesser than $F$), the complexity of stochastic greedy is reduced to $O(nmfT/L \log 1/\epsilon  + kTB)$. The reason for this is that \modelonline{} tries to maximize $LL_V(\theta + \eta \sum_{j \in S} \nabla LL_T(\theta, j), \mathcal V)$, and the complexity of evaluating $LL_V$ is $O(mf)$. Multiplying this by the complexity of the stochastic or naive greedy, we get the final complexity. While this is better than earlier, it can still be slow since there is a factor $nm$ being multiplied in the subset selection time. The second approximation we make is the \textbf{Taylor series approximation}, which computes an approximation $\hat{G}_{\theta}(S \cup e)$ based on the Taylor-series approximation. In particular, given a subset $S$, define $\theta^S = \theta + \eta \sum_{j \in S} \nabla LL_T(\theta, j)$. The Taylor-series approximation $\hat{G}_{\theta}(S \cup e) = LL_V(\theta^S) + \eta \nabla_{\theta} LL_T(\theta, e)^T LL_V(\theta^S, \mathcal V)$. Note that $LL_T(\theta, e)$ can be precomputed before the (stochastic) greedy algorithm is run, and similarly $LL_V(\theta^S, \mathcal V)$ just needs to be computed once before picking the best $e \notin S$ to add. With the Taylor series approximation, the complexity reduces to $O(k[m+n]fT/L + kTB)$ with the naive-greedy and $O([km + n\log 1/\epsilon]fT/L + kTB)$ with the stochastic greedy. Comparing to without using the Taylor series approximation, we get a speedup of $O(m)$ for naive-greedy and $O(n\log 1/\epsilon /k)$ for the stochastic greedy, which can be significant when $k$ is much smaller than $n$ (e.g., 10\% of $n$). With the Taylor series approximation, we find that the time for subset selection (for deep models) is often comparable in complexity to one epoch of full training, but since we are doing subset selection only every $L$ epochs, we can still get a speedup roughly equal to $n/k + 1/L$. However, for shallow networks or 2 layer neural networks, the subset selection time can still be orders of magnitude slower than full training. In this case, we do one final approximation, which we call the \textbf{$r$-Greedy Taylor Approximation}. In this case, we re-compute the validation log-likelihood only $r$ times (instead of $k$). In other words, we use the \emph{stale} likelihoods for $k/r$ greedy steps. Since we are using the same likelihood function, this becomes a simple modular optimization problem where we need to pick the top $k/r$ values. While in principle, this approximation can be used in conjunction with stochastic greedy, we find that the accuracy degradation is considerable, mainly because stochastic greedy picks the best item greedily from $O(n/k \log 1/\epsilon)$ random data instances, which can yield poor sets if a large number of items are selected every round of greedy (which is what happens in $r$-greedy). Rather, we use this in conjunction with naive-greedy. The complexity of the $r$-taylor approximation with naive-greedy algorithm is $O(r[m+n]fT/L + kTB)$. For smaller models (like one or two layer neural network models), we set $r = 0.03k$ (we perform ablation study on $r$ in our experiments), thereby achieving 30x speedup to just the Taylor-series approximation. For deep models, since the complexity of the gradient descent increases (i.e. $B$ is high), we can use larger values of $r$, and typically set $r \approx k$. As we demonstrate in our experiments, even after the approximations presented above, we significantly outperform other baselines (including \textsc{Craig} and Random), and is comparable to full training while being much faster, on a wide variety of datasets and models.

\subsubsection{Regularization with Other Functions}
Note that since the optimization problem in equation~\eqref{glister-disc} is an instance of submodular optimization, we can also \emph{regularize} this with another data-selection approach. This can be particularly useful, if say, the validation dataset is small and we do not want to overfit to the validation loss.
The regularized objective function is ($\lambda$ is a tradeoff parameter):
\begin{align} \label{reg-objective}
    S^{t+1} = \mbox{argmax}_{S \subseteq \mathcal U, |S| \leq k} G_{\theta^t}(S) + \lambda R(S).
\end{align}
We consider two specific kinds of regularization functions $R(S)$. The first is the supervised facility location, which is basically NN-submodular function on the training set features~\cite{wei2015submodularity}, and the second is a random function. The random function can be thought of as a small perturbation to the data selection objective. 

\subsubsection{Implementation Aspects}
The detailed pseudo-code of the final algorithm is in Algorithm \ref{alg:algorithm1}. \emph{GreedyDSS} in Algorithm~\ref{alg:algorithm1} refers to the greedy algorithms and approximations discussed above to solve equation~\eqref{reg-objective}. The parameters of GreedyDSS are: a) training set $\mathcal U$, validation set: $\mathcal V$, current parameters $\theta^{(t)}$, learning rate $\eta$, budget $k$, the parameter $r$ governing the number of times we do taylor-series approximation, and finally the regularization coefficient $\lambda$. In the interest of space, we defer the detailed algorithm to the supplementary material. We use the PyTorch \cite{paszke2017automatic} framework to implement our algorithms. To summarize the implementation aspects, the main hyper-parameters which govern the tradeoff between accuracy and efficiency are $r, L$, the regularization function $R$, coefficient $\lambda$ and the choice of the greedy algorithm. For all our experiments, we set $L = 20$. For our deep models experiments (i.e. more than 2-3 layers), we use just the Taylor-approximation with $r = k$ and stochastic greedy, while for shallow models, we use $r \approx 0.03k$ with the naive greedy. We perform ablation studies in Section~\ref{sec:experiments} to understand the effect of these parameters in experiments, and in particular, the choices of $r$ and $L$. We do not significantly tune $\lambda$ in the regularized versions and just set in a way so both components (i.e. the \textsc{Glister} loss and regularizer) have roughly equal contributions. See the supplementary material for more details of the hyper-parameters used.
\begin{figure*}
    \begin{subfigure}[b]{0.3\textwidth}
        \centering
        \includegraphics[width=\textwidth, height=2.7cm]{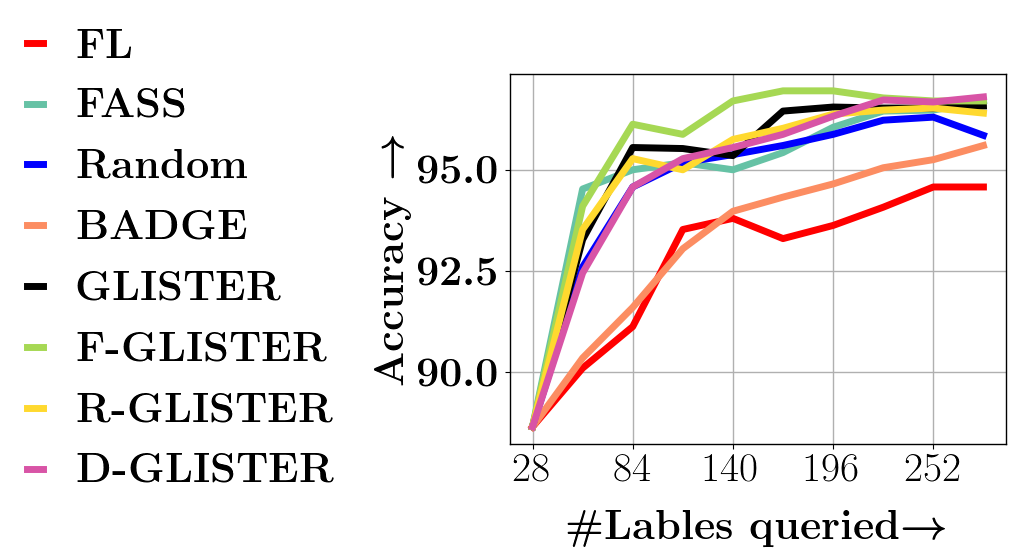}
        \subcaption{SVM-Guide}
    \end{subfigure}
    \begin{subfigure}[b]{0.20\textwidth}
        \centering
        \includegraphics[trim=270 0 0 0, clip,width=\textwidth, height=2.7cm]{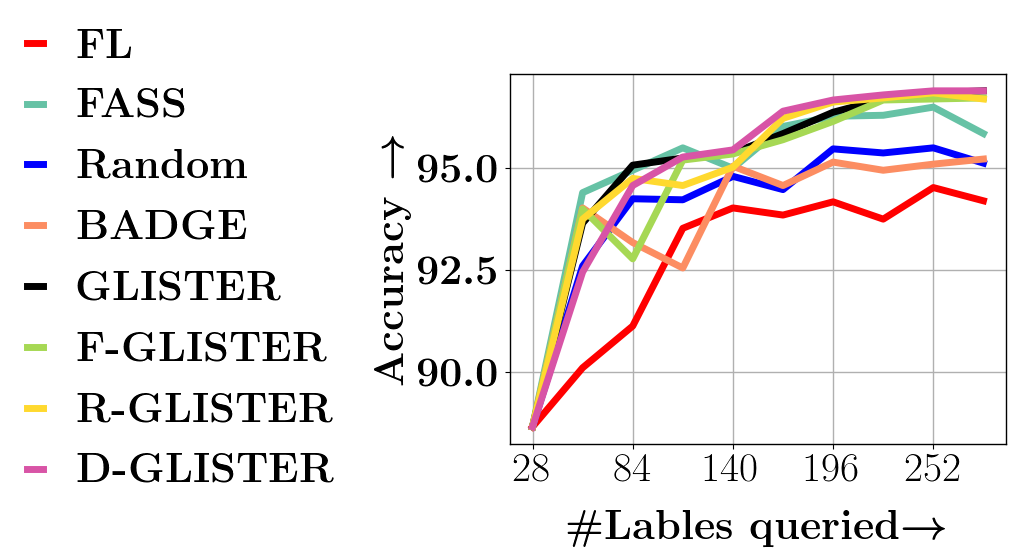}
        \subcaption{SVM Guide (Imbalance)}
    \end{subfigure}
        \begin{subfigure}[b]{0.20\textwidth}
        \centering
        \includegraphics[trim=270 0 0 0, clip,width=\textwidth, height=2.7cm]{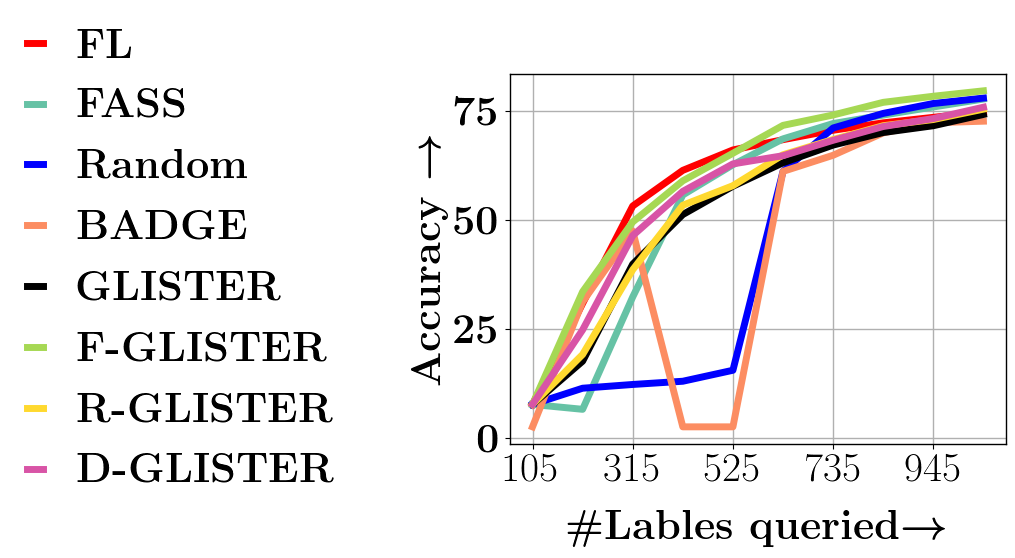}
        \subcaption{Letter}
    \end{subfigure}
    \centering
    \begin{subfigure}[b]{0.20\textwidth}
        \centering
       \includegraphics[trim=270 0 0 0, clip,width=\textwidth, height=2.7cm]{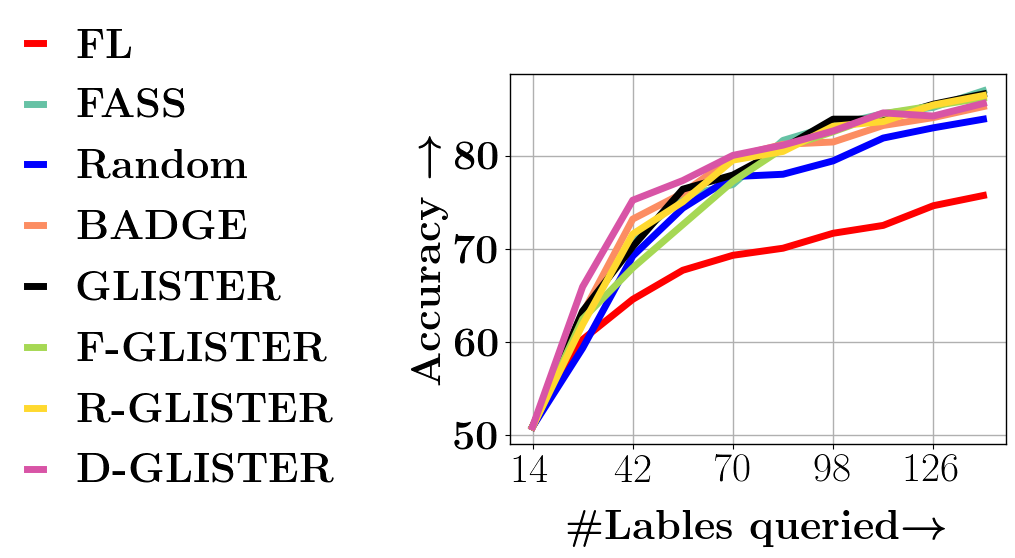}
        \subcaption{DNA Datasets}
    \end{subfigure}
    \caption{Active Learning Results. In each case, we see that versions of \textsc{Glister} out-perform existing techniques.}
    \label{fig:noise_AL_experiments}
\end{figure*}

\begin{figure}
    \begin{subfigure}[b]{0.22\textwidth}
    \centering
    \includegraphics[width=\textwidth, height=2.5cm]{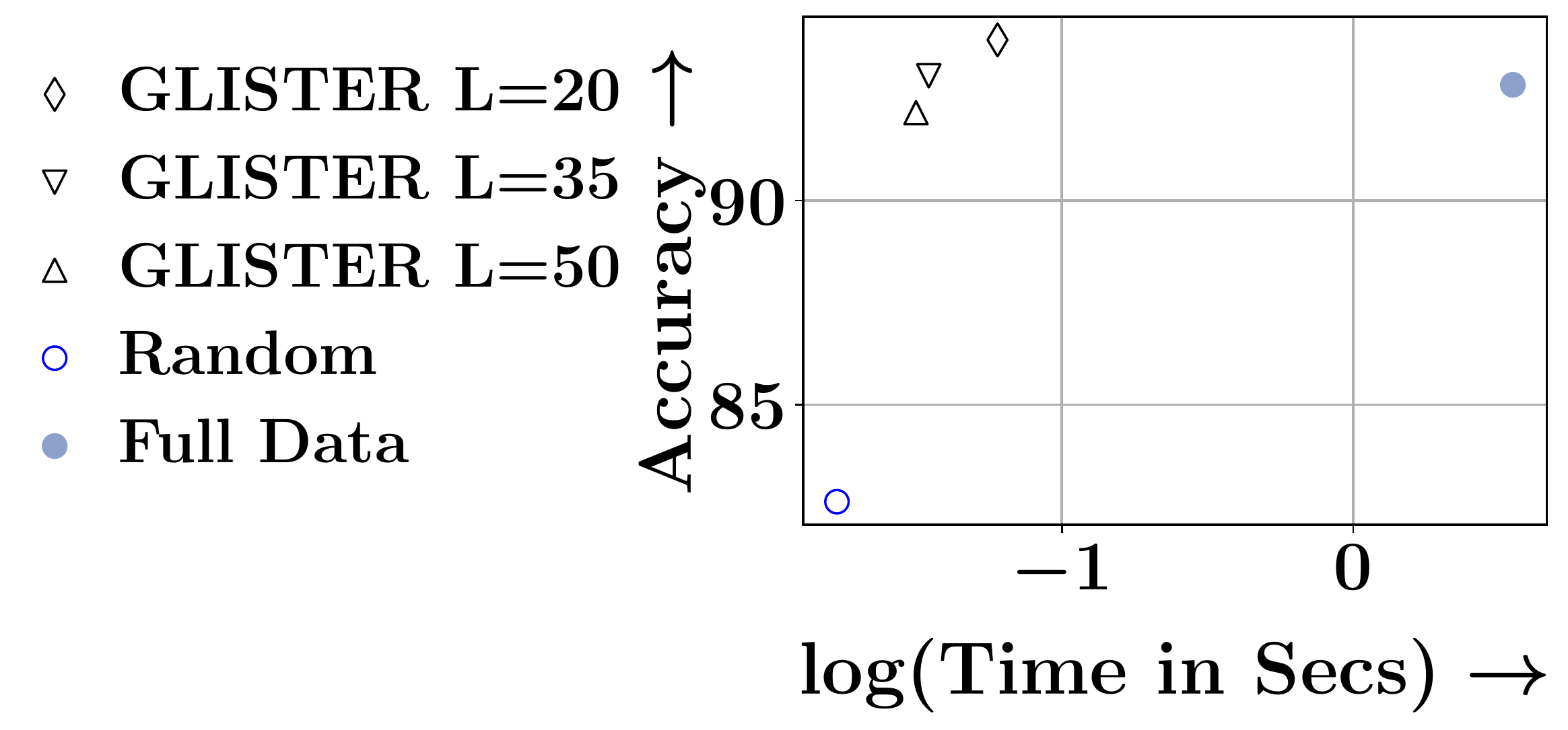}
    \subcaption{}
    \end{subfigure}
    \begin{subfigure}[b]{0.24\textwidth}
    \centering
    \includegraphics[width=\textwidth, height=2.7cm]{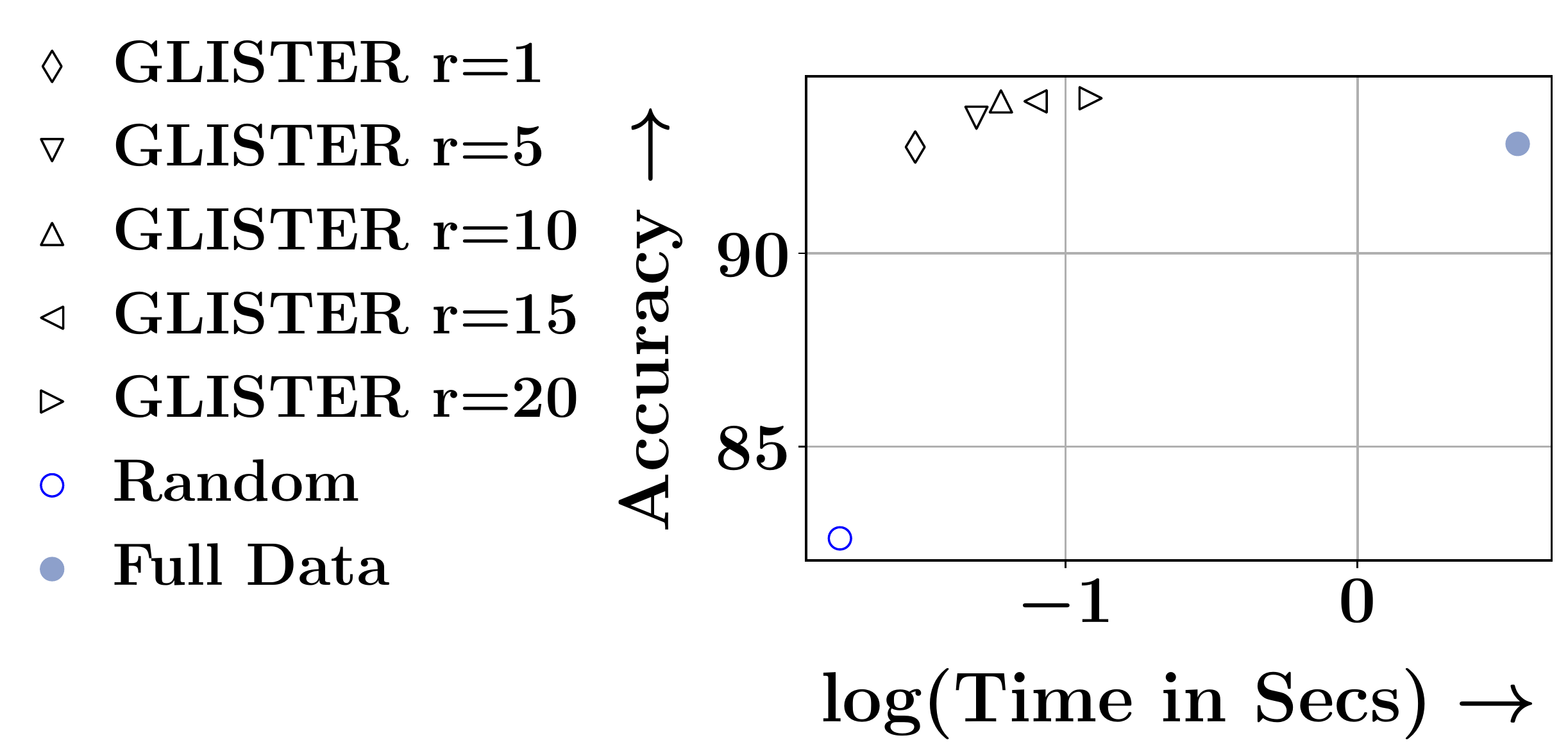}
    \subcaption{}
    \end{subfigure}
    \caption{Ablation study comparing the effect of $L$ and $r$ on DNA dataset. In our experiments, we choose $L = 20$, and for our smaller datasets, we set $r = 0.03k$. The x-axis is log scale to the base $e$.}
    \label{run-time}
\end{figure}

\subsubsection{Convergence Analysis}
In this section, we study conditions under which \modelonline\ reduces the objective value, and the conditions for convergence. To do this, we first define certain properties of the validation loss. A function $f(x) : \mathcal{R}^{d} \rightarrow \mathcal{R}$ is said to be Lipschitz-smooth with constant $\mathcal L$ if $\|\nabla f(x) - \nabla f(y)\| \leq \mathcal{L} \|x - y\|, \forall x, y \in \mathcal{R}^{d}$. Next, we say that a function $f : \mathcal{R}^{d} \rightarrow \mathcal{R}$ has $\sigma$-bounded gradients if $\|\nabla f(x)\| \leq \sigma $ for all $x \in \mathcal{R}^d$

The following result studies conditions under which the validation loss reduces with every training epoch $l$.
\begin{theorem}
\label{theorem-2}
Suppose the validation loss function $L_V$ is Lipschitz smooth with constant $\mathcal{L}$, and the gradients of training and validation losses are $\sigma_T$ and $\sigma_V$ bounded respectively. Then the validation loss always monotonically decreases with every training epoch $l$, i.e. $L_V(\theta_{l+1}) \leq L_V(\theta_{l})$  if it satisfies the condition that $\nabla_{\theta}L_V(\theta_{l}, {\mathcal V})^{T} \nabla_{\theta}L_T(\theta_{l}, S) \geq 0$ for $0 \leq l \leq T$ and the learning rate $\alpha \leq \min_l \frac{2\|\nabla_{\theta}L_V(\theta_{l}, {\mathcal V})\|\cos(\Theta_l)}{\mathcal{L}\sigma_T}$ where $\Theta_l$ is the angle between $\nabla_{\theta}L_V(\theta_{l}, {\mathcal V})$ and $\nabla_{\theta}L_T(\theta_{l}, S)$. 
\end{theorem}
The condition basically requires that for the subset selected $S_i$, the gradient on the training subset loss $L_T(\theta_l, S_i)$ is in the same direction as the gradient on the validation loss $L_V(\theta_l, {\mathcal V})$ at every epoch $l$.  Note that in our Taylor-series approximation, we anyways select  subsets $S_i$ such that the dot product between gradients of subset training loss and validation loss is maximized. So, as long as our training data have some instances that are similar to the validation dataset, our model selected subset $S_i$ should intuitively satisfy the condition mentioned in Theorem \ref{theorem-2}. 
We end this section by providing a convergence result.

The following theorem  shows that under certain conditions, \modelonline\ converges to the optimizer of the validation loss in $\mathcal{O}(1/{\epsilon}^2)$ epochs.
\begin{theorem}
\label{theorem-3}
Assume that the validation and subset training losses satisfy the conditions that $\nabla_{\theta}L_V(\theta_{l}, {\mathcal V})^{T} \nabla_{\theta}L_T(\theta_{l}, S) \geq 0$ for $0 \leq l \leq T$, and for all the subsets encountered during the training. Also, assume that $\delta_{\min} = \min_l \frac{\nabla L_T(\theta_l)}{\sigma_G}$. Then, the following convergence result holds:
\small{\begin{align*}
   &\min_l L_V(\theta_l) - L_V(\theta^{*}) \leq  \frac{R\sigma_T}{\delta_{\min}\sqrt{T}} + \frac{R\sigma_T\sum_{l=0}^{T}\sqrt{1 -\cos{\Theta_{l}}}}{T\delta_{\min}} \\
    &\text{where\hspace{0.5cm}} \cos{\Theta_l} = \frac{\nabla_{\theta}L_T(\theta_l)^T \cdot \nabla_{\theta}L_V(\theta_l)}{\left\Vert\nabla_{\theta}L_T(\theta_l)\right\Vert\left\Vert\nabla_{\theta}L_V(\theta_l)\right\Vert}\text{\hspace{0.5cm}} 
\end{align*}}
\end{theorem}

Since our Taylor approximation chooses a subset that maximizes the dot-product between the gradients of the training loss and the validation loss, we expect that the angle between the subset training gradient and validation loss gradients be close to 0 i.e., $\cos{\theta_l} \approx 1$. Finally, note that $\delta_{\min}$ needs to be greater than zero, which is also reasonable, since having close to zero gradients on the subset gradients would imply overfitting to the training loss. In \modelonline\, we only train on the subset for $L$ epochs, ensuring the training gradients on the subset do not go to zero.



\section{\modelactive\ Framework} 
In this section, we extend \textsc{Glister} to the active learning setting. We propose \modelactive\ for the mini-batch adaptive active learning where we select a batch of \textit{B} samples to be labeled for \textit{T} rounds. This method is adaptive because samples selected in the current round are affected by the previously selected points as the model gets updated. The goal here is to select a subset of size \textit{B} from the pool of unlabeled instances such that the subset selected has as much information as possible to help the model come up with an appropriate decision boundary. \textsc{Glister-Active} is very similar to \textsc{Glister-Online} except for three critical differences. First, the data-selection in Line 6 of Algorithm~\ref{alg:algorithm1} is only on the unlabeled instances. Second, we use the hypothesized labels instead of the true labels in the greedy Taylor optimization (since we do not have the true labels). The usage of hypothesized labels({\em, i.e.}, predictions from the current model) is very similar to existing active learning approaches like \textsc{Badge} and \textsc{Fass}. Thirdly, instead of selecting $k$ examples every time and running only on that subset, \textsc{Glister-Active} selects a batch of $B$ instances over the unlabeled examples and adds it to the current set of labeled examples (after obtaining the labels). Similar to \textsc{Glister-Online}, we consider both the unregularized and regularized data selection objectives. In the interest of space, we defer the algorithm and other details to the supplementary material.

\section{Experimental Results}\label{sec:experiments}
Our experimental section aims to showcase the stability and efficiency of \modelonline\ and \modelactive\ on a number of real world datasets and experimental settings. We try to address the following questions through our experiments: 1) How does \modelonline\ tradeoff accuracy and efficiency, compared to the model trained on the full dataset and other data selection techniques? 2) How does \modelonline\ work in the presence of class imbalance and noisy labels? 3) How well does \modelonline\ scale to large deep learning settings? and 4) How does \modelactive\ compare to other active learning algorithms?

\noindent \textbf{Baselines in each setting:} We compare the following baselines to our \textsc{Glister} framework. We start with data selection for efficient training. \noindent \textbf{1. Random: } Just randomly select $k$ (budget size) training examples. \textbf{2. CRAIG:} We compare against CRAIG~\cite{mirzasoleiman2019coresets} which tries to approximate the gradients of the full training sets. \textbf{3. SS + FNN:} This is the KNN submodular function from \cite{wei2015submodularity}, but using the training set. We do not compare to the NB submodular function, because as demonstrated in~\cite{wei2015submodularity} the KNN submodular function mostly outperformed NB submodular for non Naive-Bayes models. 
For the case of class imbalance and noisy settings, we assume that the training set is imbalanced (or noisy), while the validation set is balanced (or clean). For data selection experiments in this setting, we consider the baselines 1-3 above, i.e. CRAIG, Random and SS + FNN, but with some difference. First, we use a stronger version of the random baseline in the case of class imbalance, which ensures that the selected random set is balanced (and hence has the same class distibution as the validation set). For SS + FNN baseline, we select a subset from the training data using KNN submodular function, but using the validation set instead of the training set. In the case of noisy data, we consider 1, 2 and 4 as baselines for data selection. Finally, we consider the following baselines for active learning. \textbf{1. FASS: } FASS algorithm \cite{wei2015submodularity} selects a subset using KNN submodular function by filtering out the data samples with low uncertainty about predictions. \textbf{2. BADGE: } BADGE algorithm \cite{ash2020deep} selects a subset based on the diverse gradient embedding obtained using hypothesized samples. \textbf{3. Random: } In this baseline, we randomly select a subset at every iteration for the data points to be added in the labeled examples. For data selection, we consider two variants of \textsc{Glister}, one with the Facility Location as a regularized (\textsc{F-Glister}), and the second with random as a regularizer (\textsc{R-Glister}). In the case of active learning, we add one more which is diversity regularized (\textsc{D-Glister}), where the diversity is the pairwise sum of distances. 

\noindent \textbf{Datasets, Model Architecture and Experimental Setup: }  To demonstrate effectiveness of \modelonline\  on real-world datasets, we performed experiments on DNA, SVMGuide, Digits, Letter, USPS (from the UCI machine learning repository), MNIST, and CIFAR-10.
We ran experiments with shallow models and deep models. For shallow models, we used a two-layer fully connected neural network having 100 hidden nodes. We use simple SGD optimizer for training the model. The shallow model experiments were run on the first five datasets, while on MNIST and CIFAR-10 we used a deep model. For MNIST, we use LeNet model~\cite{lecun1989backpropagation}, while for CIFAR-10, we use ResNet-18~\cite{he2016deep}. Wherever the datasets do not a validation set, we split the training set into a train (90\%) and validation set (10\%).

\noindent \textbf{Data Selection for Efficient Learning: } We begin by studying the effect of data selection for efficiency (i.e., to reduce the amount of training time and compute). For this purpose, we compare for different subset sizes of 10\%, 30\%, 50\% in the shallow learning setting. We demonstrate the results on MNIST and CIFAR-10 (for deep learning). In the supplementary material, we show results on the other datasets (i.e., the UCI datasets). The results are shown in Figure \ref{fig:dss_general_imb_experiments} (top row). The first two plots (i.e., a and b) show that \textsc{Glister} and its variants significantly outperform all other baselines (including \textsc{Craig} and Random). To make \textsc{Craig} comparable to \textsc{Glister}, we run the data selection with $L = 20$ (i.e. every 20 epochs). This is mainly due to computational reasons since when run every epoch, our implementation of \textsc{Craig} was much slower than full training, effectively defeating the purpose of data selection. We also included \textsc{Craig Every} baseline for the MNIST dataset, which is CRAIG run every epoch to showcase the performance of CRAIG run every epoch. From the results, we observed that GLISTER still performs better than CRAIG run every epoch for MNIST data selection. We also observe that with just 50\% of the data, \textsc{Glister} achieves comparable accuracy to full training. We also note that facility location and random selection regularization help, particularly for larger subsets, by avoiding the overfitting to the validation set. What is also encouraging that \textsc{Glister-Online} performs well even at \textbf{very small subset sizes}, which is important for data selection (e.g., for doing hyper-parameter turnings several times on very small subsets). Perhaps surprisingly, \textsc{Craig} performs very poorly at small data-sizes. Plots c and d show the timing results on CIFAR-10 and MNIST. We see that on CIFAR-10, \textsc{Glister} \textbf{achieves a 6x speedup at 10\%, 2.5x speedup at 30\%, and 1.5x speedup at 50\%, while loosing 3\%, 1.2\% and 0.2\% respectively in accuracy}. Similarly, for MNIST, \textbf{we see a 3x speedup at 10\% subset, with a loss of only 0.2\% in accuracy}. This timing also includes the subset selection time, thereby making the comparison fair. 
\noindent \textbf{Robust Data Selection: } To check our model's generalization performance when adversaries are present in training data, we run experiments in class imbalance and Noisy label settings for the datasets mentioned above. We artificially generate class-imbalance for the above datasets by removing 90\% of the instances from 30\% of total classes available. For noisy data sets, we flip the labels for a randomly chosen subset of the data where the noise ratio determines the subset's size. In our experimental setting, we use a 30\% noise ratio for the noisy experiments. The class imbalance setting results are shown in Figure~\ref{fig:dss_general_imb_experiments} e,f, and g for CIFAR-10, MNIST, and DNA. The results demonstrate that \modelonline\ again significantly outperforms the other baselines. We note that two of the baselines (random with prior and KNN-submodular with validation set information) have knowledge about the imbalance. It is also worth noting that for CIFAR-10, \textsc{R-Glister} achieves a significant improvement over the other baselines (by around 7\%) for subset sizes 30\% and 50\%. Figure~\ref{fig:dss_general_imb_experiments} h shows the results of a noisy setting in the DNA dataset. On the smaller dataset DNA, \textsc{Glister} and its variants outperform even full training, which is again not surprising because the full data has noise. In contrast, our approach essentially filters out the noise through its data selection. \textsc{Glister} and its variants outperform other baselines by more than 10\%, which is very significant. We provide additional results for both the class imbalance and noisy settings in the supplementary material and show similar gains on the other five smaller datasets.




\noindent \textbf{Active Learning: } Next, we compare \textsc{Glister-Active} to state-of-the-art active learning algorithms including \textsc{FASS} and \textsc{Badge}. The results are shown in Figure~\ref{fig:noise_AL_experiments} (right two plots) on SVM-Guide, Letter, and DNA datasets. Again, we see that \textsc{Glister} and its variants (particularly, with diversity) outperform the existing active learning baselines, including \textsc{Badge}, which is currently the state-of-the-art batch active learning algorithm. Since \textsc{Badge} and \textsc{FASS} have been shown to outpeform techniques like uncertainty Sampling~\cite{settles2009active} and Coreset based approaches~\cite{sener2018active}, we do not compare them in this work. In our supplementary material we compare our algorithms on many more datasets and also consider active learning in the class imbalance scenario. 

\noindent \textbf{Ablation Study and Speedups: } We conclude the experimental section of this paper by studying the different approximations. We compare the different versions of Taylor approximations in Figure~\ref{run-time} and study the effect of $L$ and $r$ on the DNA dataset. In the supplementary material, we provide the ablation study for a few more datasets. For the ablation study with $L$, we study $L = 20, 35$ and $50$. For smaller values of $L$ (e.g., below 20, we see that the gains in accuracy are not worth the reduction in efficiency), while for larger values of $L$, the accuracy gets affected. Similarly, we vary $r$ from $1$ to $20$ (which is 5\% of $k$ on DNA dataset). We observe that $L = 20$ and $r = 3$\% of $k$ generally provides the best tradeoff w.r.t time vs accuracy. Thanks to this choice of $r$ and $L$, we see significant speedups through our data selection, even with smaller neural networks. For example, we see a \textbf{6.75x speedup on DNA, 4.9x speedup on SVMGuide, 4.6x on SatImage, and around 1.5x on USPS and Letter}. Detailed ablation studies on SVMGuide, DNA, SatImage, USPS, and Letter are in the supplementary.

\section{Conclusion}
We present \model, a novel framework that solves a mixed discrete-continuous bi-level optimization problem for efficient training using data subset selection while pivoting on a held-out validation set likelihood for robustness. We study the submodularity of \model\ for simpler classifiers and extend the analysis to our proposed iterative and online data selection algorithm \modelonline, that can be applied to any loss-based learning algorithm. We also extend the model to batch active learning (\modelactive). For these variants of \model, on a wide range of tasks, we empirically demonstrate  improvements (and associated trade-offs) over the state-of-the-art in terms of efficiency, accuracy, and robustness.\looseness-1
\bibliography{dssbib.bib}

\appendix
\appendix
\onecolumn
\section{Proof of NP-Hardness (Lemma~\ref{np-hard})}
In this section, we prove the NP-Hardness of the data selection step (equation~\eqref{glister-disc}), thereby proving Lemma~\ref{np-hard}
\begin{lemma*}
There exists log-likelihood functions $LL_V$ and $LL_T$ and training and validation datasets, such that equation~\eqref{glister-disc} is NP hard.
\end{lemma*}
\begin{proof}
We prove that our subset selection problem is NP-hard by proving that an instance of our data selection step can be viewed as set cover problem.

Our subset selection optimization problem is as follows:
\begin{equation}
    \begin{aligned}
        \underset{{S \subseteq {\mathcal U}, |S| \leq k}}{\max} G_{\theta^0}(S) \text{\hspace{0.2cm}where\hspace{0.2cm}}G_{\theta}(S) = LL_V(\theta + \alpha \nabla_{\theta} LL_T(\theta, S), {\mathcal V})
    \end{aligned}
\end{equation}

Let $(x_i, y_i)$ be the $i^{th}$ data-point and $x_i \in \mathcal{R}^d$ in the validation set $\mathcal{V}$ where $i \in [1, |\mathcal{V}|]$. Let $(x_i^t, y_i^t)$ be the $i^{th}$ data-point and $x_i^t \in \mathcal{R}^d$ in the training set $\mathcal{U}$ where $i \in [1, |\mathcal{U}|]$. Let $k$ be the number of classes and $\theta$ be the model parameter. We assume that the number of training points is equal to the number of features in the training data point i.e.,$|\mathcal{U}| = d$.

Assuming our validation log-likelihood is of the form $LL_V(\theta, \mathcal{V}) = \sum_{i=1}^{\mathcal{V}}\min(y_i \theta^T x_i, 1)$ and that all the data-points in the validation set $\mathcal{V}$ have a label value of 1 i.e., $\forall i, y_i = 1$. Also assume that our validation data point vector is of form $x_i = [w_{1i}, w_{2i}, ..., w_{di}]$. As we shall see below, the $w$ vector actually comes from our set-cover instance (i.e. given an instance of a weighted set cover, we can obtain an instance of our problem). Also, note that the loss function above is a shifted version of negative hinge loss, since $-LL_V(\theta, \mathcal{V}) = \sum_{i=1}^{\mathcal{V}} -\min(y_i \theta^T x_i, 1) = \sum_{i=1}^{\mathcal{V}} \max(-y_i \theta^T x_i, -1) = \sum_{i=1}^{\mathcal{V}} [\max(1 - y_i \theta^T x_i, 0) - 1]$

We assume our model to be of simple linear form $y_i = \theta^T x_i$ and our model parameters to be a zero-vector i.e., $\theta^0 = 0$. We also make an assumption that our training set is in such a form that the log-likelihood gradients for $i^{th}$ data-point is a one-hot encoded vector such that $i^{th}$ feature in the vector is equal to 1. Let us denote this one encoder vector using $1_i$. Under the above assumptions, our subset-selection optimization problem is as follows:

\begin{align}
        \underset{{S \subseteq {\mathcal U}, |S| \leq k}}{\max} \sum_{i=1}^{\mathcal{V}}\min(\sum_{j \in S} 1_j^T x_i, 1)
\end{align}

Since $1_j^Tx_i = w_{ji}$, we have:

\begin{align}
\label{setcover_instance}
        \underset{{S \subseteq {\mathcal U}, |S| \leq k}}{\max} \sum_{i=1}^{\mathcal{V}}\min(\sum_{j \in S}w_{ji}, 1)
\end{align}

The above equation~\eqref{setcover_instance} is a instance of Set-Cover Problem which is "NP-hard". Hence, we have proved that an instance of our subset selection problem can be viewed as set cover problem.

Therefore, we can say that there  exist log-likelihood functions $LL_V$ and $LL_T$ and training and validation datasets, such that equation~\eqref{glister-disc} is NP hard.
\end{proof}
\section{Proof of Submodularity of the Data Selection in \modelonline{}  (Theorem~\ref{theorem-1})}
In this section, we prove the submodularity of the data selection step (equation~\eqref{glister-disc}) with a number of loss functions, thereby proving Theorem~\ref{theorem-1}. For convenience, we first restate Theorem~\ref{theorem-1}.
\begin{theorem*}
If the validation set log-likelihood function $LL_V$ is either the negative logistic loss, the negative squared loss, nagative hinge loss or the negative perceptron loss, the optimization problem in equation~\eqref{glister-disc} is an instance of cardinality constrained non-monotone submodular maximization. When $LL_V$ is the negative cross-entropy loss, the optimization problem in equation~\eqref{glister-disc} is an instance of cardinality constrained weakly submodular maximization. 
\end{theorem*}

\subsection{Negative Logistic Loss}
We start with the negative-logistic loss. Let $(x_i, y_i)$ be the $i^{th}$ data-point in the validation set $\mathcal{V}$ where $i \in [1, |\mathcal{V}|]$ and $(x_i^t, y_i^t)$ be the $i^{th}$ data-point in the training set $\mathcal{U}$ where $i \in [1, |\mathcal{U}|]$. Let $\theta$ be the model parameter.

For convenience, we again provide the discrete optimization problem (equation~\eqref{glister-disc}). Define $G_{\theta}(S) = LL_V(\theta + \alpha \nabla_{\theta} LL_T(\theta, S), {\mathcal V})$. Then, the optimization problem is 
\begin{align} 
    S^{t+1} &= \underset{{S \subseteq {\mathcal U}, |S| \leq k}}{\operatorname{argmax\hspace{0.7mm}}} G_{\theta^t}(S)
\end{align}
Note that we can express  $LL_T(\theta,S)$ as $-\log{L_T(\theta,S)}$ and $-\log{LL_V(\theta,\mathcal{V})}$ as $-\log{L_V(\theta,\mathcal{V})}$ where $L_T$ is training loss and $L_V$ is validation loss.
The lemma below, shows that this is a submodular optimization problem, if the likelihood function $LL_V$ is the negative logistic loss. 
\begin{lemma}
If the validation set log-likelihood function $LL_V$ is the negative logistic loss, then optimizing $G_{\theta^t}(S)$ under cardinality constraints is an instance of cardinality constrained monotone submodular maximization.
\end{lemma}
\begin{proof}
For simplicity of notation, we assume our starting point is $\theta_0$ instead of $\theta_t$. Also, we will use $G(S)$ instead of $G_{\theta_0}(S)$. Given this, and since $LL_V$ is a negative logistic loss, the optimization problem can be written as:
\begin{align}
    G(S) = \sum_{i=1}^{|\mathcal{V}|}-\log\left(1 + \exp(-y_i(\theta_0^T - \alpha {\underset{j \in S}{\sum}\nabla_{\theta}L_T(x_j^t, y_j^t, \theta_0)}^T) x_i)\right)
\end{align}
To the above equation we can add a term $L_{MAX}$ to make the entire second term positive. Since the term $L_{MAX}$ is going to be constant it doesn't affect our optimization problem.
\begin{align}
    G(S) = \sum_{i=1}^{|\mathcal{V}|}\left(L_{MAX}-\log\left(1 + \exp(-y_i \theta_0^T + \alpha \underset{j \in S}{\sum}{\nabla_{\theta}L_T(x_j^t, y_j^t, \theta_0)}^T y_i x_i)\right)\right)
\end{align}
Let, $g_{ij} = {-\nabla_{\theta}L_T(x_j^t, y_j^t, \theta_0)}^T x_i y_i$,
\begin{align}
    G(S) = \sum_{i=1}^{|\mathcal{V}|}\left(L_{MAX}-\log\left(1 + \exp(-y_i \theta_0^T x_i - \alpha \underset{j \in S}{\sum}g_{ij})\right)\right)
\end{align}

We can transform to $g_{ij}$ to a new variable $g^{'}_{ij}$ such that $g_{ij} = g_{min} + g^{'}_{ij}$ where $g_{min} = \underset{i,j}{\min}g_{ij}$. The above transformation ensures that  $g^{'}_{ij} \geq 0$.
\begin{align}
    G(S) &= \sum_{i=1}^{|\mathcal{V}|}\left(L_{MAX}-\log\left(1 + \exp(-y_i \theta_0^T x_i - \alpha \underset{j \in S}{\sum}g^{'}_{ij} - \alpha |S| g_{min})\right)\right) \\
    &= \sum_{i=1}^{|\mathcal{V}|} \left(L_{MAX}-\log\left(1 + \exp(-y_i \theta_0^T x_i) \exp(-\alpha \underset{j \in S}{\sum}g^{'}_{ij}) \exp(-\alpha |S| g_{min})\right)\right)
\end{align}
Let $\exp(-y_i \theta_0^T x_i) = c_i$ and we know that $c_i \geq 0$,

\begin{align}
    G(S) = \sum_{i=1}^{|\mathcal{V}|}\left(L_{MAX}-\log\left(1 + c_i \exp(-\alpha \underset{j \in S}{\sum}g^{'}_{ij}) \exp(-\alpha |S| g_{min})\right)\right)
\end{align}
Since we are considering an subset selection of selecting an subset $S$ where $|S| = k$, we convert our set function $G(S)$ to a proxy set function $\hat{G}(S)$,
\begin{align}
    \hat{G}(S) &= \sum_{i=1}^{|\mathcal{V}|}\left(L_{MAX}-\log\left(1 + c_i \exp(-\alpha \underset{j \in S}{\sum}g^{'}_{ij}) \exp(-\alpha k g_{min})\right)\right) \\
    &= \sum_{i=1}^{|\mathcal{V}|}\left(L_{MAX}-\log\left(1 + c_i \exp(-\alpha \underset{j \in S}{\sum}g^{'}_{ij}) \exp(-\alpha k g_{min})\right)\right) \\
    &= \sum_{i=1}^{|\mathcal{V}|}\left(L_{MAX}-\log\left(1 + C_i \exp(-\alpha \underset{j \in S}{\sum}g^{'}_{ij}))\right)\right)
\end{align}
Where we define $C_i = c_i \exp(-\alpha k g_{min})$ in the last equation. Note that $g(x) = -\log(1 + C\exp(-x))$ is a concave function in $x$ (since $C > 0$), and given that $g^{'}_{ij} \geq 0, \forall i, j$, the function above is a concave over modular function, which is submodular. Note that $g(x)$ is also monotone, and hence $\hat{G}(S)$ is in fact a monotone submodular function. Finally, since we have a cardinality constraint, the optimal solution (as well as the greedy solution) of $\hat{G}(S)$ and $G(S)$ will be the same since they will both be of size $k$. Hence, maximizing $\hat{G}(S)$ is the same as maximizing $G(S)$ under cardinality constraint, which is equivalent to maximizing the likelihood function with the negative logistic loss under cardinality constraints. 
\end{proof}

\subsection{Negative Squared Loss}
Similar to the negative logistic loss case, define the notation as follows. Let $(x_i, y_i)$ be the $i^{th}$ data-point in the validation set $\mathcal{V}$ where $i \in [1, |\mathcal{V}|]$ and $(x_i^t, y_i^t)$ be the $i^{th}$ data-point in the training set $\mathcal{U}$ where $i \in [1, |\mathcal{U}|]$. Let $k$ be the number of classes and $\theta$ be the model parameter. Define $G_{\theta}(S) = LL_V(\theta + \eta \nabla_{\theta} LL_T(\theta, S), {\mathcal V})$. Then, recall that the optimization problem is 
\begin{align} 
    S^{t+1} &= \underset{{S \subseteq {\mathcal U}, |S| = k}}{\operatorname{argmax\hspace{0.7mm}}} G_{\theta^t}(S)
\end{align}
Note that we can express  $LL_T(\theta,S)$ as $-L_T(\theta,S)$ and $LL_V(\theta,\mathcal{V})$ as $-L_V(\theta,\mathcal{V})$ where $L_T$ is training loss and $L_V$ is validation loss.
The lemma below, shows that this is a submodular optimization problem, if the likelihood function $LL_V$ is the negative squared loss. Finally, note that we have cardinality equality constraint (i.e. the set must have exactly $k$ elements) instead of less than equality. It turns out that the cardinality equality constraint is required in this case, because the resulting function is non-monotone, and this requirement will become evident in the proof below. 
\begin{lemma}
If the validation set log-likelihood function $LL_V$ is the negative squared loss, then the problem of optimizing $G_{\theta^t}(S)$ under a constraint $|S| = k$ is an instance of cardinality constrained non-monotone submodular maximization. 
\end{lemma}
\begin{proof}
Again, for convenience, we assume we start at $\theta_0$, and we use $G(S)$ instead of $G_{\theta_0}(S)$. With the negative squared loss, the optimization problem becomes:
\begin{align}
    G(S) = -\sum_{i=1}^{|\mathcal{V}|} \left\Vert y_i - x_i^T(\theta_0 - \alpha \underset{j \in S}{\sum}\nabla_{\theta}L_T(x_j^t, y_j^t, \theta_0))\right\Vert^2
\end{align}
\begin{align}
    G(S) = -\sum_{i=1}^{|\mathcal{V}|} \left\Vert(y_i - x_i^T\theta_0) + x_i^T\alpha \underset{j \in S}{\sum}\nabla_{\theta}L_T(x_j^t, y_j^t, \theta_0)\right\Vert^2
\end{align}
Expanding the square, we obtain:
\begin{align}
    G(S) = -\sum_{i=1}^{|\mathcal{V}|} \left\Vert(y_i - x_i^T\theta_0)\right\Vert^2 -\sum_{i=1}^{|\mathcal{V}|}\left\Vert x_i^T\alpha \underset{j \in S}{\sum}\nabla_{\theta}L_T(x_j^t, y_j^t, \theta_0)\right\Vert^2
    -2\alpha\sum_{i=1}^{|\mathcal{V}|}\sum_{j=1}^{|S|}(y_i - x_i^T\theta_0) (x_i^T \nabla_{\theta}L_T(x_j^t, y_j^t, \theta_0))
\end{align}
Since, $\nabla_{\theta}L_V(x_i, y_i, \theta_0) = -2(y_i - x_i^T\theta_0) (x_i)$ when $L_V$ is squared loss.
\begin{align}
    G(S) = -\sum_{i=1}^{|\mathcal{V}|} \left\Vert(y_i - x_i^T\theta_0)\right\Vert^2 -\sum_{i=1}^{|\mathcal{V}|}\left\Vert x_i^T\alpha \underset{j \in S}{\sum}\nabla_{\theta}L_T(x_j^t, y_j^t, \theta_0)\right\Vert^2
    +\alpha\sum_{i=1}^{|\mathcal{V}|}\sum_{j=1}^{|S|}\nabla_{\theta}L_V(x_i, y_i, \theta_0)^T \nabla_{\theta}L_T(x_j^t, y_j^t, \theta_0))
\end{align}

\begin{equation}
    \begin{aligned}
    G(S) = -\sum_{i=1}^{|\mathcal{V}|} \left\Vert(y_i - x_i^T\theta_0)\right\Vert^2 
    +\alpha\sum_{j=1}^{|S|}\sum_{i=1}^{|\mathcal{V}|}\nabla_{\theta}L_V(x_i, y_i, \theta_0)^T \nabla_{\theta}L_T(x_j^t, y_j^t, \theta_0) \\
    - \alpha^2 \sum_{j=1}^{|S|}\sum_{k=1}^{|S|}\sum_{i=1}^{|\mathcal{V}|} (x_i^T \nabla_{\theta}L_T(x_j^t, y_j^t, \theta_0)) (x_i^T \nabla_{\theta}L_T(x_k^t, y_k^t, \theta_0))
    \end{aligned}
\end{equation}
Assuming $S_{jk} = \sum_{i=1}^{|\mathcal{V}|} (x_i^T \nabla_{\theta}L_T(x_j^t, y_j^t, \theta_0)) (x_i^T \nabla_{\theta}L_T(x_k^t, y_k^t, \theta_0))$, we have:
\begin{equation}
    \begin{aligned}
    G(S) = -\sum_{i=1}^{|\mathcal{V}|} \left\Vert(y_i - x_i^T\theta_0)\right\Vert^2 
    +\alpha\sum_{j=1}^{|S|}\sum_{i=1}^{|\mathcal{V}|}\nabla_{\theta}L_V(x_i, y_i, \theta_0)^T \nabla_{\theta}L_T(x_j^t, y_j^t, \theta_0)
    - \alpha^2 \sum_{j=1}^{|S|}\sum_{k=1}^{|S|}S_{jk}
    \end{aligned}
\end{equation}

Since, $S_{jk}$ is not always greater than zero, we can transform it to $\hat{S}_{jk}$ such that ${S}_{jk} = \hat{S}_{jk} + S_{min}$ where $S_{min} = \underset{j,k}{\min}S_{jk}$. This transformation ensures that $\hat{S}_{jk} \geq 0$.
\begin{align}
   G(S) &= -\sum_{i=1}^{|\mathcal{V}|} \left\Vert(y_i - x_i^T\theta_0)\right\Vert^2 
    +\alpha\sum_{j=1}^{|S|}\sum_{i=1}^{|\mathcal{V}|}\nabla_{\theta}L_V(x_i, y_i, \theta_0)^T \nabla_{\theta}L_T(x_j^t, y_j^t, \theta_0)
    - \alpha^2 \sum_{j=1}^{|S|}\sum_{k=1}^{|S|}(\hat{S}_{jk} + {S}_{min}) \\
    &= -\sum_{i=1}^{|\mathcal{V}|} \left\Vert(y_i - x_i^T\theta_0)\right\Vert^2 + \alpha\sum_{j=1}^{|S|}\sum_{i=1}^{|\mathcal{V}|}\nabla_{\theta}L_V(x_i, y_i, \theta_0)^T \nabla_{\theta}L_T(x_j^t, y_j^t, \theta_0) - \alpha^2 \sum_{j=1}^{|S|}\sum_{k=1}^{|S|}\hat{S}_{jk} - \alpha^2 |S|^2 {S}_{min}
\end{align}
Since we are considering an subset selection of selecting an subset $S$ where $|S| = K$, we convert our set function $G(S)$ to a proxy set function $\hat{G}(S)$,
\begin{equation}
    \begin{aligned}
    \hat{G}(S) = (-\sum_{i=1}^{|\mathcal{V}|} \left\Vert(y_i - x_i^T\theta_0)\right\Vert^2 - \alpha^2 K^2 {S}_{min}) + \alpha\sum_{j=1}^{|S|}\sum_{i=1}^{|\mathcal{V}|}\nabla_{\theta}L_V(x_i, y_i, \theta_0)^T \nabla_{\theta}L_T(x_j^t, y_j^t, \theta_0) - \alpha^2 \sum_{j=1}^{|S|}\sum_{k=1}^{|S|}\hat{S}_{jk} 
    \end{aligned}
\end{equation}

In the above equation, the first part $(-\sum_{i=1}^{|\mathcal{V}|} \left\Vert(y_i - x_i^T\theta_0)\right\Vert^2 - \alpha^2 K^2 {S}_{min})$ is constant and does not depend on subset $S$. The second part $\alpha\sum_{j=1}^{|S|}\sum_{i=1}^{|\mathcal{V}|}\nabla_{\theta}L_V(x_i, y_i, \theta_0)^T \nabla_{\theta}L_T(x_j^t, y_j^t, \theta_0)$ is a non-monotone modular in subset $S$. Note that the third part $- \alpha^2 \sum_{j=1}^{|S|}\sum_{k=1}^{|S|}\hat{S}_{jk}$ is similar to a graph cut function since $\hat{S}_{jk} \geq 0$. Hence it is a sub-modular function.

Therefore, the proxy set function $\hat{G}(S)$ is a non-monotone constrained submodular function. Finally, note that under a cardinality equality constraint, optimizing $G(S)$ is the same as optimizing $\hat{G}(S)$ since the solution by the randomized greedy algorithm and the optimal solution will both be of size $k$. Finally, note that fast algorithms exist for non-monotone submodular maximization under cardinaltity equality constraints~\cite{buchbinder2014submodular}.
\end{proof}

\subsection{Negative Hinge Loss and Negative Perceptron Loss}
Again, we define the notation. Let $(x_i, y_i)$ be the $i^{th}$ data-point in the validation set $\mathcal{V}$ where $i \in [1, |\mathcal{V}|]$ and $(x_i^t, y_i^t)$ be the $i^{th}$ data-point in the training set $\mathcal{U}$ where $i \in [1, |\mathcal{U}|]$. Let $k$ be the number of classes and $\theta$ be the model parameter. Define $G_{\theta}(S) = LL_V(\theta + \eta \nabla_{\theta} LL_T(\theta, S), {\mathcal V})$. 
Then, recall that the optimization problem is 
\begin{align} 
    S^{t+1} &= \underset{{S \subseteq {\mathcal U}, |S| = k}}{\operatorname{argmax\hspace{0.7mm}}} G_{\theta^t}(S)
\end{align}
Note that we can express  $LL_T(\theta,S)$ as $-L_T(\theta,S)$ and $LL_V(\theta,\mathcal{V})$ as $-L_V(\theta,\mathcal{V})$ where $L_T$ is training loss and $L_V$ is validation loss. 
The lemma below, shows that this is a submodular optimization problem, if the likelihood function $LL_V$ is the negative hinge loss (or negative perceptron loss).
\begin{lemma}
If the validation set log-likelihood function $LL_V$ is the negative hinge loss (or negative perceptron loss), then the  problem of optimizing $G_{\theta^t}(S)$ under cardinality constraints is an instance of cardinality constrained monotone submodular maximization. 
\end{lemma}
\begin{proof}
Similar to the proofs earlier, we assume we start at $\theta_0$, and we use $G(S)$ instead of $G_{\theta_0}(S)$. With the negative hinge loss, we obtain the following problem:
\begin{align}
    G(S) &= -\sum_{i=1}^{|\mathcal{V}|} \max\left(0, 1 - y_ix_i^T\left(\theta_0 - \underset{j \in S}{\sum}\nabla_{\theta}L_T(x_j^t. y_j^t, \theta_0)\right)\right) \\
    &= -\sum_{i=1}^{|\mathcal{V}|} \max\left(0, 1 -y_ix_i^T\theta_0 + \underset{j \in S}{\sum}y_ix_i^T\nabla_{\theta}L_T(x_j^t. y_j^t, \theta_0)\right)
\end{align}
Note that the expression for the Perceptron loss is similar, except for the $1$ in the Hinge Loss. Hence, we just prove it for the Hinge loss. Next, define $c_i = -1 + y_ix_i^T\theta_0$ and $g_{ij} = -y_ix_i^T\nabla_{\theta}L_T(x_j^t. y_j^t, \theta_0)$,
\begin{align}
    G(S) = -\sum_{i=1}^{|\mathcal{V}|} \max\left(0, -c_i - \underset{j \in S}{\sum} g_{ij}\right)
\end{align}
Since, $g_{ij}$ is not always greater than zero, we can transform it to $\hat{g}_{ij}$ such that ${g}_{ij} = \hat{g}_{ij} + g_{min}$ where $g_{min} = \underset{i, j}{\min}g_{ij}$. This transformation ensures that $\hat{g}_{ij} \geq 0$.
\begin{align}
    G(S) = -\sum_{i=1}^{|\mathcal{V}|} \max\left(0, -c_i - |S|g_{min} - \underset{j \in S}{\sum} \hat{g}_{ij}\right)
\end{align}

Since we are considering an subset selection of selecting an subset $S$ where $|S| = K$, we convert our set function $G(S)$ to a proxy set function $\hat{G}(S)$,
\begin{align}
    \hat{G}(S) = -\sum_{i=1}^{|\mathcal{V}|} \max\left(0, -c_i - Kg_{min} - \underset{j \in S}{\sum} \hat{g}_{ij}\right)
\end{align}

Define $c_i + K g_{min} = C_i$,
\begin{align}
    \hat{G}(S) &= -\sum_{i=1}^{|\mathcal{V}|} \max\left(0, - C_i - \underset{j \in S}{\sum} \hat{g}_{ij}\right) \\
    &= \sum_{i=1}^{|\mathcal{V}|} \min\left(0,  C_i + \underset{j \in S}{\sum} \hat{g}_{ij}\right)
\end{align}
Note that $g(x)= \min\left(0, C+x\right)$ is a concave function in $x$ and given that $\hat{g}_{ij} \geq 0$, the function above is a concave over modular function, which is sub-modular. Note that $g(x)$ is also a monotone function, and hence the above proxy set function $\hat{G}(S)$ is a monotone constrained sub-modular function. Finally, because $\hat{G}(S)$ is a monotone function, the optimal solution as well as the greedy solution will both achieve sets of size $K$, and hence optimizing $\hat{G}(S)$ is equivalent to optimizing $G(S)$ (under cardinality constraints), which is the problem of optimizing the likelihood function with the negative hinge loss. We can obtain a very similar expression for the negative perceptron loss, with the only difference that $c_i =  y_ix_i^T\theta_0$ instead of $c_i = -1 + y_ix_i^T\theta_0$. 
\end{proof}

\subsection{Negative Cross Entropy Loss}
Let $(x_i, y_i)$ be the $i^{th}$ data-point in the validation set $\mathcal{V}$ where $i \in [1, |\mathcal{V}|]$ and $(x_i^t, y_i^t)$ be the $i^{th}$ data-point in the training set $\mathcal{U}$ where $i \in [1, |\mathcal{U}|]$. Let $k$ be the number of classes and $\theta$ be the model parameter. Define $G_{\theta}(S) = LL_V(\theta + \eta \nabla_{\theta} LL_T(\theta, S), {\mathcal V})$. Then, recall that the optimization problem is 
\begin{align} 
    S^{t+1} &= \underset{{S \subseteq {\mathcal U}, |S| = k}}{\operatorname{argmax\hspace{0.7mm}}} G_{\theta^t}(S)
\end{align}
Note that we can express  $LL_T(\theta,S)$ as $-L_T(\theta,S)$ and $LL_V(\theta,\mathcal{V})$ as $-L_V(\theta,\mathcal{V})$ where $L_T$ is training loss and $L_V$ is validation loss.
Unlike the negative logistic, squared and hinge loss cases, with the negative cross-entropy loss, the optimization problem above is no longer submodular. However, as we show below, it is approximately submodular. Before stating the result, we first recall the definition of approximate submodularity. 

\textbf{Definition:} A function is called $\beta$-submodular~\cite{DBLP:journals/corr/abs-1811-07863}, if the gain of adding an element $e$ to set $X$ is $\beta$ times greater than or equals to the gain of adding an element $e$ to set $Y$ where $X$ is a subset of $Y$.
i.e.,
\begin{equation}
    \underset{X,Y | X \subseteq Y}{\forall}G(e|X) \geq \beta G(e|Y) 
\end{equation}

While this definition is different from the notion of approximate submodularity~\cite{das2011submodular}. However, they are closely related since (following Proposition 4 in~\cite{DBLP:journals/corr/abs-1811-07863}), the approximate submodularity parameter $\gamma \geq \beta$. This immediately implies the following result:
\begin{lemma}\cite{das2011submodular,DBLP:journals/corr/abs-1811-07863} Given a $\beta$-approximate monotone submodular function $G$, the greedy algorithm achieves a $1 - e^{-\beta}$ approximation factor for the problem of maximizing $G$ subject to cardinality constraints. 
\end{lemma}
Next, we show that $G_{\theta_t}(S)$ is a $\beta$-approximate submodular function. To do this, first define $R$ as the largest value of $||x_i||$ in the validation and training datasets (i.e. we assume that the norm of the feature vectors are bounded). Note that this is common assumption made in most convergence analysis results. Then, we show that with the cross entropy loss, $G_{\theta_t}(S)$ is a $\beta$-approximate with $\beta = 1/(2R^2 + 1)$. 
\begin{lemma}
If the validation set log-likelihood function $LL_V$ is the negative cross entropy loss and training set loss function $L_T$ is cross entropy loss, then the optimization problem in equation~\eqref{glister-disc} is an instance of cardinality constrained $\beta$-approximate submodular maximization, where $\beta = 1/(2R^2 + 1)$ ($R$ satisfies $||x_i|| \leq R$ for the training and validation data). 
\end{lemma}
Before proving our result, we note that if we normalize the features to have a norm of $1$, the approximation guarantee is $1 - e^{-0.3}$.
\begin{proof}
Similar to the proofs earlier, we assume we start at $\theta_0$, and we use $G(S)$ instead of $G_{\theta_0}(S)$. With the negative cross entropy loss, we obtain the following problem:
\begin{align}
    G(S) = \sum_{i=1}^{|\mathcal{V}|}\log\left(\frac{\exp((\theta_{0}^{y_i} - \alpha \sum_{j \in S}\nabla_{\theta}L_T(x_j^t, y_j^t, \theta_0^{y_i}))^T x_{i})}{\sum_{k}\exp((\theta_{0}^{k} - \alpha \sum_{j \in S}\nabla_{\theta}L_T(x_j^t, y_j^t, \theta_0^{k}))^T x_{i})}\right)
\end{align}
Expanding this out, we obtain:
\begin{equation}
\begin{aligned}
    G(S) =  \sum_{i=1}^{|\mathcal{V}|}\left(\log\left(\exp((\theta_{0}^{y_i} - \alpha \sum_{j \in S}\nabla_{\theta}L_T(x_j^t, y_j^t, \theta_0^{y_i}))^T x_{i})\right) -
    \log\left(\sum_{k}\exp((\theta_{0}^{k} - \alpha \sum_{j \in S}\nabla_{\theta}L_T(x_j^t, y_j^t, \theta_0^{k}))^T x_{i})\right)\right)
\end{aligned}
\end{equation}
which can be written as:
\begin{align}
    G(S) &= \sum_{i=1}^{|\mathcal{V}|}\left(\left((\theta_{0}^{y_i} - \alpha \sum_{j \in S}\nabla_{\theta}L_T(x_j^t, y_j^t, \theta_0^{y_i}))^T x_{i}\right) - \log\left(\sum_{k}\exp((\theta_{0}^{k} - \alpha \sum_{j \in S}\nabla_{\theta}L_T(x_j^t, y_j^t, \theta_0^{k}))^T x_{i})\right)\right) \\
    &= \sum_{i=1}^{|\mathcal{V}|} \left((\theta_{0}^{y_i})^T x_{i} - \alpha (\sum_{j \in S}\nabla_{\theta}L_T(x_j^t, y_j^t, \theta_0^{y_i}))^T x_{i} - \log\left(\sum_{k}\exp((\theta_{0}^{k} - \sum_{j \in S}\nabla_{\theta}L_T(x_j^t, y_j^t, \theta_0^{k}))^T x_{i})\right)\right)
\end{align}
Since, the term $(\theta_{0}^{y_i})^T x_{i}$ does not depend on the subset $S$, we can remove it from our optimization problem,
\begin{align}
    G(S) = \sum_{i=1}^{|\mathcal{V}|} \left(\sum_{j \in S} - \alpha \nabla_{\theta}{L_T(x_j^t, y_j^t, \theta_0^{y_i})}^T x_{i} - \log\left(\sum_{k}\exp((\theta_{0}^{k} - \alpha \sum_{j \in S}\nabla_{\theta}L_T(x_j^t, y_j^t, \theta_0^{k}))^T x_{i})\right)\right)
\end{align}
Assume $g_{ijk} = \nabla_{\theta}{L_T(x_j^t, y_j^t, \theta_0^{k})}^T x_{i}$,
\begin{align}
    G(S) =  \sum_{i=1}^{|\mathcal{V}|} \left( \sum_{j \in S} - \alpha g_{ij{y_{i}}} - \log\left(\sum_{k}\exp(\theta_{0}^{k^T} x_{i} - \alpha \sum_{j \in S}g_{ijk})\right)\right)
\end{align}

\begin{align}
    G(S) = \sum_{i=1}^{|\mathcal{V}|} \left(\sum_{j \in S} -\alpha g_{ijy_{i}} - \log\left(\sum_{k}\exp(\theta_{0}^{k^T} x_{i}) \exp(-\alpha \sum_{j \in S}g_{ijk})\right)\right)
\end{align}

Let $c_{ik} = \exp(\theta_{0}^{k^T} x_{i})$ and also note that $c_{ik} \geq 0$.
\begin{align}
    G(S) = \sum_{i=1}^{|\mathcal{V}|}\left( \sum_{j \in S} -\alpha g_{ijy_{i}} - \log\left(\sum_{k}c_{ik} \exp(- \alpha \sum_{j \in S}g_{ijk})\right)\right)
\end{align}
Since $g_{ijk}$ is not always greater than zero, we need to make some transformations to convert the problem into a monotone submodular function. First, we define a transformation of $g_{ijk}$ to $g_{ijk}^{'}$ such that $g_{ijk} = g^{'}_{ijk} + g_{min} - 1$ where $g_{min} = \underset{i,j,k}{\min}g_{ijk}$. This transformation ensures that $g_{ijk}^{'} \geq 1$. Also, define $g_{negmin} = \underset{i,j,k}{\min}(-g_{ijk})$, and then we define a transformation of $g_{ijk}$ to $g_{ijk}^{''}$ such that $-g_{ijk} = g^{''}_{ijk} + g_{negmin}$. Note that both $g^{'}_{ijk}$ and $g^{''}_{ijk}$ are greater than or equal to zero after these transformations. 

\begin{align}
    G(S) &= \sum_{i=1}^{|\mathcal{V}|} \left( \sum_{j \in S} \alpha (g^{''}_{ijy_{i}} + g_{negmin}) - \log\left(\sum_{k}c_{ik} \exp(- \alpha \sum_{j \in S}(g^{'}_{ijk} + g_{min} - 1))\right)\right) \\
    &= \alpha|\mathcal{V}||S|(g_{negmin}) + \sum_{i=1}^{|\mathcal{V}|} \left(\sum_{j \in S} \alpha (g^{''}_{ijy_{i}})  - \log\left(\sum_{k}c_{ik} \exp(- \alpha \sum_{j \in S}(g^{'}_{ijk})) \exp(|S|(g_{min} - 1))\right)\right)
\end{align}
Since we are considering an subset selection of selecting an subset $S$ where $|S| = K$, we convert our set function $G(S)$ to a proxy set function $\hat{G}(S)$. 

\begin{align}
    \hat{G}(S) = \alpha|\mathcal{V}|K(g_{negmin}) + \sum_{i=1}^{|\mathcal{V}|} \left(\sum_{j \in S} \alpha (g^{''}_{ijy_{i}})  - \log\left(\sum_{k}c_{ik} \exp(- \alpha \sum_{j \in S}(g^{'}_{ijk})) \exp(K(g_{min} - 1))\right)\right)
\end{align}
Next, define $C_{ik} = c_{ik} \exp(K(g_{min} - 1))$. Also, since $\alpha|\mathcal{V}|K(g_{negmin})$ is a constant, we can remove it from the optimization problem and we can define:
\begin{align}
    \hat{G}(S) =  \sum_{i=1}^{|\mathcal{V}|} \sum_{j \in S} \alpha (g^{''}_{ijy_{i}})  - \sum_{i=1}^{|\mathcal{V}|} \log\left(\sum_{k}C_{ik} \exp(- \alpha \sum_{j \in S}(g^{'}_{ijk}))\right)
\end{align}

In the above equation, the first part $G_1(X) =   \sum_{i=1}^{|\mathcal{V}|} \sum_{j \in S} \alpha (g^{''}_{ijy_{i}})$ is monotone modular function in $S$. The second part $G_2(X) = - \sum_{i=1}^{|\mathcal{V}|} \log\left(\sum_{k}C_{ik} \exp(- \alpha \sum_{j \in S}(g^{'}_{ijk}))\right)$ is a monotone function, but is unfortunately not necessary submodular. In the next part of this proof, we focus on this function and show that it is $\beta$-submodular. Moreover, since the first part is positive modular, it is easy to see that if $G_2(X)$ is $\beta$ submodular (for $\beta < 1$), $G_1(X) + G_2(X)$ will also be $\beta$-submodular. To show this, recall that a function $h(X)$ is $\beta$-submodular if $h(j | X) \geq \beta h(j | Y)$ for all subsets $X \subseteq Y$. If we assume that $G_2$ is $\beta$-submodular, then it will hold that $G_2(j | X) \geq \beta G_2(j | Y)$ for all subsets $X \subseteq Y$. Furthermore, since $G_1$ is modular, it holds that $G_1(j | X) = G_1(j | Y) \geq \beta G_1(j | Y)$ since $G_1$ is positive modular. Hence $G_1 + G_2$ is $\beta$-submodular. 

\subsubsection{Proof of $\beta$-submodularity of $G_2$:} First notice that $G_2(e|X)$ of adding an element $e$ to set $X$ is:
\begin{equation}
    G_2(e|X) = - \sum_{i=1}^{|\mathcal{V}|} \log\left(\sum_{k}C_{ik} \exp(- \alpha \sum_{j \in S}(g^{'}_{ijk}) - \alpha g^{'}_{iek})\right) + \sum_{i=1}^{|\mathcal{V}|} \log\left(\sum_{k}C_{ik} \exp(- \alpha \sum_{j \in S}(g^{'}_{ijk}))\right) 
\end{equation}
It then follows that,
\begin{equation}
\begin{aligned}
    G_2(e|X) &\geq  - \sum_{i=1}^{|\mathcal{V}|} \log\left(\sum_{k}C_{ik} \exp(- \alpha \sum_{j \in S}(g^{'}_{ijk}) - \alpha g^{'}_{min})\right) + \sum_{i=1}^{|\mathcal{V}|} \log\left(\sum_{k}C_{ik} \exp(- \alpha \sum_{j \in S}(g^{'}_{ijk}))\right)
\end{aligned}
\end{equation}
where $g^{'}_{min} = \underset{i,j,k}{\min}g^{'}_{ijk}$. This then implies: 
\begin{align}
    G_2(e|X) &\geq - \sum_{i=1}^{|\mathcal{V}|} \log\left(\sum_{k}C_{ik} \exp(- \alpha \sum_{j \in S}(g^{'}_{ijk})) \exp(- \alpha g^{'}_{min})\right) + \sum_{i=1}^{|\mathcal{V}|} \log\left(\sum_{k}C_{ik} \exp(- \alpha \sum_{j \in S}(g^{'}_{ijk}))\right) \\
    &\geq - \sum_{i=1}^{|\mathcal{V}|} \log\left(\sum_{k}C_{ik} \exp(- \alpha \sum_{j \in S}(g^{'}_{ijk}))\right) + \sum_{i=1}^{|\mathcal{V}|} \alpha g^{'}_{min} + \sum_{i=1}^{|\mathcal{V}|} \log\left(\sum_{k}C_{ik} \exp(- \alpha \sum_{j \in S}(g^{'}_{ijk}))\right) \\
    &\geq |\mathcal{V}| \alpha g^{'}_{min}
\end{align}

Similarly,
\begin{equation}
\begin{aligned}
    G_2(e|X) \leq  - \sum_{i=1}^{|\mathcal{V}|} \log\left(\sum_{k}C_{ik} \exp(- \alpha \sum_{j \in S}(g^{'}_{ijk}) - \alpha g^{'}_{max})\right) + \sum_{i=1}^{|\mathcal{V}|} \log\left(\sum_{k}C_{ik} \exp(- \alpha \sum_{j \in S}(g^{'}_{ijk}))\right) 
\end{aligned}
\end{equation}
where $g^{'}_{max} = \underset{i,j,k}{\max}g^{'}_{ijk}$. This implies that:
\begin{align}
    G_2(e|X) &\leq - \sum_{i=1}^{|\mathcal{V}|} \log\left(\sum_{k}C_{ik} \exp(- \alpha \sum_{j \in S}(g^{'}_{ijk})) \exp(- \alpha g^{'}_{max})\right) + \sum_{i=1}^{|\mathcal{V}|} \log\left(\sum_{k}C_{ik} \exp(- \alpha \sum_{j \in S}(g^{'}_{ijk}))\right) \\
    &\leq \sum_{i=1}^{|\mathcal{V}|} \log\left(\sum_{k}C_{ik} \exp(- \alpha \sum_{j \in S}(g^{'}_{ijk}))\right) + \sum_{i=1}^{|\mathcal{V}|} \alpha g^{'}_{max} + \sum_{i=1}^{|\mathcal{V}|} \log\left(\sum_{k}C_{ik} \exp(- \alpha \sum_{j \in S}(g^{'}_{ijk}))\right) \\
    &\leq |\mathcal{V}| \alpha g^{'}_{max}
\end{align}

So, from the above two bounds on $G_2(e|X)$, we have:

\begin{equation}
    \underset{X, Y | X \subseteq Y}{\forall} \frac{G_2(e|X)}{G_2(e|Y)} \geq \frac{g^{'}_{min}}{g^{'}_{max}}
\end{equation}

Since, $g_{ijk} = \nabla_{\theta}L_T(x_j^t, y_j^t, \theta_0^k)^Tx_i$ and $g^{'}_{ijk} = g_{ijk} - g_{min} + 1$, we have:

\begin{equation}
    g^{'}_{ijk} = \nabla_{\theta}L_T(x_j^t, y_j^t, \theta_0^k)^T x_i - \underset{i,j,k}{\min}\nabla_{\theta}L_T(x_j^t, y_j^t, \theta_0^k)^T x_i + 1
\end{equation}

Since, $L_T$ is also a cross-entropy loss, we know that $\nabla_{\theta}L_T(x_j^t, y_j^t, \theta_0^k) = (1_{y_j} - p)x_j$ where $1_{y_j} = 1$ when $k=y_j$ for final layer parameters. Hence,
\begin{align}
 \underset{i,j,k}{\forall}\nabla_{\theta}L_T((x_j^t, y_j^t, \theta_0^k)) \leq R \text{\hspace{0.2cm}where\hspace{0.2cm}} R \geq \underset{i}{\forall}\left\Vert x_i^t \right\Vert
\end{align}

Similarly,
\begin{align}
 \underset{i,j,k}{\forall}\nabla_{\theta}L_T((x_j^t, y_j^t, \theta_0^k)) \geq -R \text{\hspace{0.2cm}where\hspace{0.2cm}} R \geq \underset{i}{\forall}\left\Vert x_i^t \right\Vert
\end{align}

Similarly, the norm of validation set data points is bounded by R, and therefore we obtain that $g^{'}_{min} = 1$ and $g^{'}_{max} = 2R^2+1$. As a result,
\begin{align}
    \underset{X, Y | X \subseteq Y}{\forall} \frac{G_2(e|X)}{G_2(e|Y)} \geq \frac{1}{2R^2+1}
\end{align}

Hence, this implies that $G_2$ is $\beta$ submodular with $\beta = \frac{1}{2R^2+1}$, which further implies that $\hat{G}$ is $\beta$-submodular. Finally, we are interested in the constraint that $|S| = K$, the optimal solution as well as the greedy solution will both obtain sets of size $K$ and hence the optimizer of $G$ will be the same as the optimizer of $\hat{G}$. Hence the greedy algorithm will achieve a $1 - e^{-\beta}$ approximation factor, for the data selection step with the cross entropy loss.
\end{proof}

\section{Convergence of \modelonline\ and Reduction in Objective Value}
This section provides the proof for Theorem~\ref{theorem-2} and  Theorem~\ref{theorem-3}. 

\subsection{Reduction in Objective Values}
\begin{theorem*}
Suppose the validation loss function $L_V$ is Lipschitz smooth with constant L, and the gradients of training and validation losses are $\sigma_T$ and $\sigma_V$ bounded respectively. Then the validation loss always monotonically decreases with every training epoch $l$, i.e. $L_V(\theta_{l+1}) \leq L_V(\theta_{l})$  if it satisfies the condition that $\nabla_{\theta}L_V(\theta_{l}, {\mathcal V})^{T} \nabla_{\theta}L_T(\theta_{l}, S) \geq 0$ for $0 \leq l \leq T$ and the learning rate $\alpha \leq \min_l \frac{2\|\nabla_{\theta}L_V(\theta_{l}, {\mathcal V})\|\cos(\Theta_l)}{L\sigma_T}$ where $\Theta_l$ is the angle between $\nabla_{\theta}L_V(\theta_{l}, {\mathcal V})$ and $\nabla_{\theta}L_T(\theta_{l}, S)$. 
\end{theorem*}
\begin{proof}

Suppose we have a validation set $\mathcal{V}$ and the loss on the validation set is $L_V(\theta, \mathcal{V})$. Suppose the subset selected by the \modelonline\ framework is denoted by $S$ and the subset training loss is $L_T(\theta, S)$. Since validation loss $L_V$ is lipschitz smooth, we have,
\begin{equation}
    \begin{aligned}
    \label{lipschitz-smooth}
    L_V(\theta_{l+1}, \mathcal{V}) \leq L_V(\theta_l, \mathcal{V}) + \nabla_{\theta}^{T}(\theta_l, \mathcal{V}) \Delta \theta + \frac{L}{2}{\left\Vert\Delta \theta\right\Vert}^2, \text{\hspace{0.5cm}where,\hspace{0.2cm}} \Delta \theta = \theta_{l+1} - \theta_l
    \end{aligned}
\end{equation}

Since, we are using SGD to optimize the subset training loss $L_T(\theta, S)$ model parameters our update equations will be as follows:
\begin{equation}
\label{sgd}
    \theta_{l+1} = \theta_l - \alpha \nabla_{\theta}L_T(\theta_l, S)
\end{equation}

Plugging our updating rule (Eq. \eqref{sgd}) in (Eq. \eqref{lipschitz-smooth}):
\begin{equation}
    \begin{aligned}
    \label{lipschitz-smooth1}
    L_V(\theta_{l+1}, \mathcal{V}) \leq L_V(\theta_l, \mathcal{V}) + \nabla_{\theta}^{T}L_V(\theta_l, \mathcal{V}) (- \alpha \nabla_{\theta}L_T(\theta_l, S)) + \frac{L}{2}{\left\Vert- \alpha \nabla_{\theta}L_T(\theta_l, S)\right\Vert}^2
    \end{aligned}
\end{equation}

Which gives,
\begin{equation}
    \begin{aligned}
    \label{final-equation}
    L_V(\theta_{l+1}, \mathcal{V}) - L_V(\theta_l, \mathcal{V}) \leq \nabla_{\theta}^{T}L_V(\theta_l, \mathcal{V}) (- \alpha \nabla_{\theta}L_T(\theta_l, S)) + \frac{L}{2}{\left\Vert- \alpha \nabla_{\theta}L_T(\theta_l, S)\right\Vert}^2
    \end{aligned}
\end{equation}

From (Eq. \eqref{final-equation}), note that:
\begin{equation}
\begin{aligned}
\label{thm2-cndtn}
L_V(\theta_{l+1}, \mathcal{V}) \leq L_V(\theta_{l}, \mathcal{V}) \text{\hspace{0.2cm}if\hspace{0.2cm}}  \nabla_{\theta}^{T}L_V(\theta_l, \mathcal{V}) \nabla_{\theta}L_T(\theta_l, S)) - \frac{\alpha L}{2}{\left\Vert \nabla_{\theta}L_T(\theta_l, S)\right\Vert}^2 \geq 0
\end{aligned}
\end{equation}

Since we know that ${\left\Vert \nabla_{\theta}L_T(\theta_l, S)\right\Vert}^2 \geq 0$, we will have the necessary condition 
$\nabla_{\theta}^{T}L_V(\theta_l, \mathcal{V}) \nabla_{\theta}L_T(\theta_l, S)) \geq 0$.

We can also re-write the condition in (Eq:\eqref{thm2-cndtn}) as follows:

\begin{equation}
\begin{aligned}
\label{learning_rate_cndtn}
    \nabla_{\theta}^{T}L_V(\theta_l, \mathcal{V}) \nabla_{\theta}L_T(\theta_l, S)) \geq \frac{\alpha L}{2} {\nabla_{\theta}L_T(\theta_l, S)}^T\nabla_{\theta}L_T(\theta_l, S)
\end{aligned}
\end{equation}

The Eq:\eqref{learning_rate_cndtn} gives the necessary condition for learning rate i.e.,
\begin{equation}
\begin{aligned}
\label{eta_cndtn}
    \alpha \leq \frac{2}{L} \frac{\nabla_{\theta}L_V(\theta_l, \mathcal{V})^T \nabla_{\theta}L_T(\theta_l, S))}{{\nabla_{\theta}L_T(\theta_l, S)}^T\nabla_{\theta}L_T(\theta_l, S)}
\end{aligned}
\end{equation}

The above Eq:\eqref{eta_cndtn} can be written as follows:
\begin{equation}
\begin{aligned}
    \alpha \leq \frac{2}{L} \frac{\left\Vert \nabla_{\theta}L_V(\theta_l, \mathcal{V})\right\Vert \cos(\Theta_l)}{\left\Vert\nabla_{\theta}L_T(\theta_l, S)\right\Vert}\\
    \text{where\hspace{0.2cm}} \cos{\Theta_l} = \frac{\nabla_{\theta}L_T(\theta_l, S)^T \cdot \nabla_{\theta}L_V(\theta_l, \mathcal{V})}{\left\Vert\nabla_{\theta}L_T(\theta_l, S)\right\Vert\left\Vert\nabla_{\theta}L_V(\theta_l, \mathcal{V})\right\Vert}
\end{aligned}
\end{equation}

Since, we know that the gradient norm $\left\Vert\nabla_{\theta}L_T(\theta_l, S)\right\Vert \leq \sigma_T$, the condition for the learning rate can be written as follows,
\begin{equation}
\begin{aligned}
\label{step-size-cndtn}
    \alpha \leq \frac{2\left\Vert \nabla_{\theta}L_V(\theta_l, \mathcal{V})\right\Vert \cos(\Theta_l)}{L \sigma_T}\\
    \text{where\hspace{0.2cm}} \cos{\Theta_l} = \frac{\nabla_{\theta}L_T(\theta_l, S)^T \cdot \nabla_{\theta}L_V(\theta_l, \mathcal{V})}{\left\Vert\nabla_{\theta}L_T(\theta_l, S)\right\Vert\left\Vert\nabla_{\theta}L_V(\theta_l, \mathcal{V})\right\Vert}
\end{aligned}
\end{equation}

Since, the condition mentioned in Eq:\eqref{step-size-cndtn} needs to be true for all values of $l$, we have the condition for learning rate as follows:
\begin{equation}
\begin{aligned}
\label{final-step-size-cndtn}
    \alpha \leq \min_l \frac{2\left\Vert \nabla_{\theta}L_V(\theta_l, \mathcal{V})\right\Vert \cos(\Theta_l)}{L \sigma_T}\\
    \text{where\hspace{0.2cm}} \cos{\Theta_l} = \frac{\nabla_{\theta}L_T(\theta_l, S)^T \cdot \nabla_{\theta}L_V(\theta_l, \mathcal{V})}{\left\Vert\nabla_{\theta}L_T(\theta_l, S)\right\Vert\left\Vert\nabla_{\theta}L_V(\theta_l, \mathcal{V})\right\Vert}
\end{aligned}
\end{equation}
\end{proof}

\subsection{Convergence Result}
Next, we prove the convergence result.
\begin{theorem*}
Assume that the validation and subset training losses satisfy the conditions that $\nabla_{\theta}L_V(\theta_{l}, {\mathcal V})^{T} \nabla_{\theta}L_T(\theta_{l}, S) \geq 0$ for $0 \leq l \leq T$, and for all the subsets encountered during the training. Also, assume that $\delta_{\min} = \min_l \frac{\left\Vert\nabla L_T(\theta_t)\right\Vert}{\sigma_G}$ and the parameters satisfy $||\theta|| \leq \frac{R}{2}$. Then, with a learning rate set to $\alpha = \frac{R}{\sigma_T\sqrt{T}}$, the following convergence result holds:
\begin{align*}
   &\min_l L_V(\theta_l) - L_V(\theta^{*}) \leq  \frac{R\sigma_T}{\delta_{\min}\sqrt{T}} + \frac{R\sigma_T\sum_{l=0}^{T}\sqrt{1 -\cos{\Theta_{l}}}}{T\delta_{\min}} \\
    &\text{where\hspace{0.5cm}} \cos{\Theta_l} = \frac{\nabla_{\theta}L_T(\theta_l)^T \cdot \nabla_{\theta}L_V(\theta_l)}{\left\Vert\nabla_{\theta}L_T(\theta_l)\right\Vert\left\Vert\nabla_{\theta}L_V(\theta_l)\right\Vert}\text{\hspace{0.5cm}} 
\end{align*}
\end{theorem*}
\begin{proof}
Suppose the validation loss $L_V$ and training loss $L_T$ be lipschitz smooth and the gradients of validation loss and training loss are sigma bounded by $\sigma_V$ and $\sigma_T$ respectively. Let $\theta_l$ be the model parameters at epoch $l$ and $\theta^*$ be the optimal model parameters.
From the definition of Gradient Descent, we have:
\begin{equation}
\begin{aligned}
{\nabla_{\theta}L_T(\theta_l)}^T(\theta_l - \theta^*) = \frac{1}{\alpha}{(\theta_l - \theta_{l+1})}^T(\theta_l - \theta^*)
\end{aligned}
\end{equation}
\begin{equation}
\begin{aligned}
{\nabla_{\theta}L_T(\theta_l)}^T(\theta_l - \theta^*) = \frac{1}{2\alpha}\left({\left\Vert\theta_l - \theta_{l+1}\right\Vert}^2 + {\left\Vert\theta_{l} - \theta^{*}\right\Vert}^2 - {\left\Vert\theta_{l+1} - \theta^{*}\right\Vert}^2\right)
\end{aligned}
\end{equation}

\begin{equation}
\begin{aligned}
\label{eq-24}
{\nabla_{\theta}L_T(\theta_l)}^T(\theta_l - \theta^*) = \frac{1}{2\alpha}\left({\left\Vert \alpha \nabla_{\theta}L_T(\theta_l)\right\Vert}^2 + {\left\Vert\theta_{l} - \theta^{*}\right\Vert}^2 - {\left\Vert\theta_{l+1} - \theta^{*}\right\Vert}^2 \right)
\end{aligned}
\end{equation}

We can rewrite the function ${\nabla_{\theta}L_T(\theta_l)}^T(\theta_l - \theta^*)$ as follows:
\begin{equation}
\begin{aligned}
\label{eq-25}
{\nabla_{\theta}L_T(\theta_l)}^T(\theta_l - \theta^*) = {\nabla_{\theta}L_T(\theta_l)}^T(\theta_l - \theta^*) -\delta_l{\nabla_{\theta}L_V(\theta_l)}^T(\theta_l - \theta^*) + \delta_l{\nabla_{\theta}L_V(\theta_l)}^T(\theta_l - \theta^*)
\end{aligned}
\end{equation}

Combining the equations Eq:\eqref{eq-24} ,Eq:\eqref{eq-25} we have,
\begin{equation}
\begin{aligned}
{\nabla_{\theta}L_T(\theta_l)}^T(\theta_l - \theta^*) -\delta_l{\nabla_{\theta}L_V(\theta_l)}^T(\theta_l - \theta^*) + \delta_l{\nabla_{\theta}L_V(\theta_l)}^T(\theta_l - \theta^*) = \frac{1}{2\alpha}\left({\left\Vert \alpha \nabla_{\theta}L_T(\theta_l)\right\Vert}^2 + {\left\Vert\theta_{l} - \theta^{*}\right\Vert}^2 - {\left\Vert\theta_{l+1}- \theta^{*}\right\Vert}^2\right)
\end{aligned}
\end{equation}

\begin{equation}
\begin{aligned}
\label{gd-cndtn}
{\nabla_{\theta}L_V(\theta_l)}^T(\theta_l - \theta^*) = \frac{1}{2\alpha\delta_l}({\left\Vert \alpha \nabla_{\theta}L_T(\theta_l)\right\Vert}^2 + {\left\Vert\theta_{l} - \theta^{*}\right\Vert}^2 - {\left\Vert\theta_{l+1} - \theta^{*}\right\Vert}^2) - \frac{1}{\delta_l}{\left(\nabla_{\theta}L_T(\theta_l) - \delta_l\nabla_{\theta}L_V(\theta_l)\right)}^T(\theta_l - \theta^*) 
\end{aligned}
\end{equation}

Assuming $\delta_{min} = \min_l \delta_l$ and summing up the Eq:\eqref{gd-cndtn} for different values of $l \in [1, T-1]$ we have,
\begin{equation}
\begin{aligned}
\sum_{l=1}^{T-1}{\nabla_{\theta}L_V(\theta_l)}^T(\theta_l - \theta^*) \leq \frac{{\left\Vert\theta_{0} - \theta^{*}\right\Vert}^2 - {\left\Vert\theta_{T} - \theta^{*}\right\Vert}^2}{\delta_{min}} +\\ \sum_{l=1}^{T-1}(\frac{1}{2\alpha\delta_l}({\left\Vert \alpha \nabla_{\theta}L_T(\theta_l)\right\Vert}^2)) + \sum_{l=1}^{T-1}\left(\frac{1}{\delta_l}{\left(\nabla_{\theta}L_T(\theta_l) - \delta_l \nabla_{\theta}L_V(\theta_l)\right)}^T(\theta_l - \theta^*) \right)
\end{aligned}
\end{equation}

Since ${\left\Vert\theta_{T} - \theta^{*}\right\Vert}^2 \geq 0$, we have:
\begin{equation}
\begin{aligned}
\sum_{l=1}^{T-1}{\nabla_{\theta}L_V(\theta_l)}^T(\theta_l - \theta^*) \leq \frac{{\left\Vert\theta_{0} - \theta^{*}\right\Vert}^2}{\delta_{min}} + \sum_{l=1}^{T-1}(\frac{1}{2\alpha\delta_l}({\left\Vert \alpha \nabla_{\theta}L_T(\theta_l)\right\Vert}^2)) + \sum_{l=1}^{T-1}\left(\frac{1}{\delta_l}{\left(\nabla_{\theta}L_T(\theta_l) - \delta_l \nabla_{\theta}L_V(\theta_l)\right)}^T(\theta_l - \theta^*) \right)
\end{aligned}
\end{equation}

We know that $L_V(\theta_l) - L_V(\theta^*) \leq {\nabla_{\theta}L_V(\theta_l)}^T(\theta_l - \theta^*)$ from convexity of function $L_V(\theta)$. Combining this with above equation we have,
\begin{equation}
\begin{aligned}
\sum_{l=1}^{T-1}L_V(\theta_l) - L_V(\theta^*) \leq \frac{{\left\Vert\theta_{0} - \theta^{*}\right\Vert}^2}{\delta_{min}} + \sum_{l=1}^{T-1}(\frac{1}{2\alpha\delta_l}({\left\Vert \alpha \nabla_{\theta}L_T(\theta_l)\right\Vert}^2)) + \sum_{l=1}^{T-1}\left(\frac{1}{\delta_l}{\left(\nabla_{\theta}L_T(\theta_l) - \delta_l \nabla_{\theta}L_V(\theta_l)\right)}^T(\theta_l - \theta^*) \right)
\end{aligned}
\end{equation}

Since, $\left\Vert L_T(\theta)\right\Vert \leq \sigma_T$ and assuming that $\left\Vert\theta - \theta^{*}\right\Vert \leq R$, we have,
\begin{equation}
\begin{aligned}
\sum_{l=1}^{T-1}L_V(\theta_l) - L_V(\theta^*) \leq \frac{\alpha\sigma_T^2T}{2\delta_{min}} + \frac{R^2}{2\alpha\delta_{min}} + \sum_{l=1}^{T-1}\left(\frac{1}{\delta_l}{\left\Vert\nabla_{\theta}L_T(\theta_l) - \delta_l \nabla_{\theta}L_V(\theta_l)\right\Vert}R\right)
\end{aligned}
\end{equation}

\begin{equation}
\begin{aligned}
\label{avg_loss}
\frac{\sum_{l=1}^{T-1}L_V(\theta_l) - L_V(\theta^*)}{T} \leq \frac{\alpha\sigma_T^2}{2\delta_{min}} + \frac{R^2}{2\alpha\delta_{min}T} + \sum_{l=1}^{T-1}\left(\frac{1}{\delta_l T}{\left\Vert\nabla_{\theta}L_T(\theta_l) - \delta_l \nabla_{\theta}L_V(\theta_l)\right\Vert}R\right)
\end{aligned}
\end{equation}

Assuming $\delta_l = \frac{\left\Vert\nabla_{\theta_l}L_T(\theta_l)\right\Vert}{\left\Vert\nabla_{\theta_l}L_V(\theta_l)\right\Vert}$, we can write as follows:
\begin{equation}
\begin{aligned}
\label{perp_eq}
\delta_l\nabla_{\theta}L_V(\theta_l) = \nabla_{\theta}L_T(\theta_l)\cos(\Theta_l) + \widehat{\nabla_{\theta}L_T(\theta_l)}\sin(\Theta_l) \text{\hspace{0.2cm}where\hspace{0.2cm}}\widehat{\nabla_{\theta}L_T(\theta_l)} \bot \nabla_{\theta}L_T(\theta_l) \\\text{where\hspace{0.2cm}}\cos(\Theta_l) = \frac{\nabla_{\theta}L_T(\theta_l)^T \cdot \nabla_{\theta}L_V(\theta_l)}{\left\Vert\nabla_{\theta}L_T(\theta_l)\right\Vert\left\Vert\nabla_{\theta}L_V(\theta_l)\right\Vert}
\end{aligned}
\end{equation}

Combining Eq:\eqref{avg_loss} and Eq:\eqref{perp_eq}, we have:
\begin{equation}
\begin{aligned}
\frac{\sum_{l=1}^{T-1}L_V(\theta_l) - L_V(\theta^*)}{T} \leq \frac{\alpha\sigma_T^2}{2\delta_{min}} + \frac{R^2}{2\alpha\delta_{min}T} + \sum_{l=1}^{T-1}\left(\frac{1}{\delta_l T}{\left\Vert\nabla_{\theta}L_T(\theta_l) (1 - \cos(\Theta_l)) + \widehat{\nabla_{\theta}L_T(\theta_l)}\sin(\Theta_l)\right\Vert}R\right)
\end{aligned}
\end{equation}

Since $\left\Vert\nabla_{\theta}L_T(\theta_l) (1 - \cos(\Theta_l)) + \widehat{\nabla_{\theta}L_T(\theta_l)}\sin(\Theta_l)\right\Vert \leq \left\Vert\nabla_{\theta}L_T(\theta_l)\right\Vert\sqrt{1-\cos(\Theta_l)}$, we have:
\begin{equation}
\begin{aligned}
\frac{\sum_{l=1}^{T-1}L_V(\theta_l) - L_V(\theta^*)}{T} \leq \frac{\alpha\sigma_T^2}{2\delta_{min}} + \frac{R^2}{2\alpha\delta_{min}T} + \sum_{l=1}^{T-1}\frac{R}{\delta_l T}\left(\left\Vert\nabla_{\theta}L_T(\theta_l)\right\Vert\sqrt{1-\cos(\Theta_l)}\right)
\end{aligned}
\end{equation}

Since $\left\Vert\nabla_{\theta}L_T(\theta)\right\Vert \leq \sigma_T$, we have:
\begin{equation}
\begin{aligned}
\frac{\sum_{l=1}^{T-1}L_V(\theta_l) - L_V(\theta^*)}{T} \leq \frac{\alpha\sigma_T^2}{2\delta_{min}} + \frac{R^2}{2\alpha\delta_{min}T} + \sum_{l=1}^{T-1}\frac{R\sigma_T}{\delta_l T}\sqrt{1-\cos(\Theta_l)}
\end{aligned}
\end{equation}

Choosing $\alpha = \frac{R}{\sigma_T\sqrt{T}}$, we have:
\begin{equation}
\begin{aligned}
\frac{\sum_{l=1}^{T-1}L_V(\theta_l) - L_V(\theta^*)}{T} \leq \frac{R\sigma_T}{\sqrt{T}\delta_{min}} + \frac{R\sigma_T\sum_{l=1}^{T-1}\sqrt{1-\cos(\Theta_l)}}{\delta_{min} T}
\end{aligned}
\end{equation}

Since, $\min_{l}(L_V(\theta_l) - L_V(\theta^*)) \leq \frac{\sum_{l=1}^{T-1}L_V(\theta_l) - L_V(\theta^*)}{T}$, we have:
\begin{equation}
\begin{aligned}
 \min_{l}(L_V(\theta_l) - L_V(\theta^*)) \leq \frac{R\sigma_T}{\delta_{min}\sqrt{T}} + \frac{R\sigma_T\sum_{l=1}^{T-1}\sqrt{1-\cos(\Theta_l)}}{T \delta_{min}}
\end{aligned}
\end{equation}

Therefore, we conclude that our algorithm will converge in $\mathcal{O}(1/ \sqrt{T})$ steps when the validation data and training data are from similar distributions since $\cos(\Theta_l) \approx 1$.
\end{proof}

\section{Derivation of Closed Form Expressions for Special Cases}
In this section, we derive the closed form expressions of the special cases discussed in section~\ref{prob-form}. 

\subsection{Discrete Naive Bayes}

We consider the case of discrete naive bayes model and see how the log 
likelihood function for the training and validation look like in terms
of parameters obtained from a subset of training data. 

\noindent \textbf{Some notation:} The training data set is ${\mathcal U} = \{ (x^i, y^i) \}_{i=1:m}$ with
each \(x^i\) being a d-dimensional vector with values from the set
$\mathcal{X}$. Similarly each label $y^i \in \mathcal{Y}$ and $\mathcal{Y}$ is a 
finite label set. We can think of ${\mathcal U}$ as being partitioned on the basis of labels:
${\mathcal U} = \cup_{y \in {\mathcal Y}} {\mathcal U}^y$ where ${\mathcal U}^y$ is the set of data points
whose label = \(y\). We calculate the (MLE) parameters \(\theta(S)\) of the model on
subset $S \subseteq {\mathcal U}$. These involve
the conditional probabilities \(\theta_{x_j | y} (S)\) for a feature \(x_j\) (along each 
dimension $j=1..d$) given a class label \(y\) and the
prior probabilities for each class label \(\theta_y (S)\)

Let \(m_{x_j,y}(S) = \sum_{i \in S} 1[x^i_j = x_j \wedge y_j = y]\) and
\(m_y(S) = \sum_{i \in S} 1[y_j = y]\), then we will have:
\(\theta_{x_j | y} (S) = \frac{m_{x_j,y(S)}}{m_y(S)}\) and, 
\(\theta_y (S) = \frac{m_y(S)}{|S|}\)

\subsubsection{Training data log-likelihood}\label{NB-training-data-log-llk}
This has been derived in~\cite{wei2015submodularity}, but for completeness, we add it here. Using the above notation we write log-likelihood of the training data set.
In some places we denote \(\theta(S)\) by \(\theta_S\) for notational convenience.

\begin{align}
    l^{NB}(S) &= \sum_{i \in {\mathcal U}} \log p(x^i | y^i ; \theta_S) + \log p(y^i ; \theta_S) \nonumber \\
    &= \sum_{i \in {\mathcal U}} \sum_{j=1:d} \log \theta_{x_j | y} (S) + \sum_{i \in {\mathcal U}} \log  \theta_y (S) \nonumber \\
    &= \sum_{i \in {\mathcal U}} \sum_{j=1:d} \log \frac{m_{x_j,y(S)}}{m_y(S)} + \sum_{i \in {\mathcal U}} \log\frac{m_y(S)}{|S|} \nonumber \\
    &= \sum_{j=1:d} \sum_{x_j \in \mathcal{X}} \sum_{y \in \mathcal{Y}} m_{x_j,y}({\mathcal U}) \log m_{x_j,y}(S) 
- (d-1) \sum_{y \in \mathcal{Y}} m_y({\mathcal U}) \log m_y (S) - |{\mathcal U}| \log |S|
\end{align}

Note that we change the sum over ${\mathcal U}$ into a sum over all possible
combinations of \(x, y\) and just multiply each pair's (feature's) count in ${\mathcal U}$
appropriately. Also we separate the above sum into three terms as
follows for analysis of the log likelihood as a function of the subset \(S\).

\begin{itemize}
\item
  term1:
  \(f_{NB} (S) = \sum_{j=1:d} \sum_{x_j \in \mathcal{X}} \sum_{y \in \mathcal{Y}} m_{x_j,y}({\mathcal U}) \log m_{x_j,y}(S)\)
\item
  term2: \((d-1) \sum_{y \in \mathcal{Y}} m_y({\mathcal U}) \log m_y (S)\)
\item
  term3: \(|{\mathcal U}| \log |S|\)
\end{itemize}

The overall log-likelihood as a summation of the three terms is a difference 
of submodular functions over
\(S\) (since parts corresponding to ${\mathcal U}$ are constants). Also when we enforce
\(|S| = k\), then term3 is a constant and when we make the assumption
that \(S\) is \textbf{balanced} {\em i.e}, the distribution over class labels in
\(S\) is same as that of ${\mathcal U}$ which means
\(|S \cap {\mathcal U}^y| = k \frac{|{\mathcal U}^y|}{|{\mathcal U}|}\) with \(|S| = k\). This makes
term2 as constant since \(m_y (S) = |S \cap {\mathcal U}^y|\) by definition. As a result, we have the following result:
\begin{lemma}
Optimizing $l^{NB}_U(S)$ is equivalent to optimizing $f_{NB_U} (S)$ under the constraint that \(|S \cap {\mathcal U}^y| = k \frac{|{\mathcal U}^y|}{|{\mathcal U}|}\) with \(|S| = k\), which is essentially a partition matroid constraint.
\end{lemma}
Note that this is an example of a
\textbf{feature based} submodular function since the form is: $F(X) = \sum_{f \in F} w_f g(m_f(X))$ where $m_f(X) = \sum_{i \in X} m_i$ is a modular function, 
$g$ is a concave function (for us its $\log$) and $w_f$ is a non-negative weight
associated with feature $f$ from the overall feature space $F$.

\subsubsection{Validation data
log-likelihood}\label{nb-validation-data-log-likelihood}
In this section we compute the log-likelihood on validation set ${\mathcal V}$ but the parameters 
still come from the subset $S$ of training data ${\mathcal U}$. So take note of the subtle change in the equations. In this case as well, the objective function is submodular, under very similar assumptions. We first start by deriving the log-likelihood on the validation set. 
\begin{align}
   l^{NB}_V (S) &= \sum_{i \in {\mathcal V}} \log p(x^i | y^i ; \theta_S) + \log p(y^i ; \theta_S) \nonumber \\
   &= \sum_{i \in {\mathcal V}} \sum_{j=1:d} \log \theta_{x_j | y} (S) + \sum_{i \in {\mathcal V}} \log  \theta_y (S) \nonumber \\
   &= \sum_{i \in {\mathcal V}} \sum_{j=1:d} \log \frac{m_{x_j,y(S)}}{m_y(S)} + \sum_{i \in {\mathcal V}} \log\frac{m_y(S)}{|S|} \nonumber \\ 
   & = \sum_{j=1:d} \sum_{x_j \in \mathcal{X}} \sum_{y \in \mathcal{Y}} m_{x_j,y}({\mathcal V}) \log m_{x_j,y}(S) 
- (d-1) \sum_{y \in \mathcal{Y}} m_y({\mathcal V}) \log m_y (S) - |{\mathcal V}| \log |S|
\end{align}

For the second and third term to be constant, we need to make very similar assumptions. In particular, we assume that for all $y \in \mathcal Y$, $|S \cap {\mathcal V}^y| = k\frac{|{\mathcal V}^y|}{|{\mathcal V}|}$. As a result, the second term and third terms are constant and we get an instance of a feature based function:
 
$$  
f_{NB}^V (S) = \sum_{j=1:d} \sum_{x_j \in \mathcal{X}} \sum_{y \in \mathcal{Y}} m_{x_j,y}({\mathcal V}) \log m_{x_j,y}(S) 
$$
\begin{lemma}
Optimizing $l^{NB}(S)$ is equivalent to optimizing $f_{NB} (S)$ under the constraint that \(|S \cap {\mathcal U}^y| = k \frac{|{\mathcal U}^y|}{|{\mathcal U}|}\) with \(|S| = k\), which is essentially a partition matroid constraint.
\end{lemma}
It is also worth pointing out that the assumptions made are very intuitive. Since we want the functions to generalize to the validation set, we want the training set distribution (induced through the set $S$) to be able to match the validation set distribution in terms of the class label distribution. Also, intuitively, the feature based function attempts to match the distribution of $m_{x_j, y}(S)$ to the distribution of $m_{x_j, y}({\mathcal V})$~\cite{wei2014submodular,wei2015submodularity}, which is also very intuitive.

\subsection{Nearest Neighbors}
Next, we consider the nearest neighbor model with a similarity function 
\(w\) which takes in two feature vectors and outputs a positive similarity score: \(w(i,j) = d - ||x^i - x^j ||^2_2\) where
\(d = max_{v, v' \in {\mathcal U}} ||x^v - x^{v'} ||^2_2\). As is the case in previous 
sections, the model parameters come from a subset of training data.
(although it is a non-parametric model, so think
of the parameters as being equivalent to the subset itself).
In order to express the model in terms of log-likelihood, the generative
probability for a data point $x^i$ is written as:
\(p(x^i | y^i ; \theta_S)\) = 
\(= c \; e^{-||x^i - x^j ||^2_2}\) i.e just by a single sample
\(x^j \;\text{s.t:}\; j = \arg \max_{s \in S\cap {\mathcal U}^{y^i}} w(i,s)\)
and hence:
\(p(x^i | y^i ; \theta_S) = c' \exp(\max_{s \in S\cap {\mathcal U}^{y^i}} w(i,s))\).
The prior probabilities stay the same as before  \(p(y^i ; \theta_S) = \frac{m_y(S)}{|S|}\). 

\subsubsection{Training data log-likelihood}\label{knn-training-data-log-llk-1}
Again, here we follow the proof technique from~\cite{wei2015submodularity}. Note that the log-likelihood on the training set, given the parameters obtained from the subset $S$ are:
\begin{align}
    l^{NN}_V (S) &= \sum_{i \in {\mathcal U}} \log p(x^i | y^i ; \theta_S) + \log p(y^i ; \theta_S) \nonumber \\
    & = \sum_{y \in \mathcal{Y}} \sum_{i \in {\mathcal U}^y}\max_{s \in S\cap {\mathcal U}^{y^i}} w(i,s) 
+  \sum_{y \in \mathcal{Y}} m_y({\mathcal U}) \log m_y(S) - |{\mathcal U}| \log|S| + \mbox{const}
\end{align}

\begin{itemize}
\item
  term1 \(f_{NN}(S)\):
  \(\sum_{y \in \mathcal{Y}} \sum_{i \in {\mathcal U}^y}\max_{s \in S\cap {\mathcal U}^{y^i}} w(i,s)\)
\item
  term2: \(\sum_{y \in \mathcal{Y}} m_y({\mathcal U}) \log m_y(S)\)
\item
  term3: \(|{\mathcal U}| \log|S|\)
\end{itemize}

Similar to the Naive-Bayes, we make the assumption that the set $S$ is balanced, i.e. $|{\mathcal U}^Y \cap S| = k\frac{|{\mathcal U}^y|}{|{\mathcal U}|}$ and the size of the set $S$ is $k$. Under this assumption, terms 2 and 3 are a constant.
\begin{lemma}
Optimizing $l^{NN}(S)$ is equivalent to optimizing $f_{NN} (S)$ under the constraint that \(|S \cap {\mathcal U}^y| = k \frac{|{\mathcal U}^y|}{|{\mathcal U}|}\) with \(|S| = k\) (i.e. a partition matroid constraint).
\end{lemma}

\subsubsection{Validation data log-likelihood}\label{knn-validation-data-log-llk}
Along similar lines, we can derive the log-likelihood on the validation set (with parameters obtained from the training set). 
\begin{align}
    l^{NN}_V (S) &= \sum_{i \in {\mathcal V}} \log p(x^i | y^i ; \theta_S) + \log p(y^i ; \theta_S) \nonumber \\
    &= \sum_{y \in \mathcal{Y}} \sum_{i \in {\mathcal V}^y}\max_{s \in S\cap {\mathcal V}^{y^i}} w(i,s) 
+  \sum_{y \in \mathcal{Y}} m_y({\mathcal V}) \log m_y(S) - |{\mathcal V}| \log|S| + C
\end{align}

Similar to the NB case, the second and third case are constants if we assume that for all $y \in \mathcal Y$, $|S \cap {\mathcal V}^y| = k\frac{|{\mathcal V}^y|}{|{\mathcal V}|}$. Hence, optimizing $l^{NN}_V(S)$ is the same as optimizing $f^{NN}_V(S) = \sum_{y \in \mathcal{Y}} \sum_{i \in {\mathcal V}^y}\max_{s \in S\cap {\mathcal V}^{y^i}} w(i,s)$, which is an instance of facility location. 
\begin{lemma}
Optimizing $l^{NN}(S)$ is equivalent to optimizing $f_{NN} (S)$ under the constraint that \(|S \cap {\mathcal V}^y| = k \frac{|{\mathcal V}^y|}{|{\mathcal V}|}\) with \(|S| = k\) ({\em i.e.}, a partition matroid constraint).
\end{lemma}

\subsection{Linear Regression}
In the case of linear regression, denote \({\mathcal U}, |{\mathcal U}|=N\) as the training set comprising of pairs \(x^i \in \mathbf{R}^d\) and \(y^i \in \mathbf{R}\). Denote \(X_U\) as the $N \times d$ data matrix and \(Y_U\) to be the $N \times 1$ vector of values. The goal of linear regression is to find the solution \(\theta \in \mathbf{R}^d\)  such that it minimizes the squared loss on the training data. The closed form solution of this is $\theta = (X_U^T X_U)^{-1} X_U^T Y_U$

Let $(x_i, y_i)$ be the $i^{th}$ data-point in the validation set $\mathcal{V}$ where $i \in [1, |\mathcal{V}|]$ and $(x_i^t, y_i^t)$ be the $i^{th}$ data-point in the training set $\mathcal{U}$ where $i \in [1, |\mathcal{U}|]$.

Now, given a subset of the training dataset, \(S \subseteq {\mathcal U}\) (with \(|S| = K\)), we can obtain parameters $\theta_S$ on $S$, with:
\(\theta(S) = (X_S^T X_S)^{-1} X_S^T Y_S\). Here, \(X_S\) is $K \times d$ matrix and 
\(Y_S\) is the $K \times 1$ dimensional vector of values corresponding to the
rows selected. Similarly, denote our validation set as a 
matrix \(X_V \in \mathbf{R}^{M \times d}\) and \(Y_V \in \mathbf{R}^M\) assuming \(|{\mathcal V}|=M\).

Hence
the subset selection problem with respect to \textbf{validation data ${\mathcal V}$} now becomes:

\begin{align}
   &\mbox{argmax}_{S \subseteq {\mathcal V}, |S|=k} -\left\Vert Y_V - X_V \theta(S) \right\Vert^2 = \mbox{argmax}_{S \subseteq {\mathcal V}, |S|=k} -\left\Vert Y_V - X_V  (X_S^T X_S)^{-1} X_S^T Y_S \right\Vert^2
\end{align}
We denote this as $l_V^{LR}(S) = -\left\Vert Y_V - X_V  (X_S^T X_S)^{-1} X_S^T Y_S \right\Vert^2$. Also, note that \( X_S^T X_S = \sum_{i \in S} x_i x_i^T\) where \(x_i\) is the
individual $i$-th data vector. 

Next, assume that we cluster the original training dataset into $m$ clusters $C_1, C_2, ...., C_m$ and the clusters' centroids are $\{x_{c1}, x_{c2}, ..., x_{cm}\}$ respectively. We then represent each data point $(x_i, y_i)$ in training set $\mathcal{U}$ by the respective centroid of its cluster. We also assume that our subset selected $S$ preserves each cluster's original proportion of data points. i.e., $\forall i \frac{|S \cap C_{i}|}{|S|} = \frac{|C_i|}{N}$.

Because of the above assumption, we can write $X_S^T X_S \approx \sum_{i=1}^{m}\frac{|C_i|}{N}x_{ci}x_{ci}^T$. This assumption ensures that we can precompute the term ${X_S^TX_S}^{-1}$. Let's denote the precomputed matrix as $D = {X_S^TX_S}^{-1}$.

Hence we can approximate the subset selection problem (i.e. optimizing $l_V^{LR}(S)$) as:
\begin{align}
  \mbox{argmax}_{S \in \mathcal C} -\left\Vert Y_V - X_VDX_S^T Y_S \right\Vert^2
\end{align}
where $\mathcal C$ refers to the constraint $\mathcal C = \{S \subseteq \mathcal V: |S| = K, \forall {i} \frac{|S \cap C_{i}|}{|S|} = \frac{|C_i|}{N}$. For convenience, we denote this set function as $G(S)$:
\begin{align}
        G(S) &= -\left\Vert Y_V - X_VDX_S^T Y_S\right\Vert^2 \\
        &= -\left\Vert Y_V \right\Vert^2 + 2Y_V^T(X_V D X_S^T Y_S) - {(X_V D X_S^T Y_S)}^T(X_V D X_S^T Y_S) \\\
        &= -\left\Vert Y_V \right\Vert^2 + 2Y_V^T(X_V D X_S^T Y_S) - {(X_V D X_S^T Y_S)}^T(X_V D X_S^T Y_S) \\
        &= -\left\Vert Y_V \right\Vert^2 + \underset{i \in S}{\sum}\underset{j \in \mathcal{V}}{\sum}2(y_j x_j^T D) (x_i^t y_i^t)  - \underset{i \in S}{\sum}\underset{j \in S}{\sum}\underset{k \in \mathcal{V}}{\sum}(x_k^T D x_i^t y_i^t)(x_k^T D x_j^t y_j^t) 
\end{align}

Next, define $s_{ij} = \underset{k \in \mathcal{V}}{\sum}(x_k^T D x_i^t y_i^t)(x_k^T D x_j^t y_j^t)$,
\begin{equation}
    \begin{aligned}
        G(S) = -\left\Vert Y_V \right\Vert^2 + \underset{i \in S}{\sum}\underset{j \in \mathcal{V}}{\sum}2(y_j x_j^T D) (x_i^t y_i^t)  - \underset{i \in S}{\sum}\underset{j \in S}{\sum}s_{ij}
    \end{aligned}
\end{equation}

Since, $s_{ij}$ is not always greater than zero, we can transform it to $\hat{s}_{ij}$ such that ${s}_{ij} = \hat{s}_{ij} + s_{min}$ where $s_{min} = \underset{i,j}{\min}s_{ij}$. This transformation ensures that $\hat{s}_{ij} \geq 0$.
\begin{equation}
    \begin{aligned}
        G(S) = -\left\Vert Y_V \right\Vert^2 + \underset{i \in S}{\sum}\underset{j \in \mathcal{V}}{\sum}2(y_j x_j^T D) (x_i^t y_i^t)  - \underset{i \in S}{\sum}\underset{j \in S}{\sum}(\hat{s}_{ij} + s_{min})
    \end{aligned}
\end{equation}

\begin{equation}
    \begin{aligned}
        G(S) = -\left\Vert Y_V \right\Vert^2 + \underset{i \in S}{\sum}\underset{j \in \mathcal{V}}{\sum}2(y_j x_j^T D) (x_i^t y_i^t)  - \underset{i \in S}{\sum}\underset{j \in S}{\sum}\hat{s}_{ij}  - |S|^2 (s_{min})
    \end{aligned}
\end{equation}
Since we are considering an subset selection of selecting an subset $S$ where $|S| = K$, we convert our set function $G(S)$ to the proxy function $\hat{G}(S)$:
\begin{equation}
    \begin{aligned}
        \hat{G}(S) = -\left\Vert Y_V \right\Vert^2 - K^2 (s_{min}) + \underset{i \in S}{\sum}\underset{j \in \mathcal{V}}{\sum}2(y_j x_j^T D) (x_i^t y_i^t)  - \underset{i \in S}{\sum}\underset{j \in S}{\sum}\hat{s}_{ij}  
    \end{aligned}
\end{equation}

In the above equation:
\begin{itemize}
    \item The first part $-\left\Vert Y_V \right\Vert^2 - K^2 (s_{min})$ is constant and does not depend on subset $S$. So, this term does not affect our optimization problem.
    \item The second part $\underset{i \in S}{\sum}\underset{j \in \mathcal{V}}{\sum}2(y_j x_j^T D) (x_i^t y_i^t)$ is a non-monotone modular function in subset $S$.
    \item Note that the third part $- \underset{i \in S}{\sum}\underset{j \in S}{\sum}\hat{s}_{ij}$ is similar to the graph cut function since $\hat{s}_{jk} \geq 0$. Hence it is a submodular function.
\end{itemize}
Therefore, the proxy set function $\hat{G}(S)$ is a non-monotone submodular function. Finally, we define the linear regression submodular function:
\begin{equation}
    \begin{aligned}
        F_{LR}(S) = \underset{i \in S}{\sum}\underset{j \in \mathcal{V}}{\sum}2(y_j x_j^T D) (x_i^t y_i^t)  - \underset{i \in S}{\sum}\underset{j \in S}{\sum}\hat{s}_{ij}  
    \end{aligned}
\end{equation}
Finally, we summarize our reductions as follows. Optimizing $F_{LR}(S)$ is the same as optimizing $\hat{G}(S)$, which is equivalent to optimizing $G(S)$ under cardinality equality constraints $|S| = k$. Finally, this means that optimizing $F_{LR}(S)$ subject to the constraints $|S| = k$ and $\forall {i} \frac{|S \cap C_{i}|}{|S|} = \frac{|C_i|}{N}$ is an \emph{approximation} of optimizing the negative linear regression loss $l_V^{LR}(S)$ subject to cardinality equality constraints $|S| = K$. Note the reason for the approximation is the clustering performed so as to approximate the $X_S^T X_S$. Since $F_{LR}(S)$ is a non-monotone submodular function, the resulting problem involves optimizing a non-monotone submodular function subject to matroid base constraints, for which efficient algorithms exist~\cite{lee2009non}.

\section{Additional Details on GreedyDSS}
\begin{algorithm}[ht]
  \caption{GreedyTaylorApprox(${\mathcal U}$, ${\mathcal V}$, $\theta^{0}$, $\eta$, $k$, $r$, $\lambda$, $R$)}
  \label{alg:algorithm2}
  \begin{algorithmic}[1]
    \STATE Initialize $S = \emptyset$, $U = {\mathcal U}$, $t = 0$
    \WHILE {$t < r$}
        \FORALL {$e\in U$} 
            \STATE Set $\theta^{(t)}_e = \theta^{(t)} + \eta \nabla_{\theta}LL_{T}(e, \theta)|_{\theta^{(t)}}$
           \STATE \begin{varwidth}{\linewidth}Set $\hat{G}_{\theta}(e) = G_{\theta_e}(e | S^k) + \lambda R(e | S^k)$;\end{varwidth}
        \ENDFOR    
        \STATE  $S_t = \underset{X \subseteq U;|X| = k/r}{\operatorname{argmax}} \sum_{e \in S} \hat{G}_{\theta}(e)$
        \STATE $S = S \cup S_t, U = U \backslash S_t$
        \STATE Update $\theta^{(t+1)} = \theta^{(t)} + \underset{e \in S_t}{\sum}\nabla_{\theta}LL_{T}(e, \theta)|_{\theta^{(t)}}$; 
        \STATE $t = t+1$
    \ENDWHILE
    \STATE Return $S$
  \end{algorithmic}
\end{algorithm} 

\section{Additional Details on \modelactive\ }
\modelactive\  performs mini-batch adaptive active learning, where we select a batch of $B$ samples to be labeled for $T$ rounds. As any active learning algorithm, \modelactive\  selects a subset of size $B$ from the pool of unlabeled instances such that the subset selected has as much information as possible to help the model come up with appropriate decision boundary. \modelactive\ picks those points which reduce the validation loss the best. We have a very similar algorithm for \modelactive\ as \modelonline\  which is shown in Algorithm \ref{alg:algorithm3}. The algorithm uses $\operatorname{\text{GreedyDSS}}$ (refer to the approximations subsection in the main paper) just like Algorithm \ref{alg:algorithm1} for \modelonline\ . However, here we pass only the unlabeled pool samples with the hypothesised lables generated by the model after training with the labled samples instead of full training set as in Algorithm \ref{alg:algorithm1}. Similarly, we set $k$ (budget) as $B$. We don’t retrain the model from scratch after each round $t \in {1,2,..,T}$ rather continue training the model as new samples are selected based on a criterion linked to the previously trained model. Finally, in our experiments, we assume we have access to a very small labeled validation set, and this is particularly useful in scenarios such as distribution shift. Note that in principle, our \modelactive{} can also be used to consider the training set, and in particular, using the hypothesized labels in the training. In other words, we can replace $\mathcal V$ in Line 4 of Algorithm \ref{alg:algorithm3} with $\hat{\mathcal U}$, which is the hypothesized labels obtained by the current model $\theta^{(t)}$.

\begin{algorithm}[h]
\caption{Active Learning Framework}
\label{alg:algorithm3}
\begin{algorithmic}[1]
\REQUIRE Input: Unlabeled data ${\mathcal U}$, Validation data ${\mathcal V}$, Labeled Data ${\mathcal L}$, $\theta^{(0)}$ model parameters initialization 
\REQUIRE $\eta$: learning rate. $T$: number of rounds, $E$: number epochs, $B$: number of points selected each round, $r$: number of times to perform taylor approximation during the greedy, $\lambda= $ Regularization coefficient
\FOR { $t = 1, \cdots, T$}
    \STATE Train the deep model using Labeled set $\mathcal L$, by running the model for $E$ epochs with learning rate $\eta$.
    \STATE Generate the hypothesised labels for the Unlabeled pool of samples ${\mathcal U}$ using the model parameters $\theta^{(t)}$
    \STATE $\hat L=\operatorname{\text{GreedyDSS}} ({\mathcal U},{\mathcal V}, \theta^{(t)},\eta,B,r,\lambda)$
    \STATE Obtain Labels on $\hat L$.
    \STATE ${\mathcal L} = {\mathcal L} \cup \hat L$, ${\mathcal U} = {\mathcal U} \backslash \hat L$.
\ENDFOR
\STATE Output the final model parameters $\theta^{(T)}$
\end{algorithmic}
\end{algorithm}

\section{Dataset Description}

The real world dataset were taken from  \textbf{LIBSVM} (a library for Support Vector Machines (SVMs)) \cite{chang2001libsvm},from \textbf{sklearn.datasets} package \cite{scikit-learn} and \textbf{UCI} machine learning repository \cite{Dua:2019}. From \textbf{LIBSVM} namely, \textbf{dna, svmguide1, letter, ijcnn1, connect-4, usps and  a9a(adult)} and from \textbf{UCI} namely, 
Census Income and Covertype were taken datasets were taken. \textbf{sklearn-digits} is from \textbf{sklearn.datasets} package \cite{scikit-learn}.  In addition to these, two standard datasets namely, MNIST and CIFAR10 are used to demonstrate effectiveness and stability of our model.

\begin{table}[H]
  \centering
  \begin{adjustbox}{width=1\textwidth}
      \begin{tabular}{|c|c|c|c|c|c|} 
     \hline 
     \textbf{Name} & \textbf{No. of classes} & \textbf{No. samples for} & \textbf{No. samples for} & \textbf{No. samples for} & \textbf{No. of features} \\  
     ~ & ~ & \textbf{training} & \textbf{validation} & \textbf{testing} & ~ \\ [0.5ex] 
     \hline
     sklearn-digits & 10 & 1797 & - & - & 64\\ 
     \hline
     dna & 3 & 1,400 & 600 & 1,186 & 180\\ 
     \hline
     satimage & 6 & 3,104 & 1,331& 2,000 & 36\\ 
     \hline
     svmguide1 & 2 & 3,089 & - & 4,000 & 4\\ 
     \hline
     usps & 10  & 7,291 & - & 2,007 & 256\\ 
     \hline
     letter & 26 & 10,500 & 4,500 & 5,000 & 16\\ 
     \hline
     \text{connect\_4} & 3 & 67,557 & - & - & 126 \\ 
     \hline
     ijcnn1 & 2  & 35,000 & 14990 & 91701 & 22\\ 
     \hline
     CIFAR10 & 10  & 50,000 & - & 10,000 & 32x32x3\\ 
     \hline
     MNIST & 10  & 60,000 & - & 10,000 & 28x28\\ 
     \hline
     \end{tabular}
 \end{adjustbox}
 \caption{Description of the datasets}
  \label{tab:d}
\end{table}

Table \ref{tab:d} gives a brief description about the datasets. Here not all datasets have a explicit validation and test set.for such datasets 10\% and 20\% samples from the training set are used as validation and test set respectively. The sizes reported for the census income dataset in the table is after removing the instances with missing values.

\begin{table}[H]
\centering
{\rowcolors{3}{white!70!black!30}{white!100!black!0}
\begin{tabular}{ |p{6cm}|p{5cm}|p{5cm}|  }
\hline
\multicolumn{3}{|c|}{Time Complexities} \\
\hline
GLISTER Approximations & Naive Greedy & Stochastic Greedy\\
\hline
No Approximation    & $\mathcal{O}(nkmFT/L + kTB)$    & $\mathcal{O}(nmFT/L \log 1/\epsilon + kTB)$\\
Last Layer Approximation& $\mathcal{O}(nkmfT/L + kTB)$  & $\mathcal{O}(nmfT/L \log 1/\epsilon + kTB)$ \\
Last Layer \& Taylor Approximations& $\mathcal{O}(k[m+n]fT/L + kTB)$ & $\mathcal{O}([km + n\log 1/\epsilon]fT/L + kTB)$\\
Last Layer, Taylor \& R approximations & $\mathcal{O}(r[m+n]fT/L + kTB)$ & $\mathcal{O}([rm + n\log 1/\epsilon]fT/L + kTB)$\\
\hline
\end{tabular}
}
 \caption{Time Complexities Table}
  \label{tab:t}
\end{table}

Table \ref{tab:t} gives a comparison of time complexities for various GLISTER approximations for both Naive Greedy and Stochastic greedy selection methods.

\section{Experimental Settings}

We ran experiments with shallow models and deep models. For shallow models, we used a two-layer fully connected neural network having 100 hidden nodes. We use simple SGD optimizer for training the model with a learning rate of 0.05. For  MNIST we used a LeNet like model ~\cite{lecun1989backpropagation} and we trained for 100 epochs,for CIFAR-10 we use ResNet-18~\cite{he2016deep} and we trained for 150 epochs. For all other datasets fully connected two layer shallow network was used and we trained for 200 epochs.

To demonstrate effectiveness of our method in robust learning setting, we artificially generate class-imbalance for the above datasets by removing 90\% of the instances from 30\% of total classes available. Whereas for noisy data sets, we flip the labels for a randomly chosen subset of the data where the noise ratio determines the subset's size. In our experimental setting, we use a 30\% noise ratio for the noisy experiments.
\subsection{Other specific settings for \modelonline}
Here we discuss various parameters defined in Algorithm~\ref{alg:algorithm1}, their significance and the values we used in the experiments.
\begin{itemize}
    \item \textbf{k} determines no. of points with which the model will be trained.
    \item \textbf{L} determines the no. of times subset selection will happen. It is set to 20 except for the experiments where we demonstrate the effect of different L values on the test accuracy. Thus, we do a subset selection every $20^{th}$ epoch. In the additional experimental details below, we also compare the effect of different values of $L$ both on performance and time.
    \item \textbf{r} determines the no. of times validation loss is recalculated by doing a complete forward pass. We demonstrate that the trade off between good test accuracy and lower training time is closely related to \textbf{r} for our method. We set \textbf{r} $\approx 0.03$\textbf{k}. In the additional experimental details below, we also compare the effect of different values of $r$ both on performance and time.
    \item \textbf{$\lambda$} determines how much regularization we want. When we use random function as a regularizing function it determines what fraction of points ( (1- \textbf{$\lambda$})k ) in the final subset selected would be randomly selected points where as when we use facility location as the regularizing function then \textbf{$\lambda$} determines how much weightage is to be given to facility location selected points' in computing the final validation loss. We use \textbf{$\lambda$} = 0.9 for Rand-Reg \modelonline\ and \textbf{$\lambda$} = 100 for Fac Loc Reg \modelonline. 
\end{itemize}

\subsection{Other specific settings for \modelactive}
Here we discuss various parameters specifically required by Algorithm~\ref{alg:algorithm3} other then the ones that are common with Algorithm~\ref{alg:algorithm1}. 
\begin{itemize}
    \item \textbf{B} just like \textbf{k} determines no. of points for which we can obtain labels in each round. 
    \item \textbf{R} it represents the total no. of round. A round comprises of selecting subset of points to be labeled and training the model with the entire with complete labeled data points. We use \textbf{R} = 10 for all our experiments.
    \item \textbf{T} is total no. of epochs we train our model in each round. We use \textbf{T} = 200 for all our experiments.
\end{itemize}

\section{Additional Experiments}

\subsection{Data Selection for Efficient Learning}

\begin{figure*}[!htbp]
    \centering
    \begin{subfigure}[b]{0.4\textwidth}
        \centering
        \includegraphics[width=\textwidth, height=3cm]{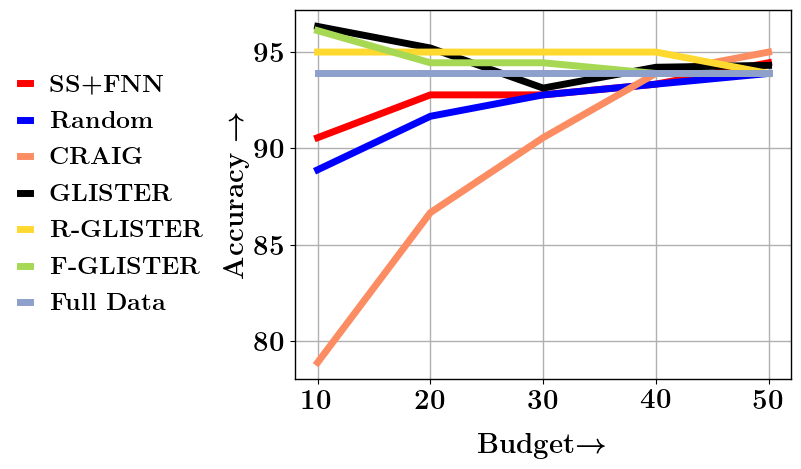}
        \subcaption{}
    \end{subfigure}
    \hfill
    \begin{subfigure}[b]{0.29\textwidth}
        \centering
        \includegraphics[trim=155 0 0 0, clip,width=\textwidth, height=3cm]{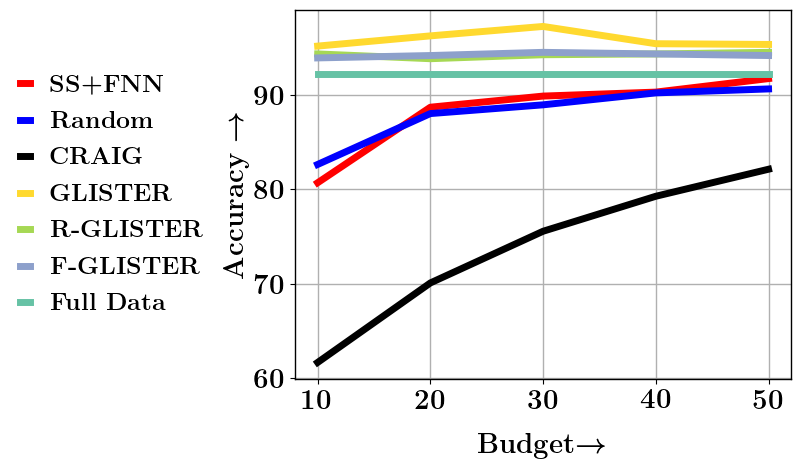}
        \subcaption{}
    \end{subfigure}
    \hfill
    \begin{subfigure}[b]{0.29\textwidth}
        \centering
        \includegraphics[trim=155 0 0 0, clip,width=\textwidth, height=3cm]{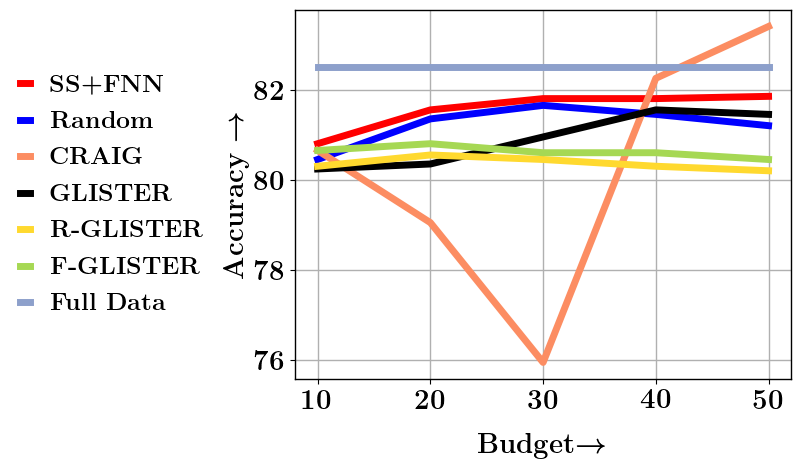}
        \subcaption{}
    \end{subfigure}
    \hfill
    \begin{subfigure}[b]{0.32\textwidth}
        \centering
        \includegraphics[trim=155 0 0 0, clip,width=\textwidth, height=3cm]{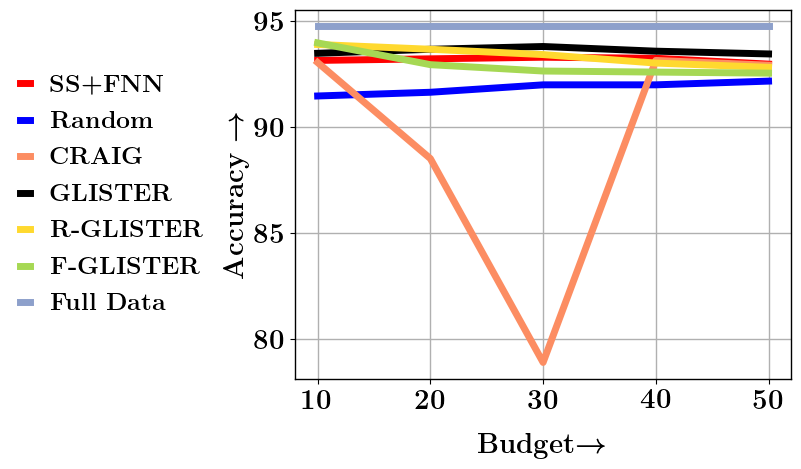}
        \subcaption{}
    \end{subfigure}
    \hfill
    \begin{subfigure}[b]{0.32\textwidth}
        \centering
        \includegraphics[trim=155 0 0 0, clip,width=\textwidth, height=3cm]{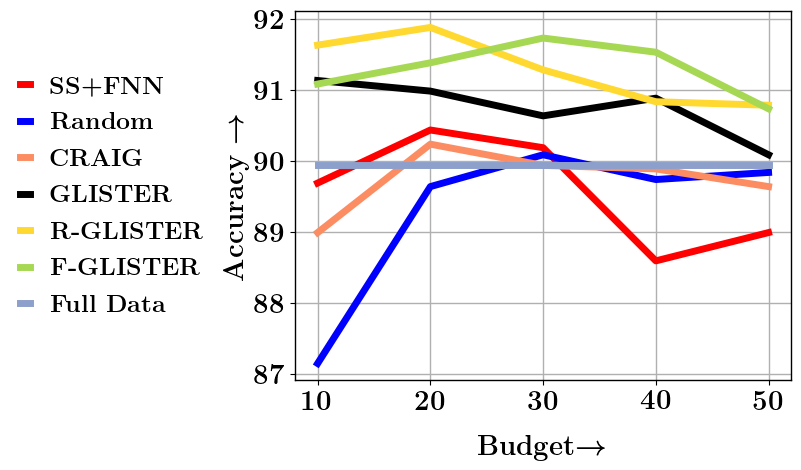}
        \subcaption{}
    \end{subfigure}
    \hfill
    \begin{subfigure}[b]{0.32\textwidth}
        \centering
        \includegraphics[trim=155 0 0 0, clip,width=\textwidth, height=3cm]{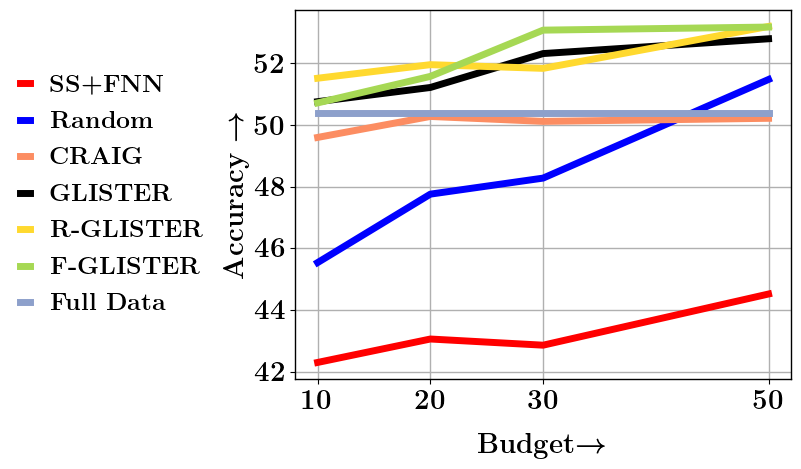}
        \subcaption{}
    \end{subfigure}
    \hfill
    \caption{\textbf{Top Row}: Data Selection for Efficient Learning. Datasets: 
    (a) Digits (b) DNA, (c) SatImage, \textbf{Second from Top Row}: (d) SVMGuide, (e)USPS, (f) Letter}
    \label{fig:dss_experiments}
\end{figure*}

We extend our discussion on effect of data selection for efficiency or faster training from the main paper with results on few more datasets as shown in figure \ref{fig:dss_experiments}. We compare subset sizes of 10\%, 30\%, 50\% in the shallow learning setting for the new datasets such as sklearn-digits, satimage, svmguide and letter. Here also we find that our method outperform other baselines significantly and are able to achieve comparable performance to full training for most of the datasets even when using smaller subset sizes. 

\subsection{Robust Learning}
\subsubsection{Class Imbalance:}
\begin{figure*}[!htbp]
    \centering
    \begin{subfigure}[b]{0.32\textwidth}
        \centering
        \includegraphics[width=\textwidth, height=3cm]{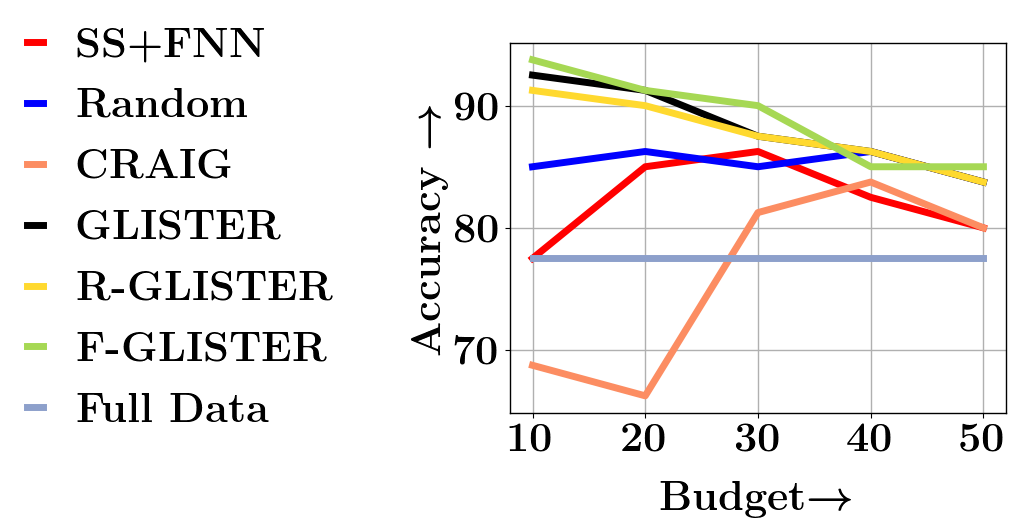}
        \subcaption{}
    \end{subfigure}
    \hfill
    \begin{subfigure}[b]{0.22\textwidth}
        \centering
        \includegraphics[trim=155 0 0 0, clip,width=\textwidth, height=3cm]{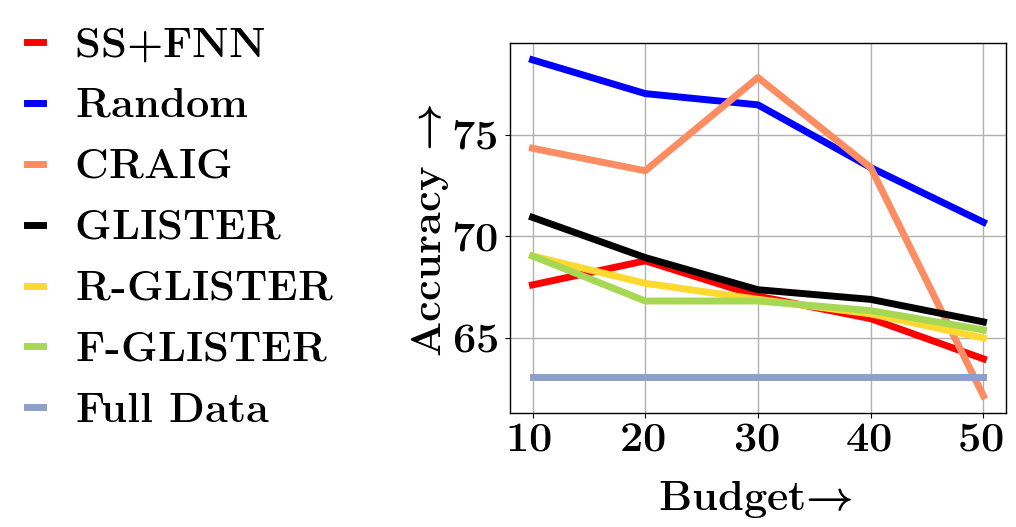}
        \subcaption{}
    \end{subfigure}
    \hfill
    \begin{subfigure}[b]{0.22\textwidth}
        \centering
        \includegraphics[trim=155 0 0 0, clip,width=\textwidth, height=3cm]{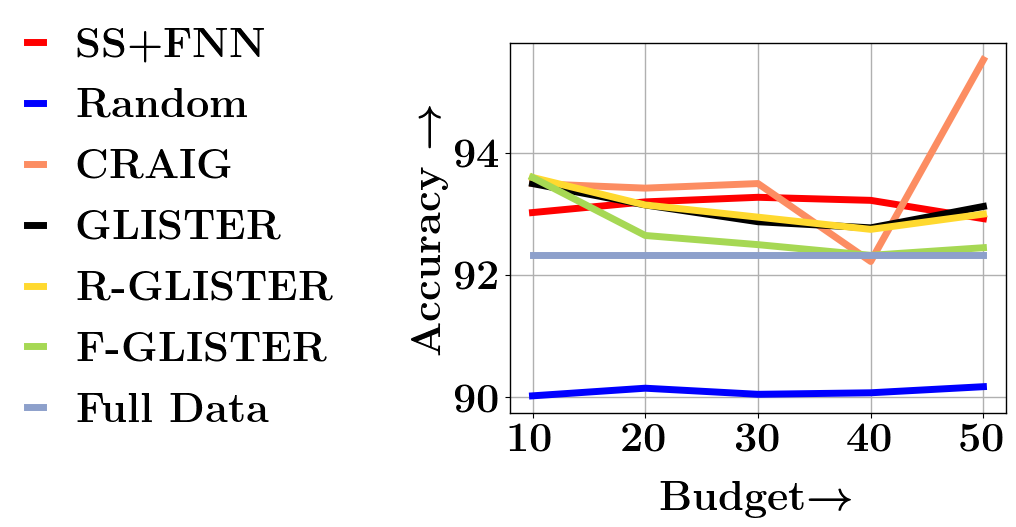}
        \subcaption{}
    \end{subfigure}
    \hfill
    \begin{subfigure}[b]{0.22\textwidth}
        \centering
        \includegraphics[trim=155 0 0 0, clip,width=\textwidth, height=3cm]{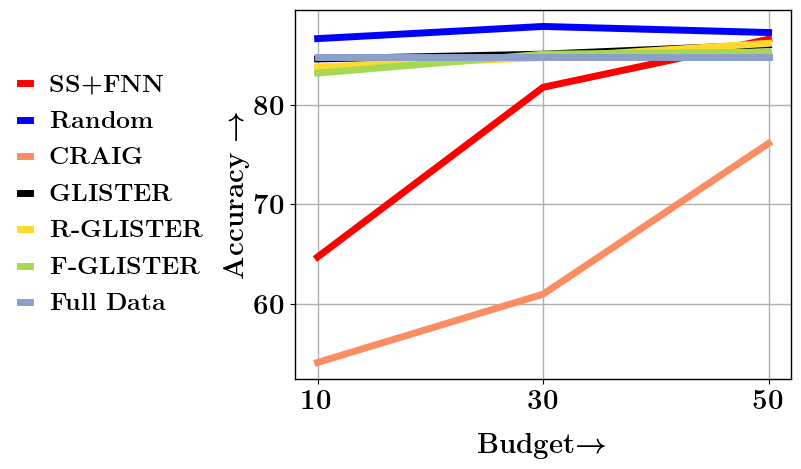}
        \subcaption{}
    \end{subfigure}
    \hfill
    \caption{Data Selection for Robust Learning in Class Imbalance Setting. Datasets: 
    (a) Digits ClassImb, (b) SatImage ClassImb, (c) SVMGuide ClassImb, (d)USPS ClassImb}
    \label{fig:dss_experiments_classImb}
\end{figure*}

We extend our discussion on effect of Robust data selection for better generalization from the main paper in class imbalance setting with results on few more datasets as shown in figure \ref{fig:dss_experiments_classImb}. We compare subset sizes of 10\%, 30\%, 50\% in the shallow learning setting for the new datasets such as sklearn-digits, satimage, svmguide. Here also we find that our method outperform other baselines significantly. We also see that our method often outperforms full training which essentially showcases the generalization capacity of our \modelonline\ Framework.

\subsubsection{Noisy Labels Setting}
\begin{figure*}[!htbp]
    \centering
    \begin{subfigure}[b]{0.40\textwidth}
        \centering
        \includegraphics[width=\textwidth, height=3cm]{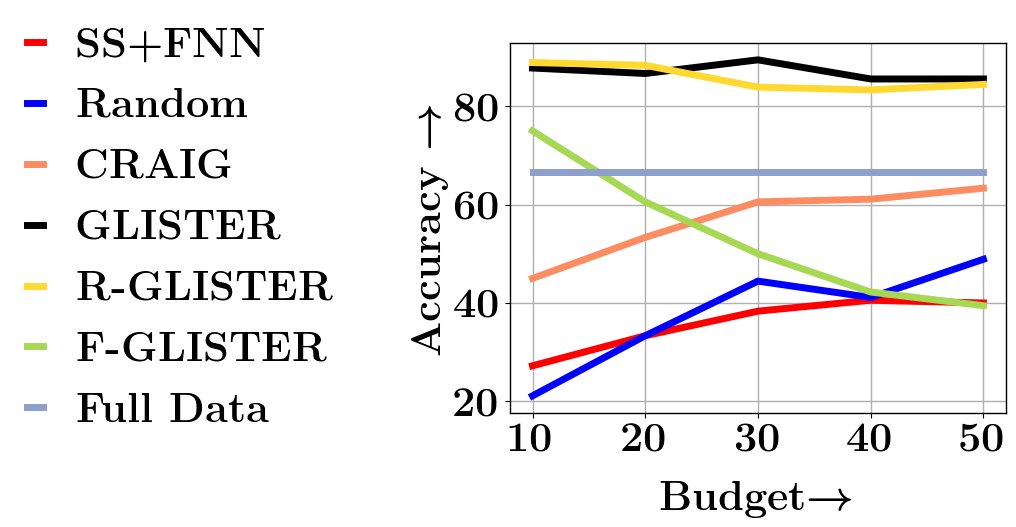}
        \subcaption{}
    \end{subfigure}
    \hfill
    \begin{subfigure}[b]{0.29\textwidth}
        \centering
        \includegraphics[trim=155 0 0 0, clip,width=\textwidth, height=3cm]{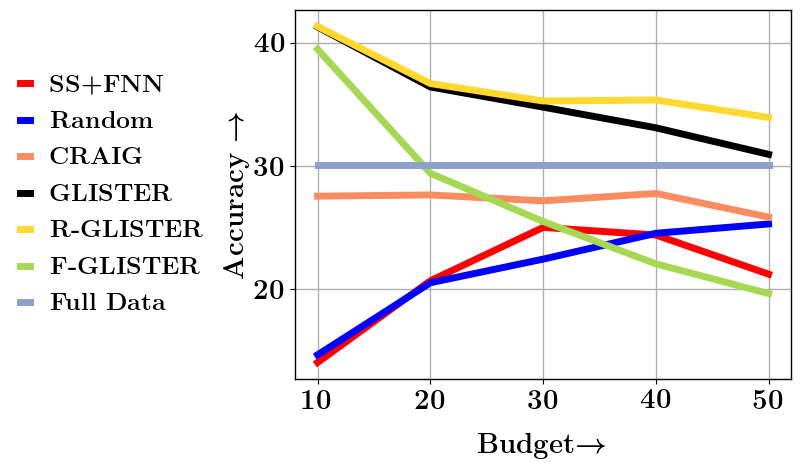}
        \subcaption{}
    \end{subfigure}
    \hfill
    \begin{subfigure}[b]{0.29\textwidth}
        \centering
        \includegraphics[trim=155 0 0 0, clip,width=\textwidth, height=3cm]{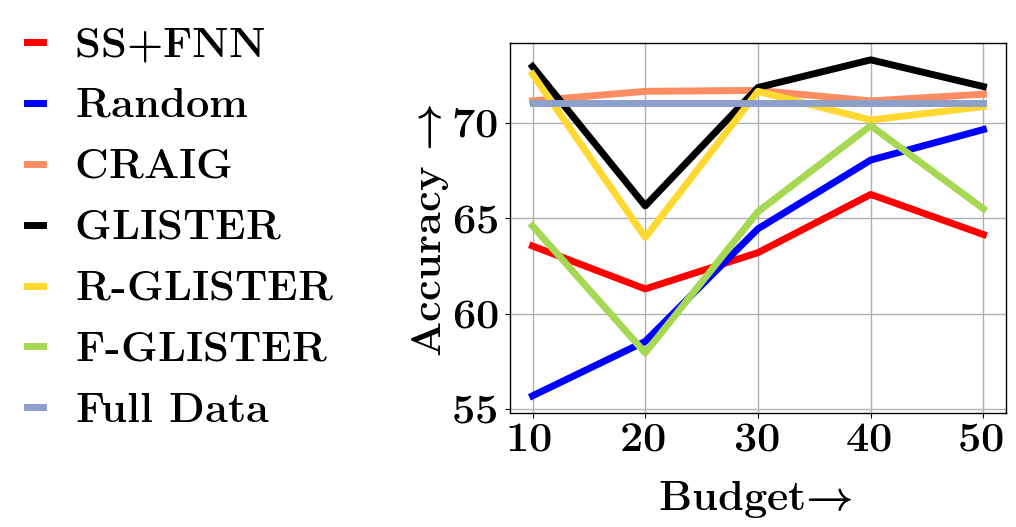}
        \subcaption{}
    \end{subfigure}
    \hfill
    \begin{subfigure}[b]{0.29\textwidth}
        \centering
        \includegraphics[trim=155 0 0 0, clip,width=\textwidth, height=3cm]{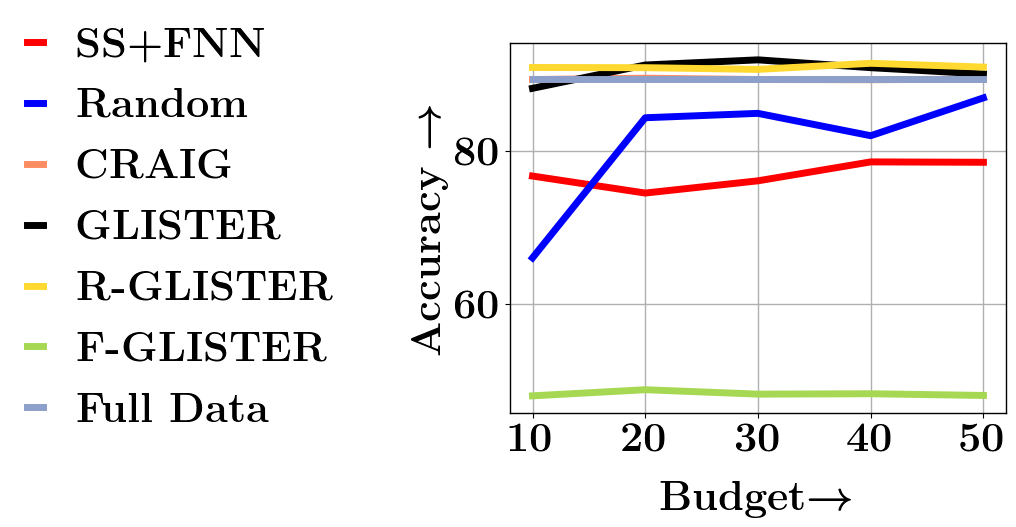}
        \subcaption{}
    \end{subfigure}
    \hfill
    \begin{subfigure}[b]{0.29\textwidth}
        \centering
        \includegraphics[trim=155 0 0 0, clip,width=\textwidth, height=3cm]{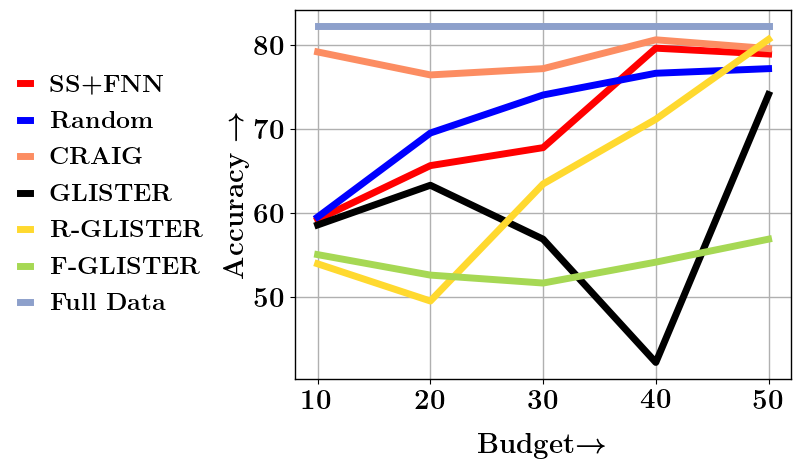}
        \subcaption{}
    \end{subfigure}
    \hfill
    \caption{\textbf{Top Row}: Data Selection for Robust Learning in Noisy Labels Setting. Datasets: 
    (a) Noisy Digits Dataset, (b) Noisy Letter dataset, (c) Noisy SatImage Dataset, \textbf{Second from Top Row}: (d) Noisy SVMGuide Dataset, (e)Noisy USPS Dataset}
    \label{fig:dss_experiments_noisy}
\end{figure*}

We extend our discussion on effect of Robust data selection for better generalization from the main paper in Noisy Labels Setting with results on few more datasets as shown in figure \ref{fig:dss_experiments_noisy}. We compare subset sizes of 10\%, 30\%, 50\% in the shallow learning setting for the new datasets such as sklearn-digits, satimage, svmguide. We used a Noise ratio of 80\% in our experiments. From the results shown in figure \ref{fig:dss_experiments_noisy}, it is evident that our method outperformed other baselines significantly and achieved an accuracy comparable to performance of the model trained on dataset with no noise  as shown in figure~\ref{fig:dss_experiments}. This highlights the generalization capacity of \modelonline\ even with much larger noise ratios.

\subsection{Active Learning Results}
\begin{figure*}[!htbp]
    \centering
    \begin{subfigure}[b]{0.32\textwidth}
        \centering
        \includegraphics[width=\textwidth, height=3cm]{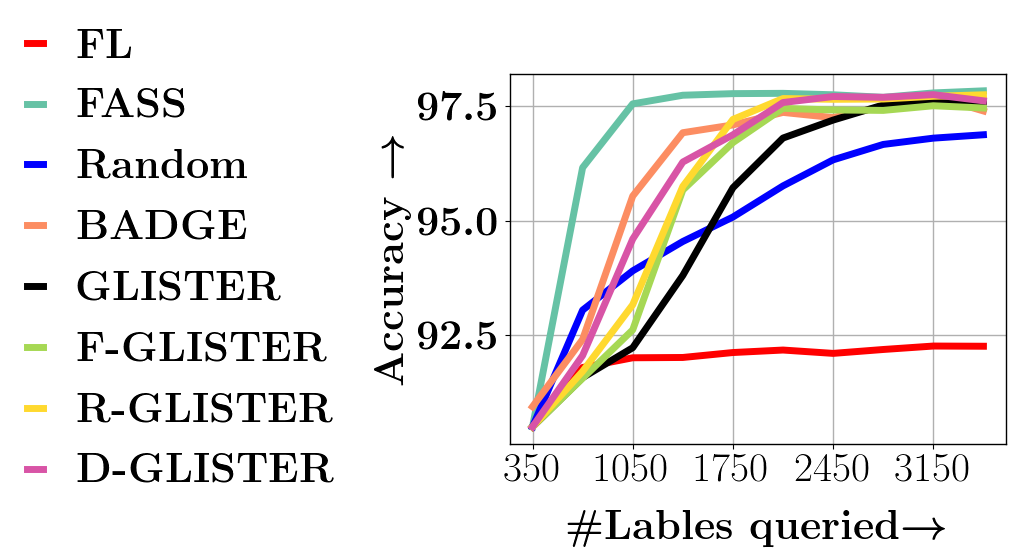}
        \subcaption{}
    \end{subfigure}
    \hfill
    \begin{subfigure}[b]{0.32\textwidth}
        \centering
        \includegraphics[trim=155 0 0 0, clip,width=\textwidth, height=3cm]{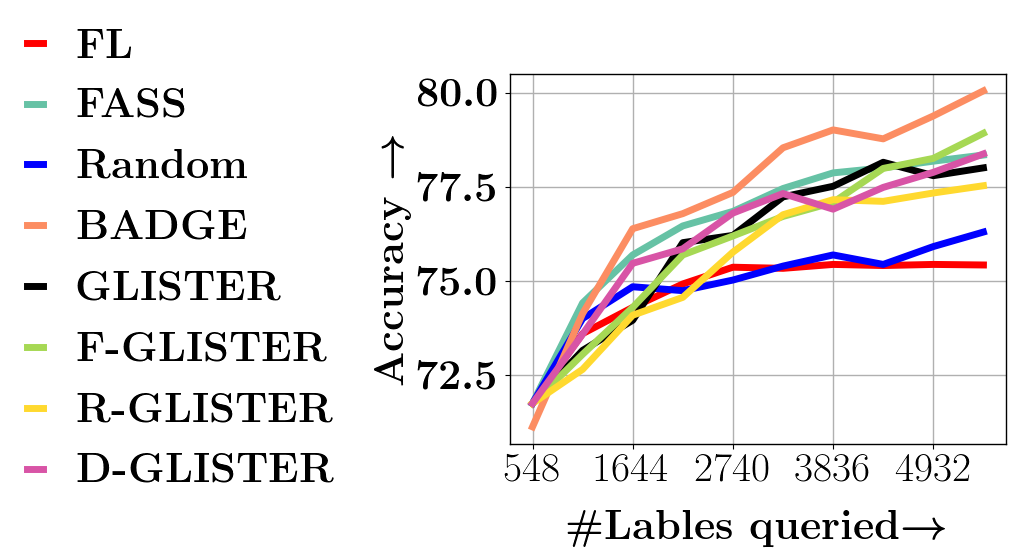}
        \subcaption{}
    \end{subfigure}
    \hfill
    \begin{subfigure}[b]{0.32\textwidth}
        \centering
        \includegraphics[trim=155 0 0 0, clip,width=\textwidth, height=3cm]{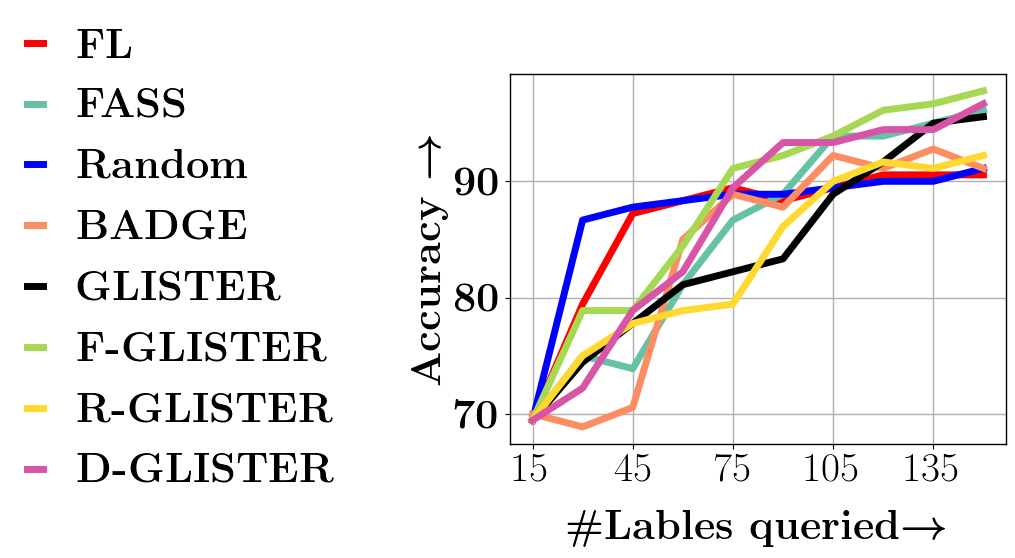}
        \subcaption{}
    \end{subfigure}
    \hfill
    \begin{subfigure}[b]{0.32\textwidth}
        \centering
        \includegraphics[trim=155 0 0 0, clip,width=\textwidth, height=3cm]{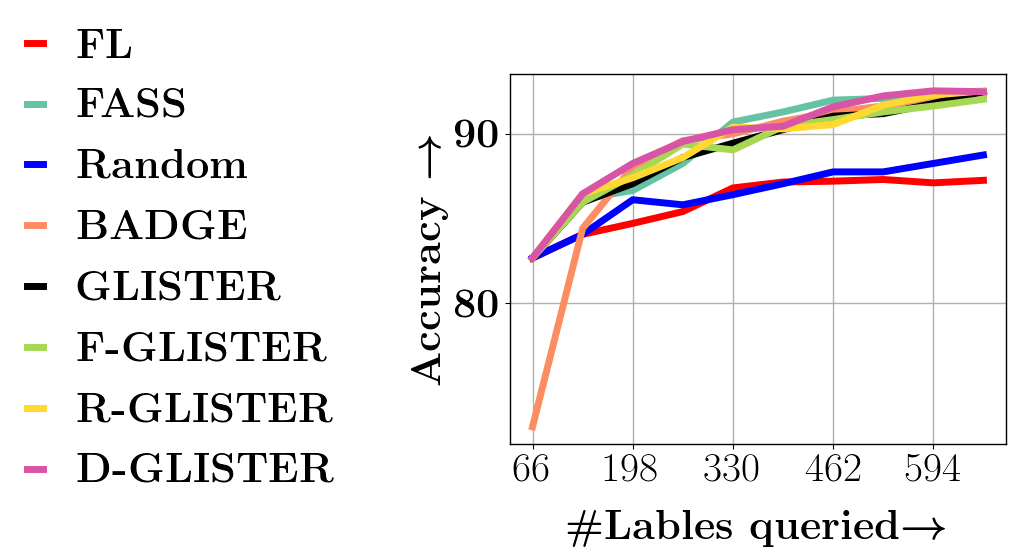}
        \subcaption{}
    \end{subfigure}
    \hfill
    \begin{subfigure}[b]{0.32\textwidth}
        \centering
        \includegraphics[trim=155 0 0 0, clip,width=\textwidth, height=3cm]{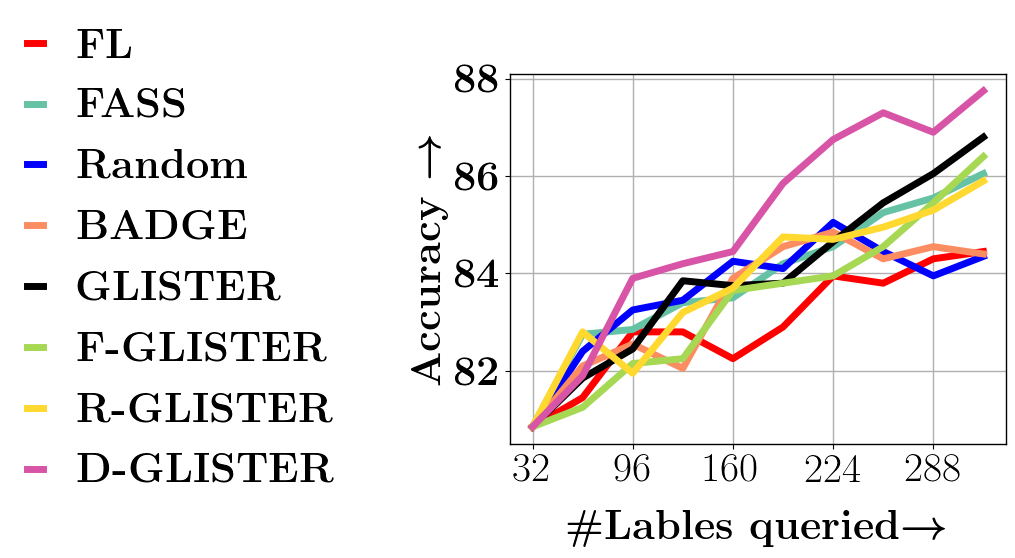}
        \subcaption{}
    \end{subfigure}
    \hfill
    \begin{subfigure}[b]{0.32\textwidth}
        \centering
        \includegraphics[trim=155 0 0 0, clip,width=\textwidth, height=3cm]{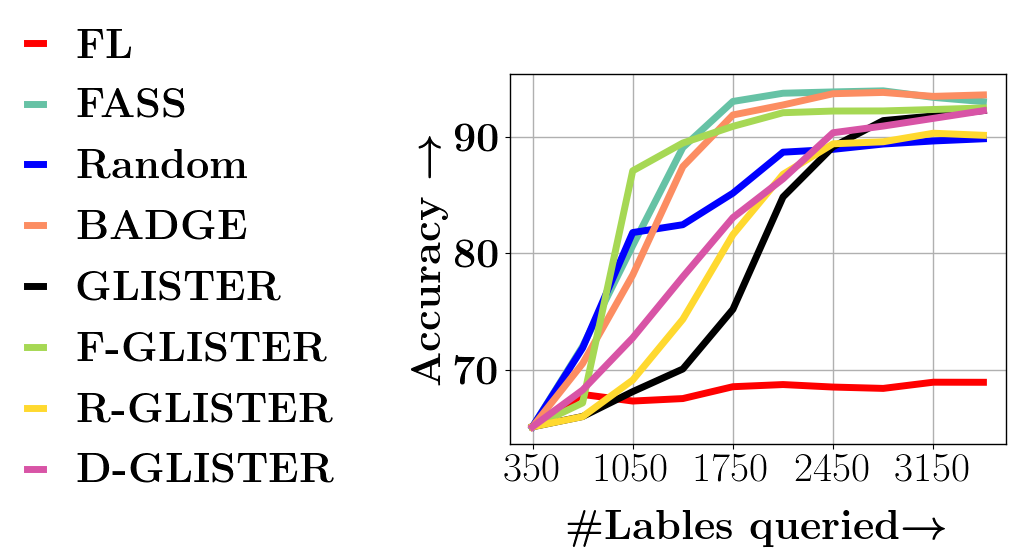}
        \subcaption{}
    \end{subfigure}
    \hfill
    \begin{subfigure}[b]{0.32\textwidth}
        \centering
        \includegraphics[trim=155 0 0 0, clip,width=\textwidth, height=3cm]{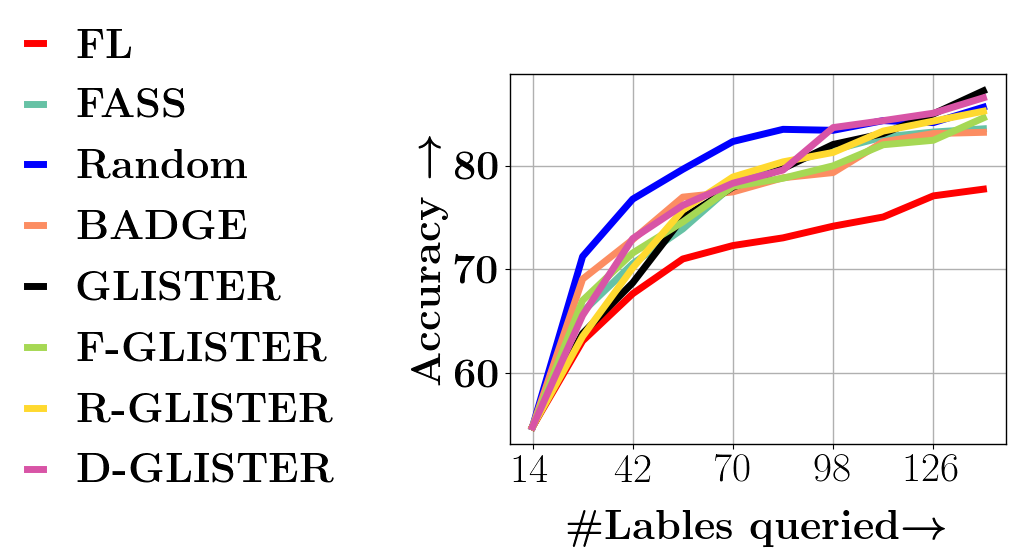}
        \subcaption{}
    \end{subfigure}
    \hfill
    \begin{subfigure}[b]{0.32\textwidth}
        \centering
        \includegraphics[trim=155 0 0 0, clip,width=\textwidth, height=3cm]{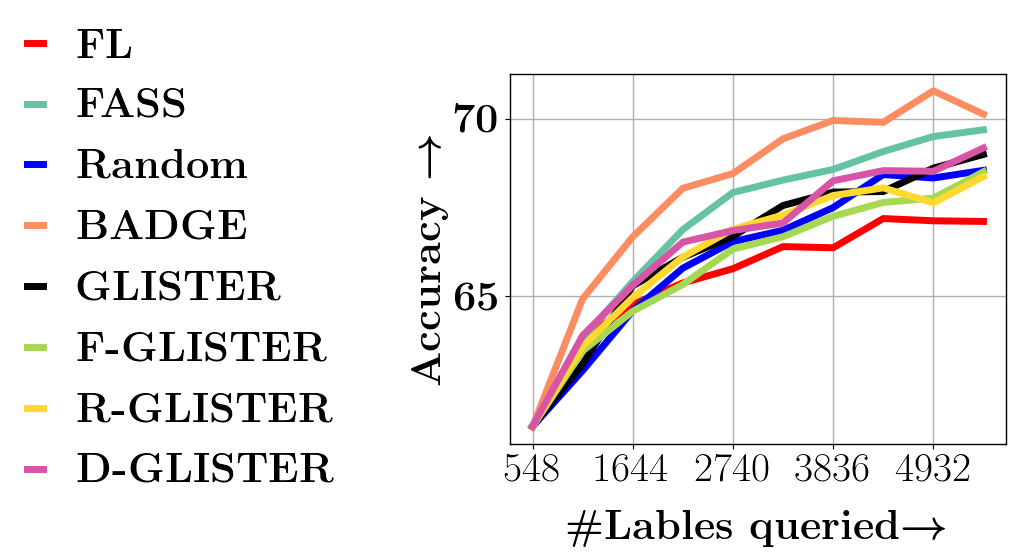}
        \subcaption{}
    \end{subfigure}
    \hfill
    \begin{subfigure}[b]{0.32\textwidth}
        \centering
        \includegraphics[trim=155 0 0 0, clip,width=\textwidth, height=3cm]{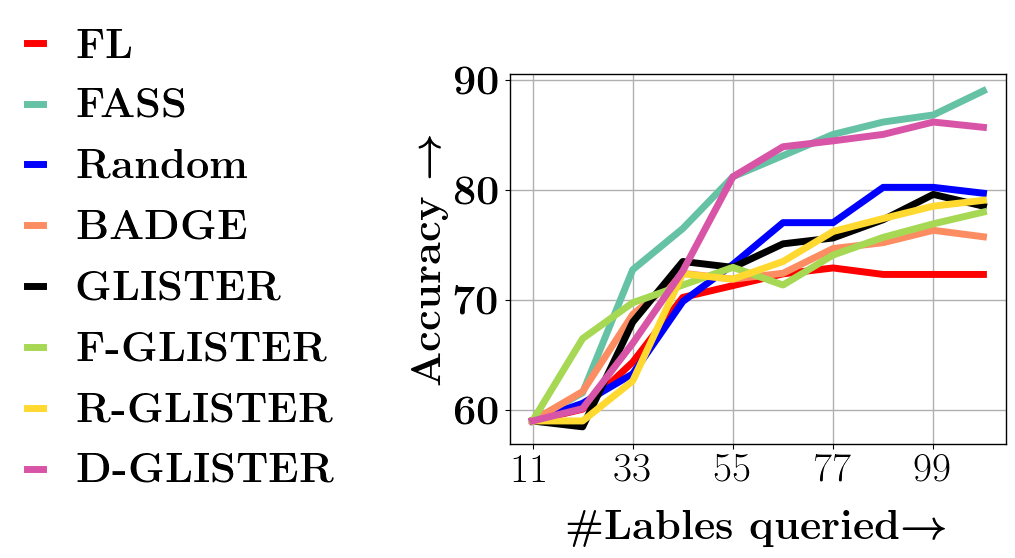}
        \subcaption{}
    \end{subfigure}
    \hfill
    \begin{subfigure}[b]{0.32\textwidth}
        \centering
        \includegraphics[trim=155 0 0 0, clip,width=\textwidth, height=3cm]{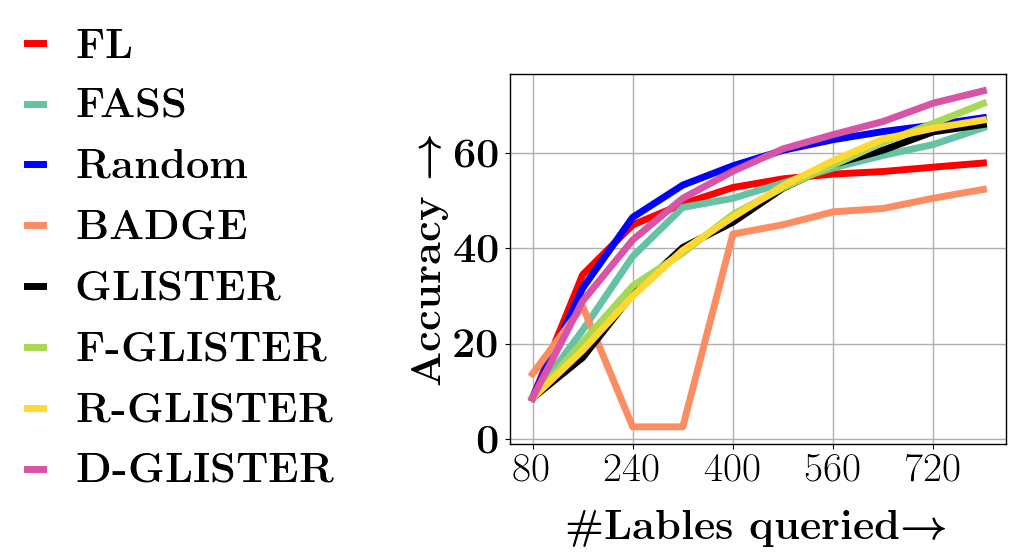}
        \subcaption{}
    \end{subfigure}
    \hfill
    \begin{subfigure}[b]{0.32\textwidth}
        \centering
        \includegraphics[trim=155 0 0 0, clip,width=\textwidth, height=3cm]{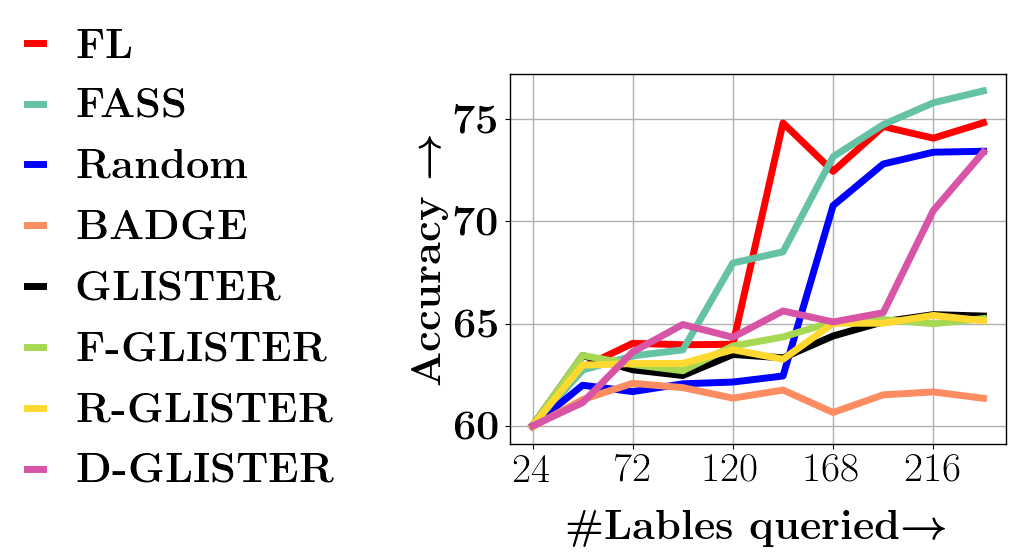}
        \subcaption{}
    \end{subfigure}
    \hfill
    \begin{subfigure}[b]{0.32\textwidth}
        \centering
        \includegraphics[trim=155 0 0 0, clip,width=\textwidth, height=3cm]{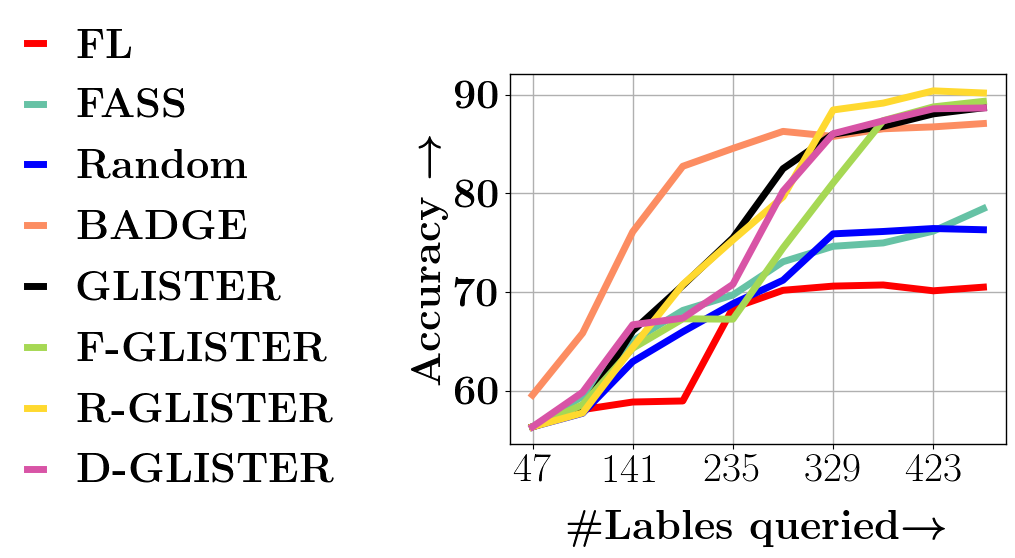}
        \subcaption{}
    \end{subfigure}
    \caption{\textbf{Top Row}: Data Selection for ACTIVE Learning. Datasets: (a) IJCNN1 (b) Connect-4, (c) Digits, \textbf{Second from Top Row}:(d) USPS, (e) SatImage,\\Data Selection for ACTIVE Learning in CLass Imbalance Setting. Datasets: (f) IJCNN1 ClassImb,\textbf{Third from Top Row}:  (g) DNA ClassImb, (h) Connect-4 ClassImb, (i) Digits ClassImb, \textbf{Fourth Row}:(j) Letter ClassImb, (k) SatImage ClassImb, (l) USPS ClassImb} \label{fig:active_experiments}
\end{figure*}

We extend our discussion of ACTIVE learning setting discussed in main paper by showing results on more datasets as shown in figure~\ref{fig:active_experiments}. We compare the performance of \modelactive\ to other state of the art baselines on both normal datasets and datasets with class imbalance.
From the results shown in figure~\ref{fig:active_experiments} on normal datasets, we can say that our \modelactive\ framework performed comparable to the state of the art baselines BADGE \cite{ash2020deep} and FASS \cite{wei2015submodularity} that are explicitly designed for Active Learning. We also observed that our \modelactive\ often outperformed the baselines when our dataset has class imbalance in it. This highlights the usefulness of our \modelactive\ framework for Robust Active Learning when our datasets are prone to adversaries.

\subsection{Ablation Studies}
\subsubsection{Convergence with varying L values}

\begin{figure*}[!htbp]
    \centering
    \begin{subfigure}[b]{0.32\textwidth}
        \centering
        \includegraphics[width=\textwidth, height=3cm]{new_fig/L_analysis/dna_L.pdf}
        \subcaption{}
    \end{subfigure}
    \hfill
    \begin{subfigure}[b]{0.32\textwidth}
        \centering
        \includegraphics[width=\textwidth, height=3cm]{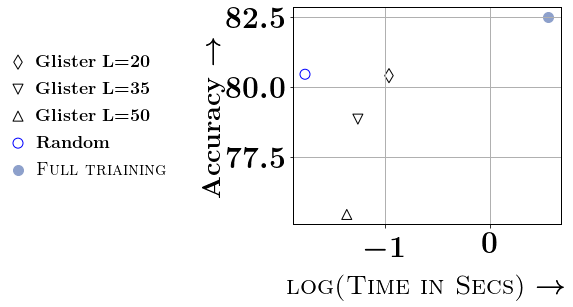}
        \subcaption{}
    \end{subfigure}
    \hfill
    \begin{subfigure}[b]{0.32\textwidth}
        \centering
        \includegraphics[width=\textwidth, height=3cm]{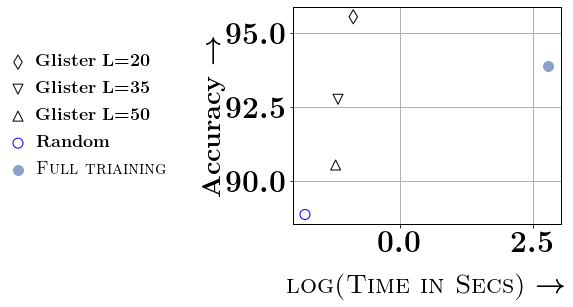}
        \subcaption{}
    \end{subfigure}
    \hfill
    \begin{subfigure}[b]{0.32\textwidth}
        \centering
        \includegraphics[width=\textwidth, height=3cm]{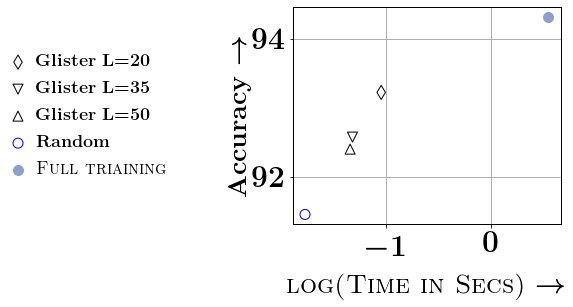}
        \subcaption{}
    \end{subfigure}
    \hfill
    \begin{subfigure}[b]{0.32\textwidth}
        \centering
        \includegraphics[width=\textwidth, height=3cm]{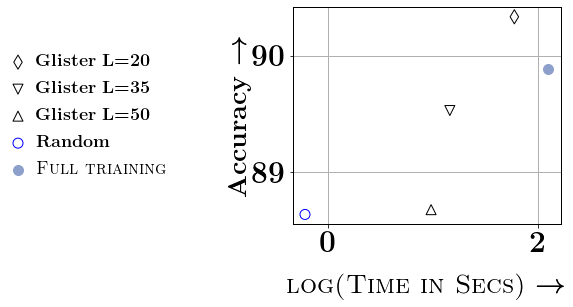}
        \subcaption{}
    \end{subfigure}
    \caption{\textbf{Top Row}: Convergence time for Data Selection for efficient Learning with varying L values. Datasets: 
    (a) DNA (b) Satimage, (c) Digits \textbf{Second from Top Row}:(d) Svmguide1,(e) USPS.} \label{fig:L_convergence_experiments}
\end{figure*}

In this section, we analyzed the effect of varying $L$ values on our model convergence rate on various datasets as shown in the figure~\ref{fig:L_convergence_experiments}. From the results, it is evident that our \modelonline\ framework has faster convergence for lower $L$ values. But using lower $L$ values leads to high computational time. This is evident from the results on some datsets where $L=50$ has faster convergence than $L=20$. Hence, we should be careful about the above trade-off when deciding the $L$ value. After extensive analysis, we used $L=20$ in our experiments.

\subsubsection{Convergence with varying r values}
\begin{figure*}[!htbp]
    \centering
    \begin{subfigure}[b]{0.32\textwidth}
        \centering
        \includegraphics[width=\textwidth, height=3cm]{new_fig/r_analysis/dna_r.pdf}
        \subcaption{}
    \end{subfigure}
    \hfill
    \begin{subfigure}[b]{0.32\textwidth}
        \centering
        \includegraphics[width=\textwidth, height=3cm]{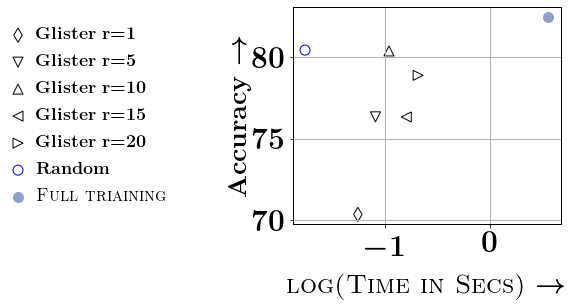}
        \subcaption{}
    \end{subfigure}
    \centering
    \begin{subfigure}[b]{0.32\textwidth}
        \centering
        \includegraphics[width=\textwidth, height=3cm]{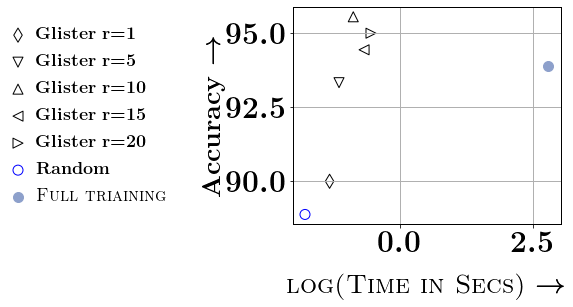}
        \subcaption{}
    \end{subfigure}
    \hfill
    \begin{subfigure}[b]{0.32\textwidth}
        \centering
        \includegraphics[width=\textwidth, height=3cm]{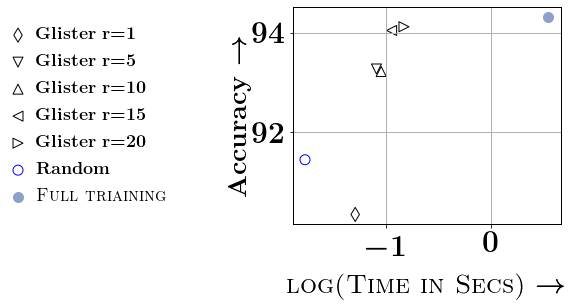}
        \subcaption{}
    \end{subfigure}
    \hfill
    \begin{subfigure}[b]{0.32\textwidth}
        \centering
        \includegraphics[width=\textwidth, height=3cm]{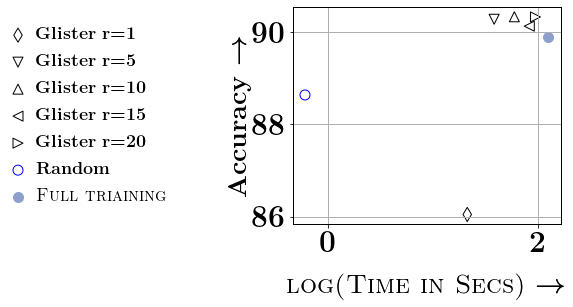}
        \subcaption{}
    \end{subfigure}
    \caption{\textbf{Top Row}: Convergence time for Data Selection for efficient Learning with varying r values. Datasets: 
    (a) DNA,  (b) Satimage, (c) Digits, \textbf{Second from Top Row}:(d) Svmguide1, (e) USPS.} \label{fig:r_convergence_experiments}
\end{figure*}

In this section, we analyzed the effect of varying $r$ values on our model convergence rate on various datasets as shown in the figure~\ref{fig:r_convergence_experiments}. From the results, it is evident that our \modelonline\ framework is highly unstable for low values of $r$ and that the stability of our model increases with $r$. But using large $r$ values leads to high computational time. Hence, it is not always advisable to use large $r$ values. It is also evident from our results shown in figure~\ref{fig:r_convergence_experiments}, since our \modelonline\ framework has faster convergence when $r=10$ instead of $r=20$ in the majority of the datasets. Hence, after extensive analysis we used $r=0.03K$ in our experiments.

\subsection{Synthetic Experiments}

\begin{figure*}[!htpb]
    \centering
    \begin{subfigure}[b]{0.19\textwidth}
        \centering
        \includegraphics[width=\textwidth, height=3cm]{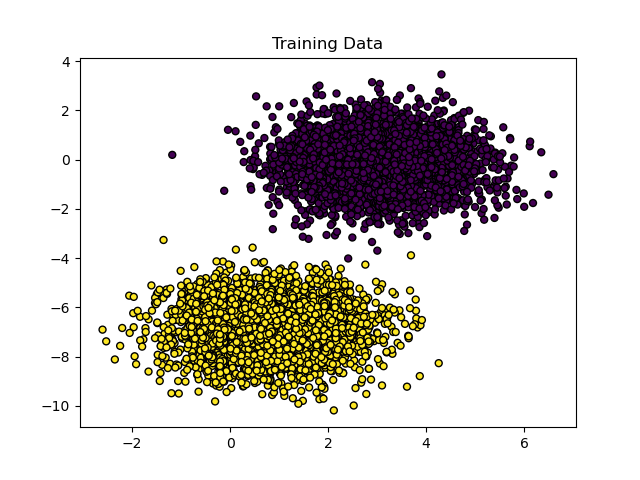}
        \caption{\label{synthetic_subset}}
    \end{subfigure}
    \hfill
    \begin{subfigure}[b]{0.19\textwidth}
        \centering
        \includegraphics[width=\textwidth, height=3cm]{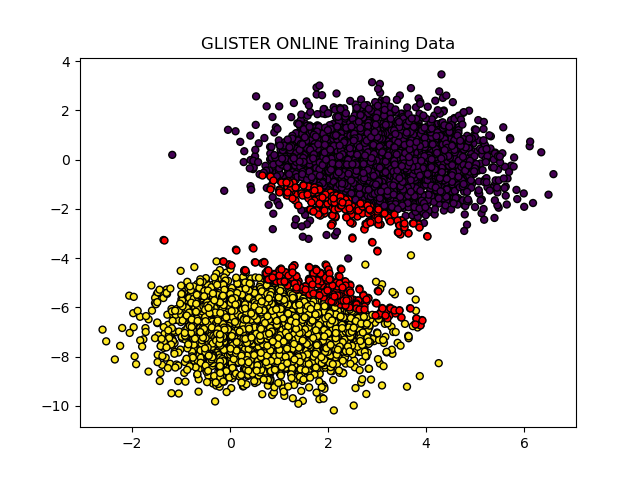}
        \caption{\label{glister-subset}}
    \end{subfigure}
    \hfill
    \begin{subfigure}[b]{0.19\textwidth}
        \centering
        \includegraphics[width=\textwidth, height=3cm]{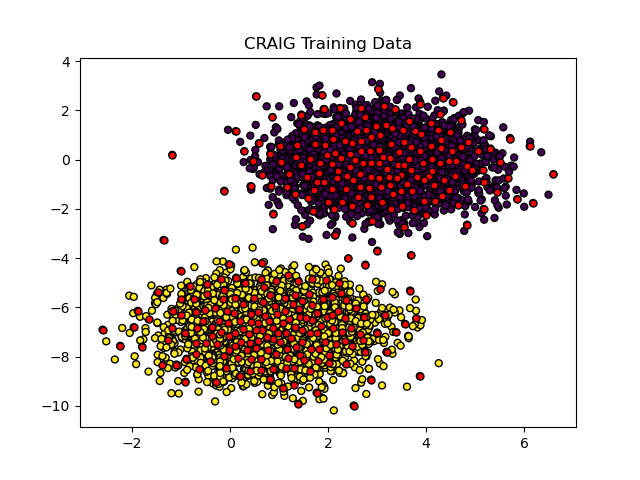}
        \caption{\label{craig-subset}}
    \end{subfigure}
    \hfill
    \begin{subfigure}[b]{0.19\textwidth}
        \centering
        \includegraphics[width=\textwidth, height=3cm]{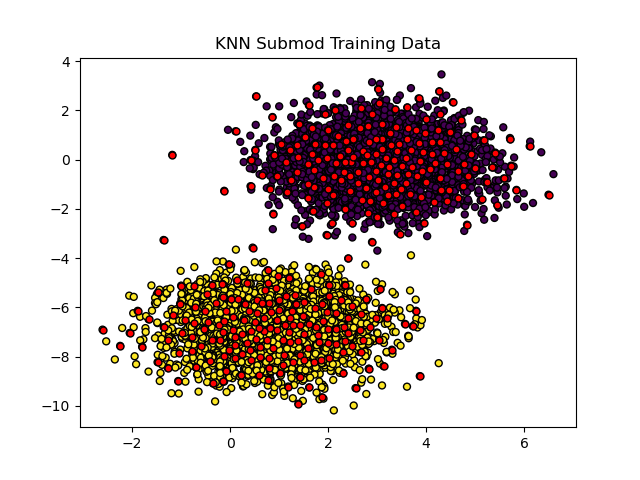}
        \caption{\label{knnsubmod-subset}}
    \end{subfigure}
    \hfill
    \begin{subfigure}[b]{0.19\textwidth}
        \centering
        \includegraphics[width=\textwidth, height=3cm]{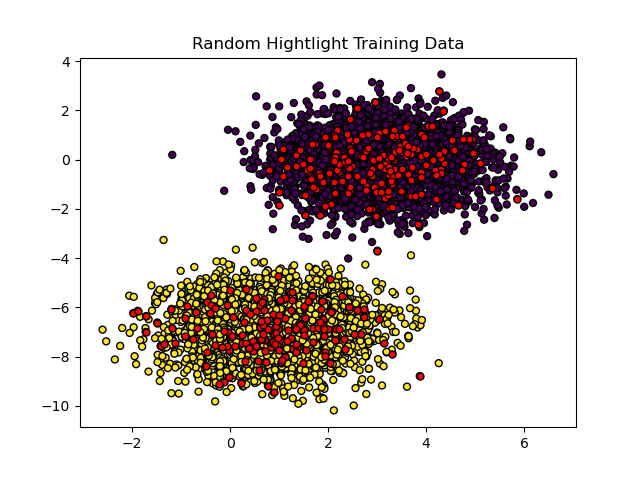}
        \caption{\label{random-subset}}
    \end{subfigure}
    \small{
    \caption{
    (a) Training Data (b) "\modelonline" Subset, (c) CRAIG Subset, (d) KNN Submodular Subset and (e) Random Subset}
    \label{fig:synthetic_data_dss_linsep_2}
    }
\end{figure*}

\begin{figure*}[!htpb]
    \centering
    \begin{subfigure}[b]{0.19\textwidth}
        \centering
        \includegraphics[width=\textwidth, height=3cm]{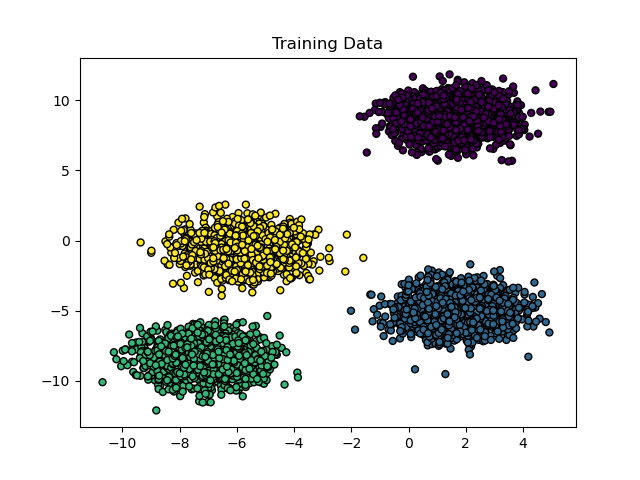}
        \caption{\label{synthetic_subset_4}}
    \end{subfigure}
    \hfill
    \begin{subfigure}[b]{0.19\textwidth}
        \centering
        \includegraphics[width=\textwidth, height=3cm]{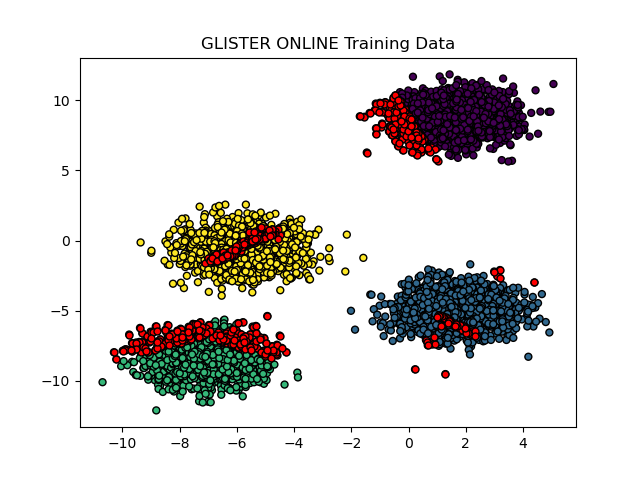}
        \caption{\label{glister-subset_4}}
    \end{subfigure}
    \hfill
    \begin{subfigure}[b]{0.19\textwidth}
        \centering
        \includegraphics[width=\textwidth, height=3cm]{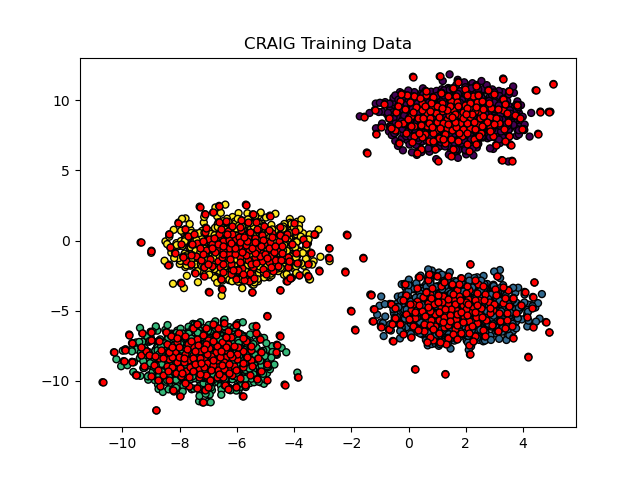}
        \caption{\label{craig-subset_4}}
    \end{subfigure}
    \hfill
    \begin{subfigure}[b]{0.19\textwidth}
        \centering
        \includegraphics[width=\textwidth, height=3cm]{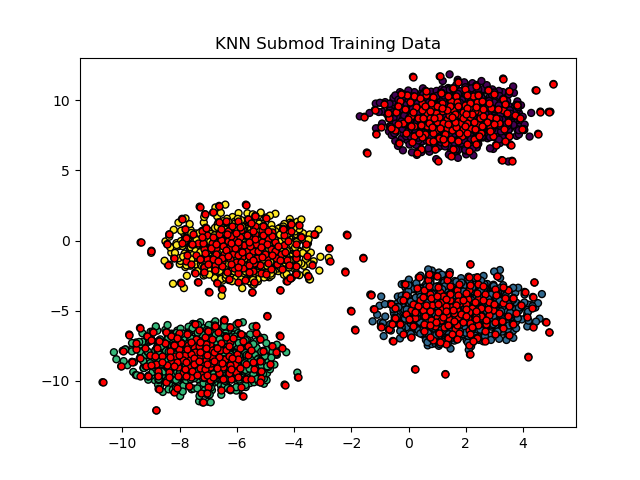}
        \caption{\label{knnsubmod-subset_4}}
    \end{subfigure}
    \hfill
    \begin{subfigure}[b]{0.19\textwidth}
        \centering
        \includegraphics[width=\textwidth, height=3cm]{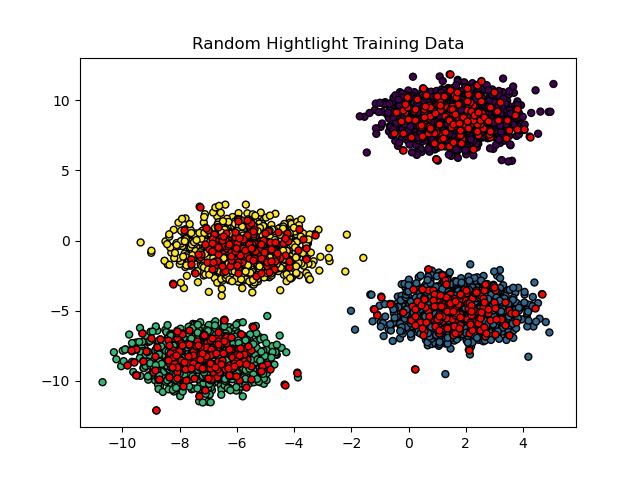}
        \caption{\label{random-subset_4}}
    \end{subfigure}
    \small{
    \caption{
    (a) Training Data (b) "\modelonline" Subset, (c) CRAIG Subset, (d) KNN Submodular Subset and (e) Random Subset}
    \label{fig:synthetic_data_dss_linsep_4}
    }
\end{figure*}

\begin{figure*}[!htpb]
    \centering
    \begin{subfigure}[b]{0.19\textwidth}
        \centering
        \includegraphics[width=\textwidth, height=3cm]{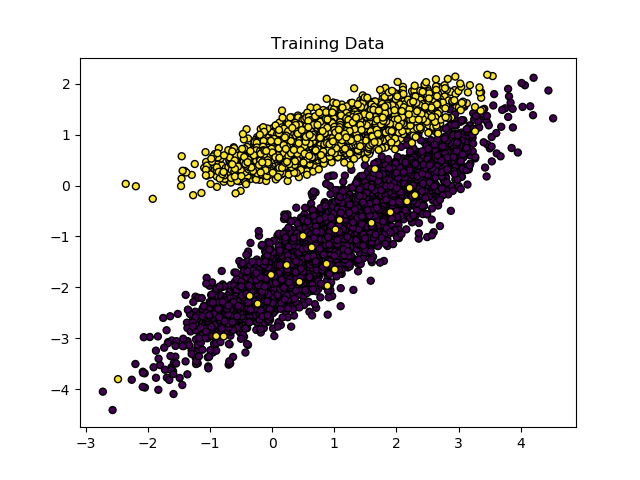}
        \caption{\label{synthetic_subset_bin_out}}
    \end{subfigure}
    \hfill
    \begin{subfigure}[b]{0.19\textwidth}
        \centering
        \includegraphics[width=\textwidth, height=3cm]{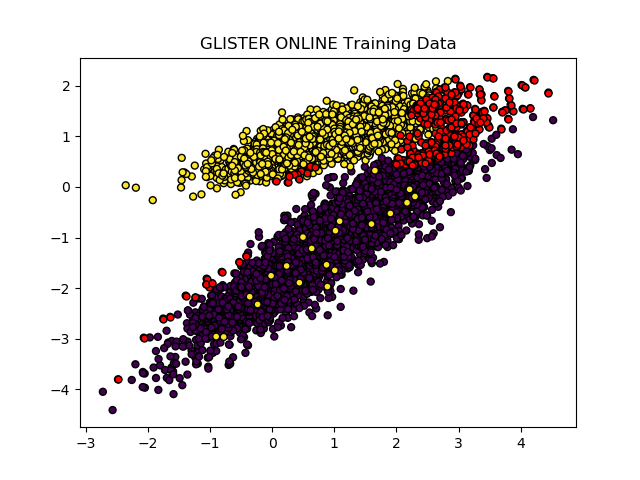}
        \caption{\label{glister-subset_bin_out}}
    \end{subfigure}
    \hfill
    \begin{subfigure}[b]{0.19\textwidth}
        \centering
        \includegraphics[width=\textwidth, height=3cm]{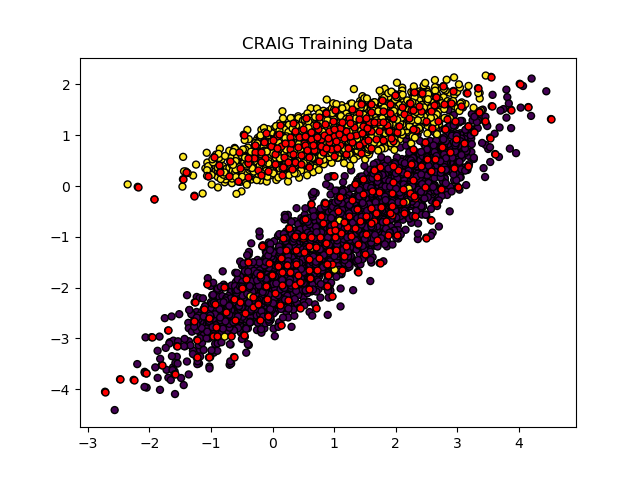}
        \caption{\label{craig-subset_bin_out}}
    \end{subfigure}
    \hfill
    \begin{subfigure}[b]{0.19\textwidth}
        \centering
        \includegraphics[width=\textwidth, height=3cm]{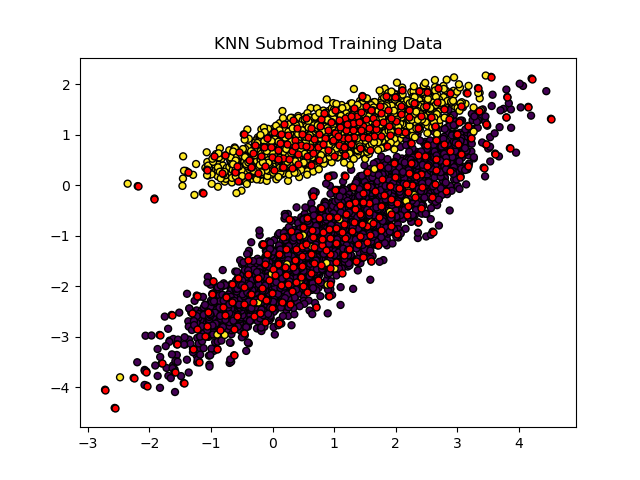}
        \caption{\label{knnsubmod-subset_bin_out}}
    \end{subfigure}
    \hfill
    \begin{subfigure}[b]{0.19\textwidth}
        \centering
        \includegraphics[width=\textwidth, height=3cm]{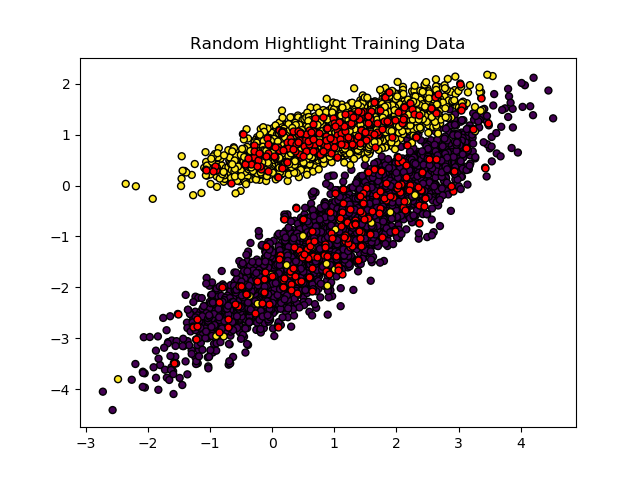}
        \caption{\label{random-binary-out}}
    \end{subfigure}
    \small{
    \caption{
    (a) Training Data (b) "\modelonline" Subset, (c) CRAIG Subset, (d) KNN Submodular Subset and (e) Random Subset}
    \label{fig:synthetic_data_dss_bin_out}
    }
\end{figure*}

\begin{figure*}[!htpb]
    \centering
    \begin{subfigure}[b]{0.19\textwidth}
        \centering
        \includegraphics[width=\textwidth, height=3cm]{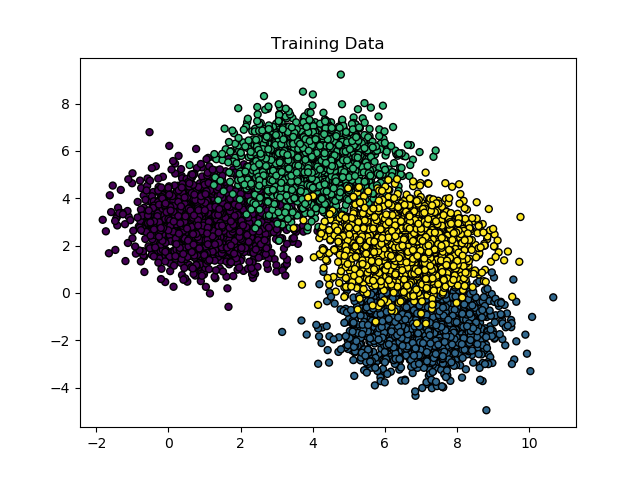}
        \caption{\label{synthetic_subset_four_close}}
    \end{subfigure}
    \hfill
    \begin{subfigure}[b]{0.19\textwidth}
        \centering
        \includegraphics[width=\textwidth, height=3cm]{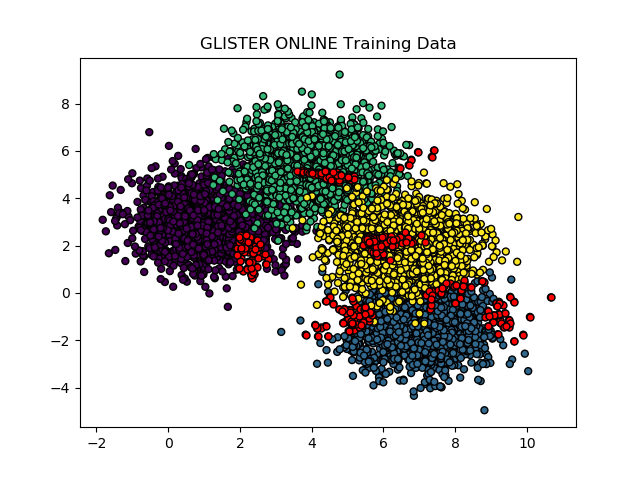}
        \caption{\label{glister-subset_four_close}}
    \end{subfigure}
    \hfill
    \begin{subfigure}[b]{0.19\textwidth}
        \centering
        \includegraphics[width=\textwidth, height=3cm]{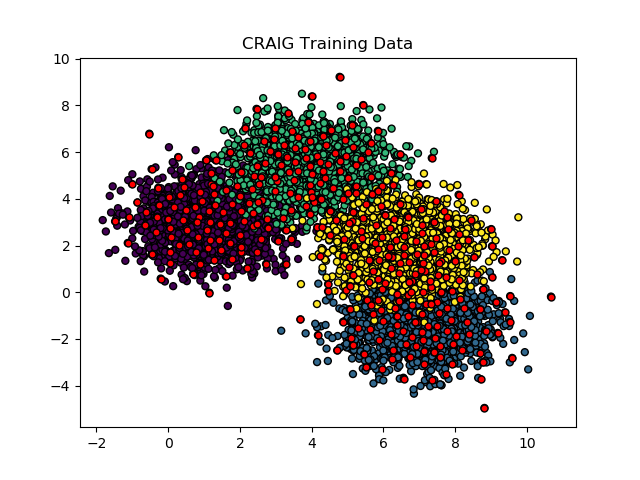}
        \caption{\label{craig-subset_four_close}}
    \end{subfigure}
    \hfill
    \begin{subfigure}[b]{0.19\textwidth}
        \centering
        \includegraphics[width=\textwidth, height=3cm]{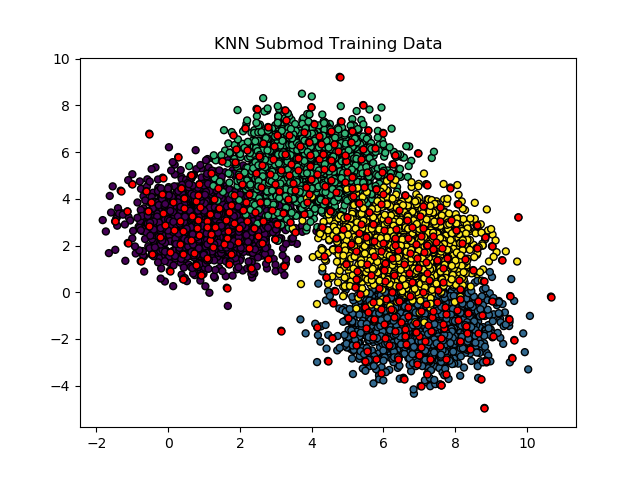}
        \caption{\label{knnsubmod-subset_four_close}}
    \end{subfigure}
    \hfill
    \begin{subfigure}[b]{0.19\textwidth}
        \centering
        \includegraphics[width=\textwidth, height=3cm]{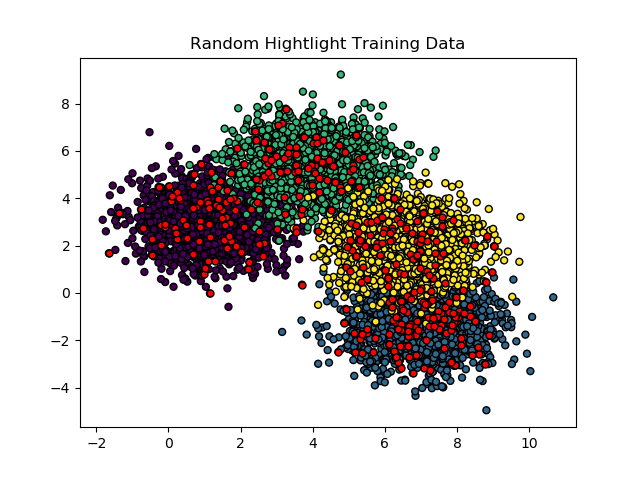}
        \caption{\label{random-subset_four_close}}
    \end{subfigure}
    \small{
    \caption{
    (a) Training Data (b) "\modelonline" Subset, (c) CRAIG Subset, (d) KNN Submodular Subset and (e) Random Subset}
    \label{fig:synthetic_data_dss_four_close}
    }
\end{figure*}

We specifically designed a set of synthetic data experiments to visualize how the subset selection process of our "\modelonline" framework works. Our synthetic datasets include linearly separable binary data, multi-class separable data, and linearly separable binary data with slack. The linearly separable binary dataset, as shown in fig.\ref{synthetic_subset}, comprises two-dimensional feature data points from two non-overlapping data points clusters of class 0 and class 1. The multi-class separable data comprises two-dimensional feature data points from four non-overlapping data points clusters of classes 0,1,2,3 which is shown in fig.\ref{synthetic_subset_4}. A overlapping version of the same is shown in fig.\ref{synthetic_subset_four_close}. Whereas the linearly separable dataset with slack comprises two-dimensional feature data points from two overlapping data points clusters of class 0 and class 1 as shown in fig.\ref{synthetic_subset_bin_out}.
 
 The result in \ref{fig:synthetic_data_dss_linsep_2} shows the subset selected by various methods for a linearly binary separable dataset. From fig. \ref{glister-subset}, we can see that our \modelonline\ framework selects a subset that is close to the boundary, whereas other methods like CRAIG\cite{mirzasoleiman2019coresets}, Random, KNNSubmod \cite{wei2015submodularity} selects subsets that are representative of the training dataset as shown in figures \ref{craig-subset}, \ref{random-subset}, \ref{knnsubmod-subset}  respectively. Similarly, in fig. \ref{fig:synthetic_data_dss_linsep_4} points selected by the methods are highlighted for linearly separable dataset with four classes, in figure~\ref{fig:synthetic_data_dss_bin_out} points selected by the methods are highlighted for binary dataset with outliers and in figure~\ref{fig:synthetic_data_dss_four_close} points selected by the methods are highlighted for overlapping dataset with four classes.
 
 From the results shown, it is clearly evident that our \modelonline\ Framework tries to select datapoints that are close to the  decision boundary where as other methods like CRAIG \cite{mirzasoleiman2019coresets}, KNNSubmod \cite{wei2015submodularity} selects representative points of the whole training set.
 
\begin{figure*}[!htpb]
    \centering
    \begin{subfigure}[b]{0.24\textwidth}
        \centering
        \includegraphics[width=\textwidth, height=3cm]{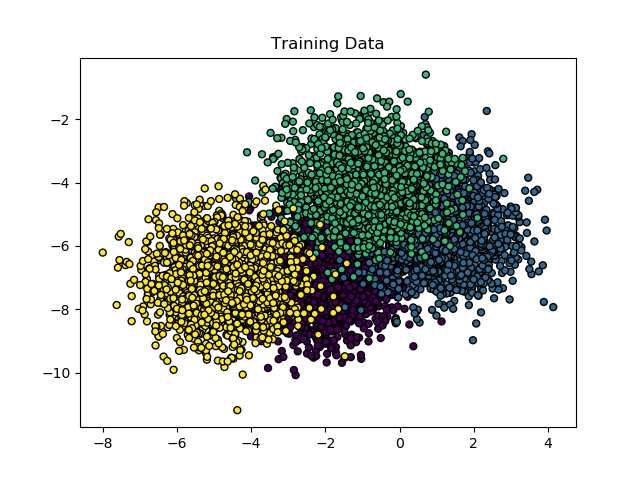}
        \caption{\label{synthetic_subset_four_close_shift}}
    \end{subfigure}
    \hfill
    \begin{subfigure}[b]{0.24\textwidth}
        \centering
        \includegraphics[width=\textwidth, height=3cm]{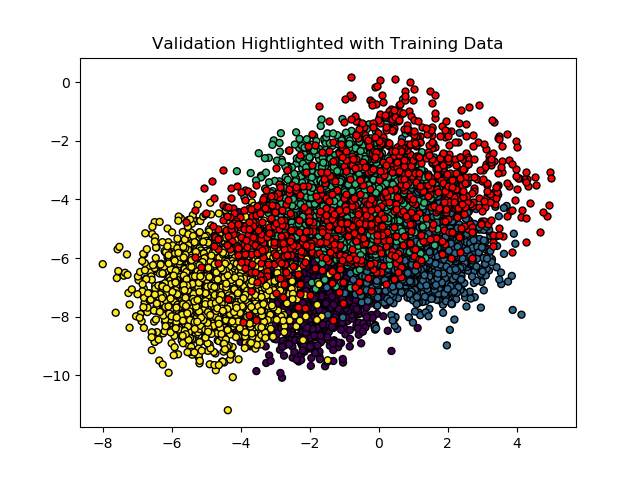}
        \caption{\label{synthetic_subset_four_close_shift_val}}
    \end{subfigure}
    \hfill
    \begin{subfigure}[b]{0.24\textwidth}
        \centering
        \includegraphics[width=\textwidth, height=3cm]{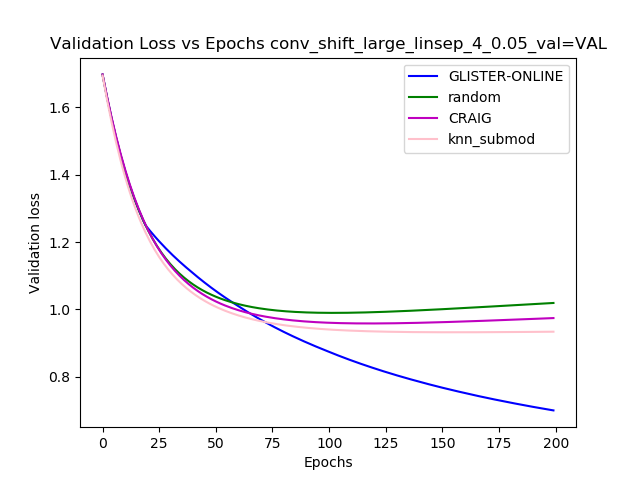}
        \caption{\label{val_plot_four_close_shift}}
    \end{subfigure}
    \hfill
    \begin{subfigure}[b]{0.24\textwidth}
        \centering
        \includegraphics[width=\textwidth, height=3cm]{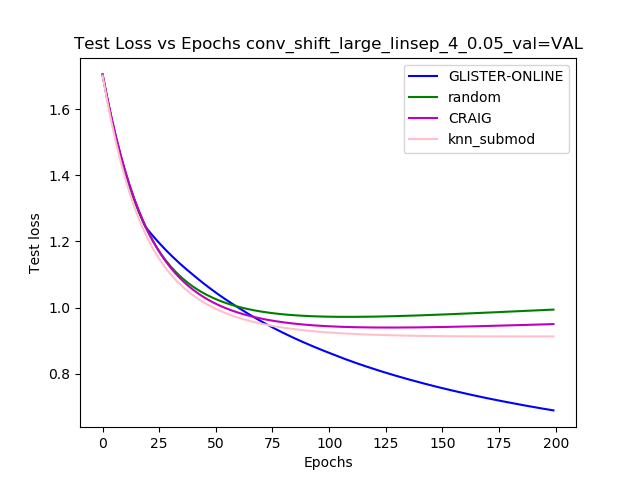}
        \caption{\label{tst_plot_four_close_shift_val}}
    \end{subfigure}
    \small{
    \caption{
    (a) Training Data (b) Validation data, (c) Validation loss (d) Test loss}
    \label{fig:synthetic_data_dss_four_close_shift}
    }
\end{figure*}

\begin{figure*}[!htpb]
    \centering
    \begin{subfigure}[b]{0.24\textwidth}
        \centering
        \includegraphics[width=\textwidth, height=3cm]{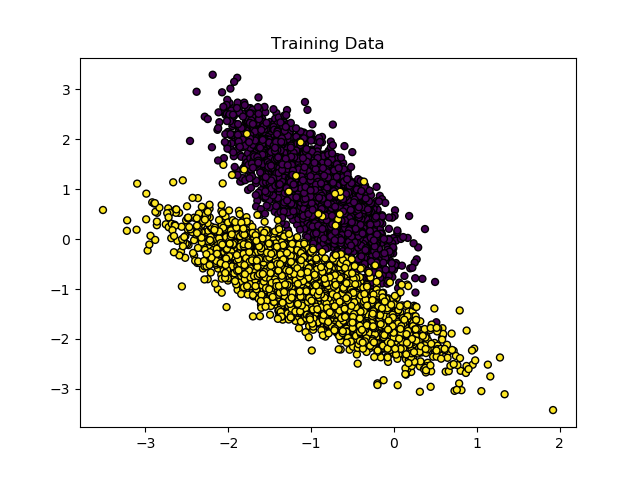}
        \caption{\label{synthetic_subset_binary_shift}}
    \end{subfigure}
    \hfill
    \begin{subfigure}[b]{0.24\textwidth}
        \centering
        \includegraphics[width=\textwidth, height=3cm]{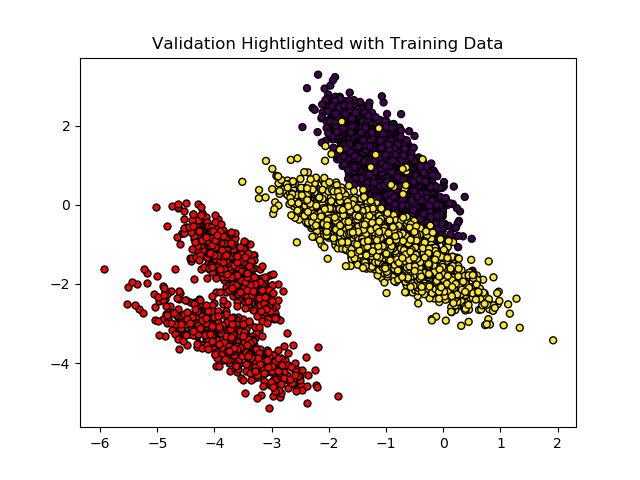}
        \caption{\label{synthetic_subset_binary_shift_val}}
    \end{subfigure}
    \hfill
    \begin{subfigure}[b]{0.24\textwidth}
        \centering
        \includegraphics[width=\textwidth, height=3cm]{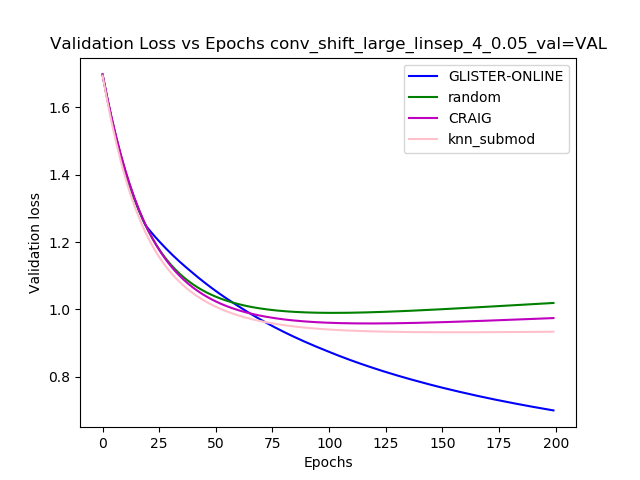}
        \caption{\label{val_plot_binary_shift}}
    \end{subfigure}
    \hfill
    \begin{subfigure}[b]{0.24\textwidth}
        \centering
        \includegraphics[width=\textwidth, height=3cm]{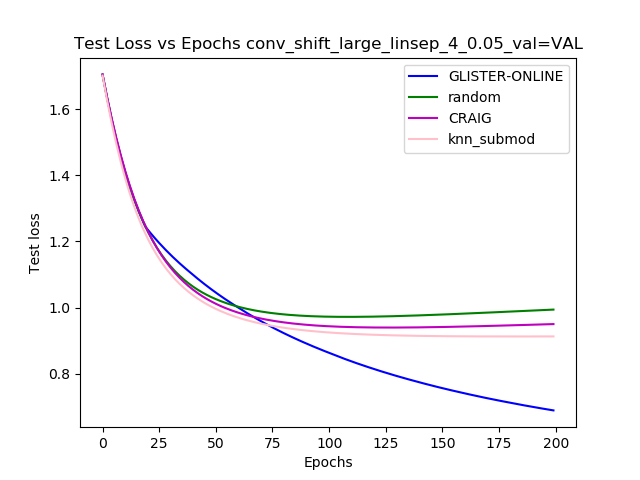}
        \caption{\label{tst_plot_binary_shift_val}}
    \end{subfigure}
    \small{
    \caption{
    (a) Training Data (b) Validation data, (c) Validation loss, (d) Test loss}
    \label{fig:synthetic_data_dss_binary_shift}
    }
\end{figure*}

Since \modelonline\  is driven by validation data, \modelonline\ can better handle situations where there is distribution shift in test and validation data from the training data, This what is called as the covariate shift. To illustrate this we use two synthetic datasets which are very similar to dataset, as shown in figure~\ref{synthetic_subset_four_close} and figure~\ref{synthetic_subset_bin_out} except that their validation dataset is shifted as shown in figure~\ref{synthetic_subset_four_close_shift_val} and figure~\ref{synthetic_subset_binary_shift_val} respectively. Figures \ref{val_plot_four_close_shift} and \ref{tst_plot_four_close_shift_val} shows how effective the methods - CRAIG\cite{mirzasoleiman2019coresets}, Random, KNNSubmod \cite{wei2015submodularity} and \modelonline\ are reducing validation and test loss respectively. Clearly \modelonline\ outperforms other methods. A similar trend is seen Figures \ref{val_plot_binary_shift} and \ref{tst_plot_binary_shift_val} for the binary dataset.

\end{document}